%% file: main.tex
\documentclass[]{onecat}
\usepackage{caption} 
\usepackage{hyperref}
\usepackage{cleveref}
\usepackage{verbatim}
\usepackage{amssymb} 
\usepackage{graphicx}
\usepackage{floatrow}
\usepackage{subcaption}
\usepackage{listings}
\usepackage{subcaption}
\usepackage{algorithm}
\usepackage{graphicx}
\usepackage{floatrow}
\usepackage{makecell}
\usepackage{wrapfig}  

\usepackage{subcaption} %

\usepackage[toc,page,header]{appendix}

\usepackage{minitoc}
\newcommand{\ourmodel}{OneCAT}

\title{\includegraphics[width=0.075\textwidth]{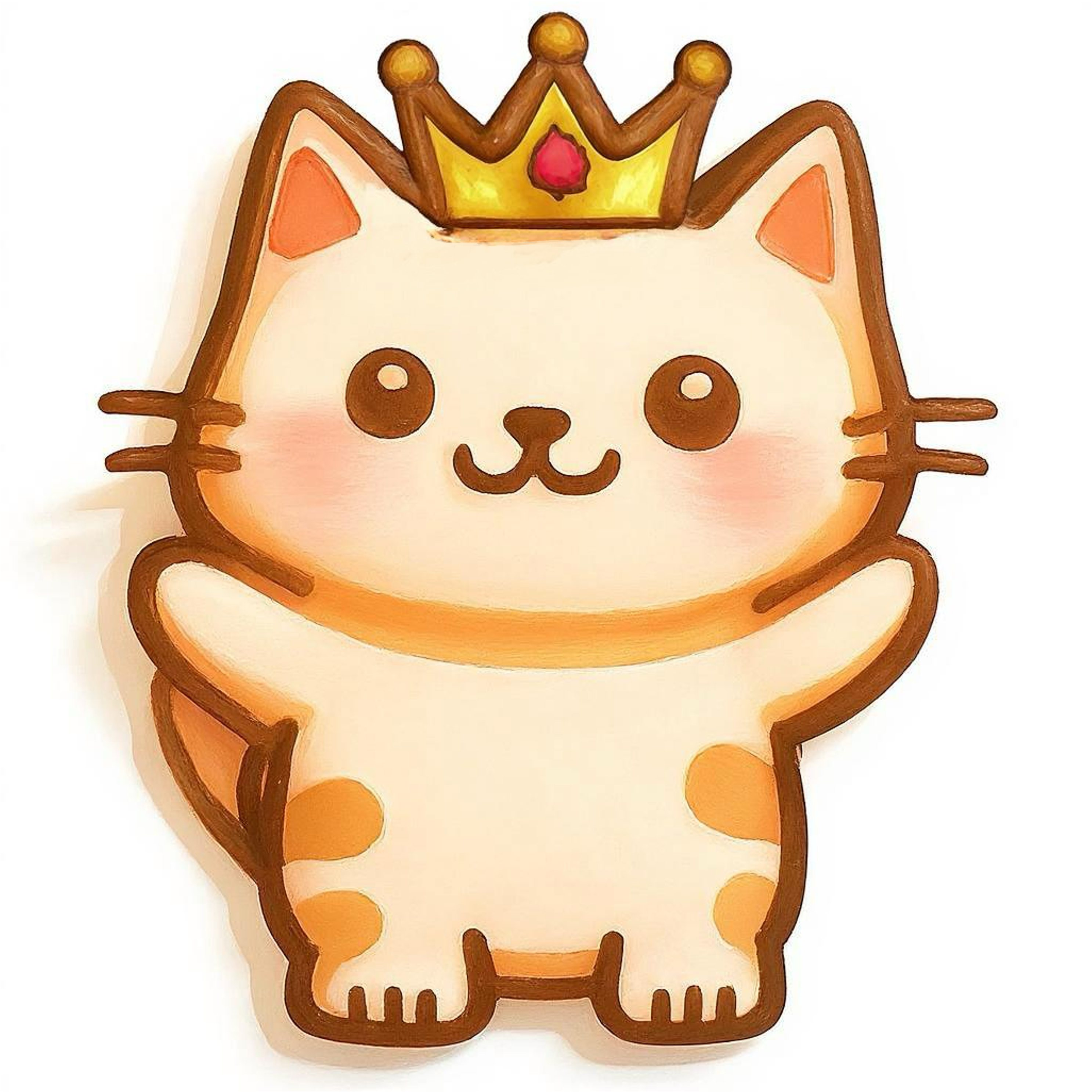}OneCAT: Decoder-Only Auto-Regressive Model for Unified Understanding and Generation}

\author{%
\parbox{\textwidth}{\centering
Han Li$^{1,2*}$, Xinyu Peng$^{1,2*}$, Yaoming Wang$^{1\P\dagger}$, Zelin Peng$^{1,2\ddagger}$, Xin Chen$^{1}$\\[2mm]Rongxiang Weng$^{1}$, Jingang Wang$^{1\P}$, Xunliang Cai$^{1}$, Wenrui Dai$^{2\P}$, Hongkai Xiong$^{2}$
}}

\affiliation{%
\parbox{\textwidth}{\centering\small
$^1$Meituan Inc,
$^2$Shanghai Jiao Tong University,
}}

\contribution[*]{Equal contribution}
\contribution[\P]{Corresponding Author}
\contribution[\dagger]{Project lead}

\footnotetext[3]{Work was done during their internship.}

\abstract{

We introduce \ourmodel, a unified multimodal model that seamlessly integrates understanding, generation, and editing within a single decoder-only transformer architecture. OneCAT uniquely eliminates the need for external components such as Vision Transformers (ViT) or vision tokenizer during inference, leading to significant efficiency gains, especially for high-resolution image inputs and outputs. This is achieved through a modality-specific Mixture-of-Experts (MoE) design trained with a unified autoregressive (AR) objective, which also natively supports dynamic resolutions. Furthermore, we pioneer to achieve multi-scale visual autoregressive mechanism within the Large Language Model (LLM) with proposed scale-aware adapter (SAA) that drastically reduces decoding latency compared to diffusion-based methods while maintaining state-of-the-art performance. Our findings demonstrate the powerful potential of pure autoregressive modeling as a sufficient and elegant foundation for unified multimodal intelligence. As a result, \ourmodel{}  outperforms existing  unified models across benchmarks for multimodal understanding, generation, and editing.

}
\date{\today}
\checkdata[Corresponding]{\url{wangyaoming03@meituan.com;wangjingang02@meituan.com;daiwenrui@sjtu.edu.cn}}
\checkdata[Project Page]{\url{https://onecat-ai.github.io/}}

\begin{document}
\maketitle
\vspace{-15pt}
\input{sections/introduction}

\input{sections/related_work}

\input{sections/method}
\input{sections/results}
\input{sections/conclusion}

\clearpage

\bibliographystyle{plainnat}
\bibliography{main}

\input{sections/appendix}

\end{document}

%% file: sections/introduction.tex
\section{Introduction}
Modular approaches using separate modules for understanding~\citep{Qwen2-VL_arxiv_2024,Qwen2.5_arxiv_2025,InternVL2.5_arxiv_2024}, generation~\citep{DALL-E_3_arxiv_2020,FLUX.1-dev_github_2024,SD3-Medium_ICML_2024}, and editing tasks~\citep{Magicbrush_2023_nips,Instruct-Pix2Pix_2023_CVPR,ICEdit_arxiv_2025} are becoming dominant for multi-modal frameworks. Despite producing capable systems, such designs inherently create complex multi-stage pipelines suffering from architectural bottlenecks that limit deep, early-stage fusion of cross-modal information and introduce significant inference latency, presenting a major barrier to both efficiency and performance.
Unified multimodal LLMs aim to address these limitations by integrating these disparate abilities within a single end-to-end architecture~\citep{Emu3-Gen_arxiv_2024,  Janus-pro_arxiv_2025,BAGEL-7B_arxiv_2025}, but many methods remain tethered to the modular paradigm~\citep{Metaqueries_arxiv_2025,BLIP3-O_arxiv_2025}. This motivates us to achieve a true revolution on the fundamental architecture that unlocks the full potential of unified systems and eschews heavy external components. In this paper, we propose a single decoder-only autoregressive model trained under a unified objective to provide elegant and potent foundation for general-purpose multimodal intelligence.

\textbf{Unified architecture.} 
We propose the first encoder-free framework for unified multimodal LLM (MLLM), where raw visual inputs are directly tokenized into patch embeddings and are processed alongside text tokens within a single decoder model. The critical innovation is a modality-specific MoE layer that dynamically routes continuous vision tokens, discrete vision tokens, and text tokens to specialized experts. It enables deep and early-stage feature fusion without requiring exquisite encoders (\textit{i.e.}, ViT and vision tokenizer) for efficient inference. For generative tasks, we pioneer to embed the multi-scale autoregressive mechanism~\citep{VAR_nips_2024} into the LLM and propose a scale-aware adapter (SAA) to extract scale-specific representation for augmentation. Thus, image tokens can be predicted from low to high resolutions, while next-token prediction is adopted for text tokens. This design significantly enhances the speed and quality of image generation by simultaneously circumventing the high latency of diffusion models and learning a coarse-to-fine generative process.

\begin{figure}[H]     
  \vspace*{-0.05\textwidth}
  \hspace*{-0.145\textwidth}
  
\includegraphics[width=1.0\textwidth]{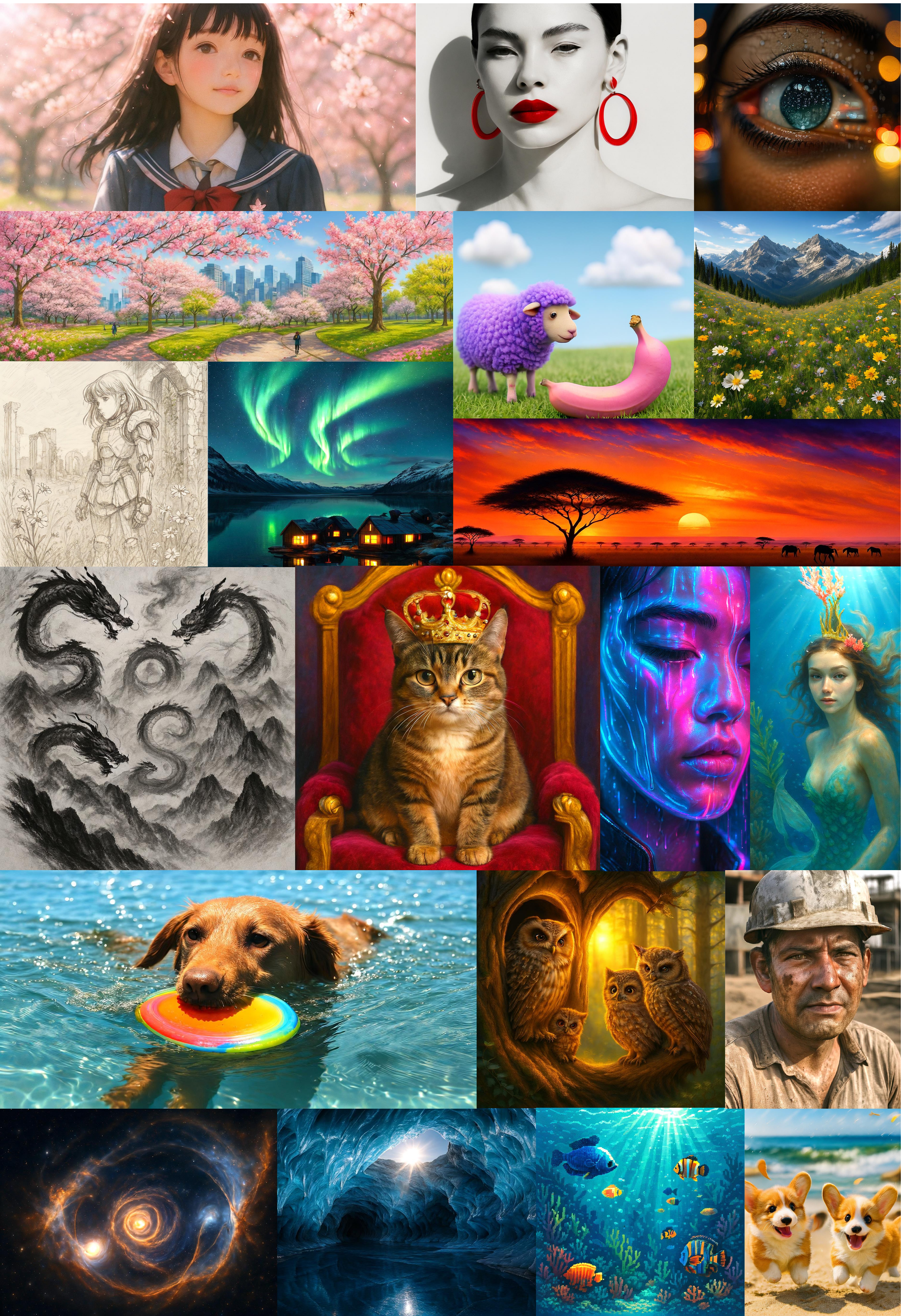}
  \caption{Showcase of the text-to-image generation abilities of the \textbf{\ourmodel{}} model.}
  \label{fig:1}
\end{figure}

\begin{figure}[H]     
  \vspace*{-0.05\textwidth}
  \hspace*{-0.145\textwidth}
  
\includegraphics[width=1.0\textwidth]{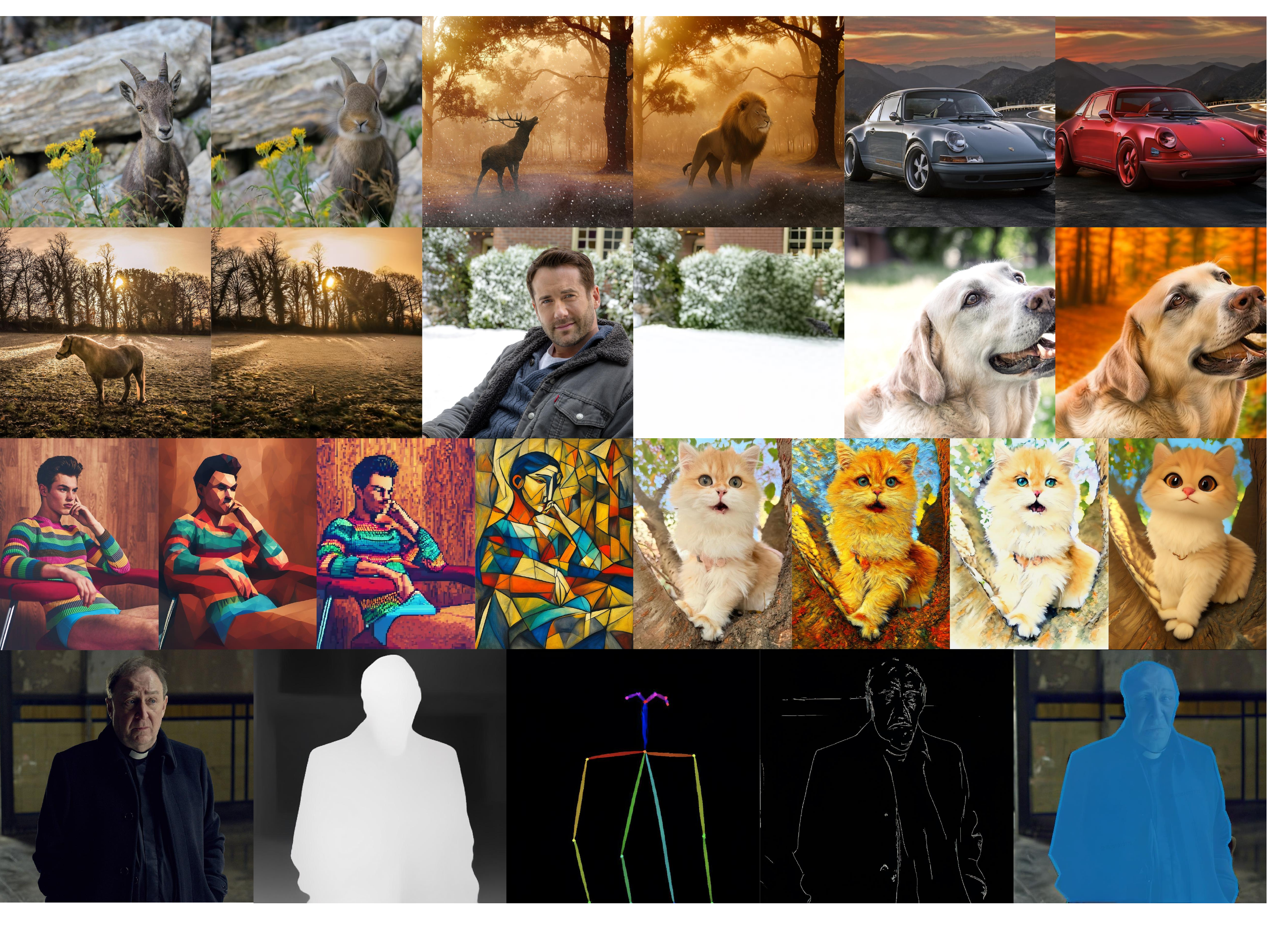}
\caption{Showcase of the image editing abilities of the \textbf{\ourmodel{}} model, including general image editing tasks such as object removal, background adjustment, color adjustment, subject replacement, and style transfer; as well as perceptual tasks including depth estimation, pose estimation, object segmentation, and Canny edge detection.}
  \label{fig:2}
\end{figure}

\textbf{Unified training paradigm.} Training encoder-free MLLMs for strong visual
perception ability is notoriously data-intensive~\citep{Mono-InternVL_CVPR_2025}. To mitigate this, existing methods like VoRA~\citep{vora_2025_arxiv} and EvE~\citep{EvE_nips_2024} perform distillation to align the internal hidden states of an LLM student with a pre-trained ViT teacher. However, they suffer from supervisory bottleneck problem that the expressive capacity of an LLM is restricted by a small teacher. We address this problem with a novel distillation strategy that first customizes a powerful MLLM teacher and then efficiently transfers its comprehensive visual perception to the proposed encoder-free unified MLLM. On such basis, we use large-scale heterogeneous multimodal data and employ a unified expert pretraining, mid-training, and supervised fine-tuning to force the shared decoder to achieve generalized representation that can seamlessly switch between comprehension, generation, and editing tasks.

Building upon these innovations, we present \underline{On}ly D\underline{eC}oder \underline{A}uto-regressive \underline{T}ransformer (\ourmodel{}), a unified multimodal model. Comprehensive evaluations demonstrate that \ourmodel{} sets a new state-of-the-art for unified models. More importantly, the encoder-free design provides a significant inference speedup, particularly for high-resolution inputs and outputs. \ourmodel{} demonstrates the viability and superiority of a pure decoder architecture, and offers a more first-principle-aligned paradigm for multimodal modeling. It facilitates earlier cross-modal fusion through its unified MoE structure and enhances semantic consistency via unified AR objective to provide valuable insights and a powerful new baseline for developments of next-generation unified multimodal systems. For examples of our model's impressive image generation capabilities, please refer to Fig.~\ref{fig:1}. Its advanced image editing functionalities are further showcased in Fig.~\ref{fig:2}.

%% file: sections/related_work.tex
\section{Related Work}
\subsection{Compositional MLLMs}

The field of Multimodal Large Language Models (MLLMs) has rapidly evolved, converging on a dominant \textbf{compositional architecture}. This paradigm connects a pre-trained vision encoder (\textit{i.e.,} CLIP~\citep{MLLMs_2021_ICML_CLIP}, SigLIP~\citep{MLLMs_2023_arxiv_siglip}, and InternViT~\citep{internvit}), to a powerful LLM through a trainable connector. Pioneering works~\citep{VLLMs_flamingo_2022_nips, VLLMs_ICML_2023_BLIP_2} propose sophisticated connector designs. For example,  Flamingo~\citep{VLLMs_flamingo_2022_nips} introduces gated cross-attention layers to inject visual information into an LLM, while BLIP-2~\citep{VLLMs_ICML_2023_BLIP_2} develops the Q-Former to bridge the modality gap between the vision encoder and LLM. A significant shift occurs with LLaVA~\citep{VLLMs_MLP_Llava_2023_nips}, which simplifies the connector to a lightweight MLP, which become a foundational blueprint for subsequent MLLMs. For example, recent state-of-the-art models like the InternVL series~\citep{internvit,InternVL2.5_arxiv_2024, InternVL2_arxiv_2024,zhu2025internvl3exploringadvancedtraining,internvl3.5} and the Qwen-VL series~\citep{Qwen2-VL_arxiv_2024,Qwen2.5_arxiv_2025,Qwen-VL_arxiv_2023} adopt this core principle and achieve superior performance by leveraging larger-scale training data and more powerful vision and language foundation models. However, this successful compositional design has inherent drawbacks. The separate nature of the vision and language components complicates the end-to-end optimization process and introduces two critical bottlenecks. First, the sequential nature of the architecture, where the vision encoder must fully process an image before the LLM can begin its generation, leads to high inference latency, especially for the \textbf{prefilling} stage. Second, the connector acts as an information bottleneck. In this so-called \textbf{late fusion} pipeline, complex visual information is compressed into a compact representation for the LLM, inevitably causing a loss of fine-grained visual detail. These fundamental limitations are now motivating a shift in the field towards more deeply integrated, such as decoder-only models, that aim to overcome these challenges.

\subsection{Decoder-only MLLMs}

Decoder-only MLLMs, also known as monolithic MLLMs, have recently emerged as a minimalist yet powerful alternative to the  compositional MLLMs. This paradigm aims to achieve greater efficiency and a more direct \textbf{early fusion} of modalities by removing the separate vision encoder or tokenizer. For example, Fuyu-8B~\citep{fuyu-8b_2023} processes vision patches by feeding them through a simple linear patch embedding layer directly into the LLM, which markedly reduce inference latency. Inspired by this success, subsequent works~\citep{EvE_nips_2024,Mono-InternVL_CVPR_2025,EvEv2.0_arxiv_2025,SAIL_2025_ICCV,vora_2025_arxiv,shukor2025scaling}, further advance decode-only MLLMs by targeting their training processes and architectures. EvE~\citep{EvE_nips_2024} and VoRA~\citep{vora_2025_arxiv} aligns the LLM's hidden states with semantic features from a pre-trained ViT. However, directly using a smaller model (e.g., a ViT with hundreds of millions of parameters) as the teacher to distill knowledge into a significantly larger LLM (with several billion parameters) may restrict the expressive capacity of the LLM.  Differently, Mono-InternVL~\citep{Mono-InternVL_CVPR_2025} and EvEv2.0~\citep{EvEv2.0_arxiv_2025} adopt a Mixture-of-Experts (MoE) framework, introducing a dedicated \textit{visual expert} to handle visual-specific features more effectively. HoLVE~\citep{HoVLE_CVPR_2025} prepends a causal transformer to the LLM to explicitly convert both visual and textual inputs into a shared space. Despite these promising advancements, the overall training efficiency of decoder-only MLLMs remains a significant challenge. More importantly, the potential for the decoder-only architecture to create unified models that can seamlessly integrate multimodal understanding, generation, and even image editing capabilities remains a largely unexplored research avenue.

\subsection{Unified Visual Understanding and Generation
}
Building on the success of MLLMs, the convergence of visual understanding and generation into a unified framework now represents a key research frontier~\citep{Janus_CVPR_2025,Janus-pro_arxiv_2025,xie2025reconstructionalignmentimprovesunified,Transfusion-7B_arxiv_2024,Emu3-Gen_arxiv_2024,show-o_arxiv_2024,li2025synergen,Metaqueries_arxiv_2025,BLIP3-O_arxiv_2025,UniWorld-V1_arxiv_2025,BAGEL-7B_arxiv_2025,Mogao_arxiv_2025}.
Pioneering unified MLLMs such as 
Chameleon~\citep{Chameleon-7B_arxiv_2024},Transfusion~\citep{Transfusion-7B_arxiv_2024}, emu3~\citep{Emu3-Gen_arxiv_2024}, show-o~\citep{show-o_arxiv_2024} and Synergen-VL~\citep{li2025synergen} utilize vison tokenizer (e.g., VQ-VAE) to convert images into discrete tokens, enabling seamless multimodal understanding and generation within a single model. However, the discretization inevitably results in lossy visual information and weakens in extracting semantic contents.  Janus series~\citep{Janus_CVPR_2025,Janus-pro_arxiv_2025} decouples visual encoding for understanding and generation using two separate encoders, but may compromise performance due to task conflicts in shared LLM parameter space.
Metaqueries~\citep{Metaqueries_arxiv_2025}, BLIP3-O~\citep{BLIP3-O_arxiv_2025}, Uniworld-V~\citep{UniWorld-V1_arxiv_2025} assembles off-the-shelf specialized MLMMs and diffusion models by tuning adapters and learnable query tokens,  which sacrifices true architectural unification for modularity.
BAGEL~\citep{BAGEL-7B_arxiv_2025} and Mogao~\citep{Mogao_arxiv_2025} employ a Mixture-of-Transformers (MoT) architecture, dedicating different components to autoregressive text generation and diffusion-based visual generation. However, while powerful, this hybrid approach inherits the significant inference latency of diffusion models and still requires separate encoders and tokenizers during the inference.

In contrast to these approaches, our OneCAT introduces a pure decoder-only architecture. By integrating modality-specific experts directly within the decoder, OneCAT achieves versatile multimodal capabilities without the need for external vision encoders or tokenizers at inference time, thus resolving the trade-off between architectural purity and inference efficiency.

\subsection{Next Scale Prediction for Visual Generation}
Autoregressive models based on next-token prediction(NTP) have long faced efficiency challenges in high-resolution image generation due to the quadratic growth of sequence length with image size. Similarly, diffusion models—though widely successful—often suffer from slow iterative sampling. To address these limitations, VAR~\citep{VAR_nips_2024} introduced the next-scale prediction(NSP) paradigm, which encodes images into hierarchical discrete tokens via a multi-scale VAE and generates them autoregressively from low to high resolution, significantly reducing the number of decoding steps. Building upon this, Infinity~\citep{infinity_CVPR_2025} further enhanced this approach with bit-level prediction and extended tokenizer vocabulary, achieving superior generation quality while maintaining efficient inference. To enable unified understanding and generation, VARGPT~\citep{zhuang2025vargpt} stack the transformer from pretrained VAR~\citep{VAR_nips_2024} as a visual decoder atop a LLM. However, since the visual tokens (\textit{i.e.,} the input of the visual decoder) \textbf{must be decoded token-by-token} through the LLM before subsequent next-scale prediction, this approach compromises the inference efficiency that is the key advantage of the NSP.

In contrast, our proposed OneCAT seamlessly unifies next-token prediction for text generation and next-scale prediction for visual generation \textbf{within a single decoder-only transformer of the LLM}, and proposes the scale-aware adapter to further exploit the scale-specific representation for enhanced visual generation.

%% file: sections/method.tex
\section{OneCAT}
\label{sec-architecture}
\begin{figure}[!t]
\centering
\includegraphics[width=0.99 \linewidth]{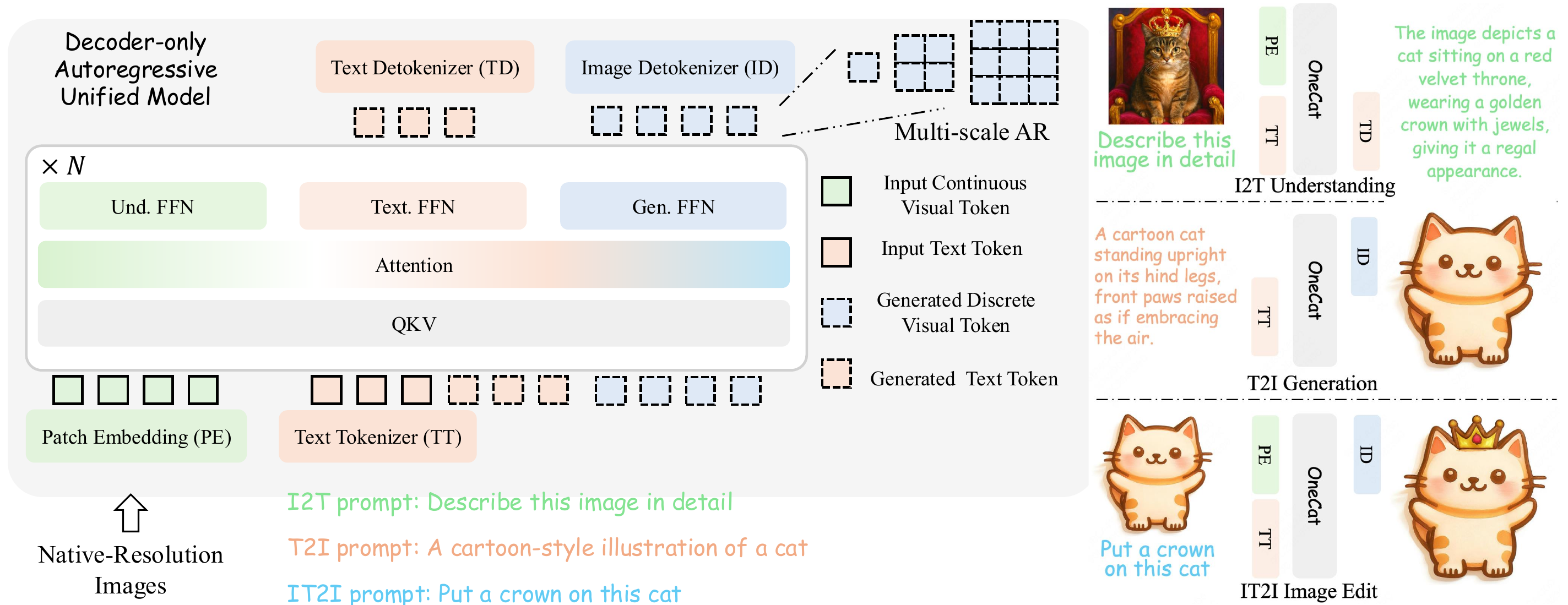}
\caption{Inference pipeline of OneCAT, a decoder-only autoregressive unified model that seamlessly supports multimodal understanding, image generation and image editing. }
\label{figs-framework}
\end{figure}

As depicted in Fig.~\ref{figs-framework}, \ourmodel{} employs a pure decoder-only architecture and eliminates the need for additional vision encoder or tokenizer \textbf{during inference}. This streamlined design significantly simplifies the model structure and reduces computational cost. Unlike existing unified MLLMs~\citep{Janus-pro_arxiv_2025,BAGEL-7B_arxiv_2025} using semantic encoders like ViTs for understanding, we follow recent encoder-free MLLMs~\citep{Mono-InternVL_CVPR_2025} and use a lightweight \texttt{Patch Embedding} layer to convert raw images into continuous visual tokens for efficient and lossless processing. The same \texttt{Patch Embedding} layer is also used to process reference images for editing tasks, thereby superseding separate VAE tokenizer  used~\citep{BAGEL-7B_arxiv_2025,Mogao_arxiv_2025} and further enhancing inference efficiency.

For text generation, \ourmodel{} adheres to the \textbf{Next-Token Prediction} paradigm. For visual generation, it innovatively employs the \textbf{Next-Scale Prediction}~\citep{VAR_nips_2024}, where  images are generated in a hierarchical  coarse-to-fine manner to progressively from the lowest to the highest resolution scales. Refer to Appendix~\ref{sec-preliminary-VAR} for more preliminaries for next-scale prediction.

\subsection{Modality-MoE}

\ourmodel{} proposes a Modality-MoE architecture with three specialized feed-forward network (FFN) experts: a  \texttt{Text.~FFN} designed for text tokens for language comprehension, a  \texttt{Und.~FFN} designed for continuous visual tokens for visual understanding, and a \texttt{Gen.~FFN} for discrete visual tokens for image generation. We employ a hard routing mechanism that assigns tokens to a specific expert based on their modality and the task at hand. Other QKV and attention layers are shared to promote  parameter efficiency and robust cross-modal alignment for instruction-following.

OneCAT is initialized from the pre-trained Qwen2.5 LLM~\citep{qwen2.5} to exploit its strong ability in language modeling. To construct the proposed Modality-MoE, we replicate the FFN layer from each Qwen2.5 transformer block to form three distinct experts. The core functionality for each multimodal task is handled as below.

\paragraph{Multimodal Understanding.} We employ a simple yet effective patch embedding layer to convert raw images into  continuous visual tokens. This layer consists of a 14$\times$14 convolution for image patchifying, a 2$\times$2 pixel unshuffle layer, and a two-layer Multilayer Perceptron (MLP) for projecting the visual features to match the LLM's hidden state dimension. 

\paragraph{Text-to-Image Generation.} We leverage a pre-trained multi-scale VAE model from Infinity~\citep{infinity_CVPR_2025} to map images between the pixel and latent spaces. This VAE operates with a downsampling ratio of 16 and a latent channel size of 32, and uses a bitwise quantizer~\citep{BSQ_2024_arxiv} to enlarge the vocabulary. During training, the VAE tokenizer encodes target images into multi-scale discrete visual tokens to serve as ground-truth. During inference, this tokenizer is \textbf{not required}, and only the detokenizer is used to reconstruct the image from generated multi-scale visual tokens. 

\paragraph{Image Editing.} OneCAT seamlessly supports image editing task by leveraging a reference image and instruction input. 
The reference image is processed by the patch embedding layer, and the resulting continuous visual tokens are also routed to the \texttt{Und.~FFN} to serve as the visual condition. The patch embedding layer provides a \textbf{near-lossless representation} of the reference image, and allows the LLM's shallower layers to obtain low-level features for pixel-level consistency, while the deeper layers to extract high-level features for semantical comprehension. Guided by this hierarchical visual context, LLM predicts new discrete visual tokens autoregressively. This design enables powerful conditional generation without any architectural modifications, showing the versatility of our unified decoder-only design.

\subsection{Scale-Aware Adapter for Hierarchical Generation}
\label{sec:sa_lora}

The tokens produced by the multi-scale VAE are inherently hierarchical (as shown in  Appendix~\ref{sec-visual-saa}).  Specifically, lower-scale tokens mainly encode low-frequency global information such as color, illumination, and coarse structure, while higher-scale tokens capture high-frequency details including fine textures and intricate patterns. Processing these functionally divergent tokens equally with a shared \texttt{Gen.~FFN} layer is not optimal and could limit the representation capacity. 

To address this, we introduce the \textbf{Scale-Aware Adapter (SAA)}, a novel architectural component integrated with the \texttt{Gen.~FFN}. The SAA comprises a set of parallel modules that serve as skip connections over each linear layer of the \texttt{Gen.~FFN}. Each SAA is dedicated to processing tokens from a specific scale, with the total number of SAA of a \texttt{Gen.~FFN} matching the number of VAE scales. During inference, discrete visual tokens are routed to corresponding scale-specific adapters based on scale indices. To ensure parameter efficiency, each adapter is achieved using a low-rank decomposition (rank $r$= 64), inspired by LoRA~\citep{LORA_ICLR_2022}. However, unlike LoRA that is typically used for fine-tuning, the SAA modules are jointly trained in an end-to-end manner as permanent components of the LLM.

\subsection{Multimodal versatile attention mechanism}
\begin{figure}[!t]
\centering
\includegraphics[width=1\linewidth]{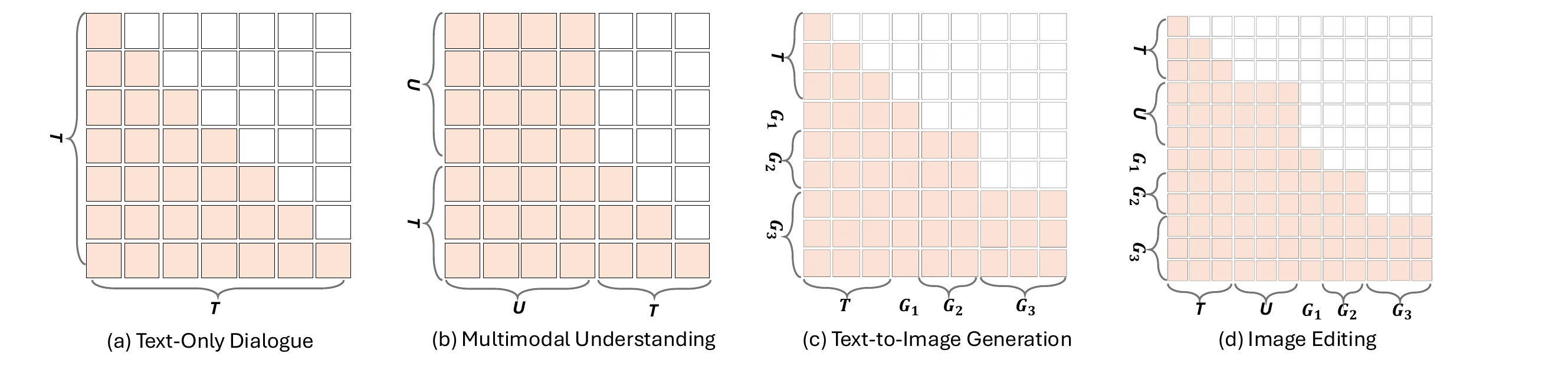}
\caption{Multimodal versatile attention mechanism. $T$ denotes the text tokens. $U$ denotes the continuous visual tokens for multimodal understanding or reference image tokens for image editing. $G_i$ denotes the $i$-th scale discrete visual tokens for visual generation.}
\label{figs-mask}
\end{figure}

We employ a multimodal versatile attention mechanism based on  FlexAttention~\citep{pytorch_team_flexattention_2024} to enable flexible processing of diverse modalities and tasks within a single LLM. As shown in Fig.~\ref{figs-mask}, text tokens $T$ use causal attention for autoregressive generation, while continuous visual tokens $U$ apply full attention for global interaction. Multiscale discrete visual tokens $G$ use block causal attention. Tokens within the same scale attend to each other freely, while attention across scales follows a causal attention.

\section{Model Training Pipeline}
\begin{figure}[!t]
\centering
\includegraphics[width=0.99\linewidth]{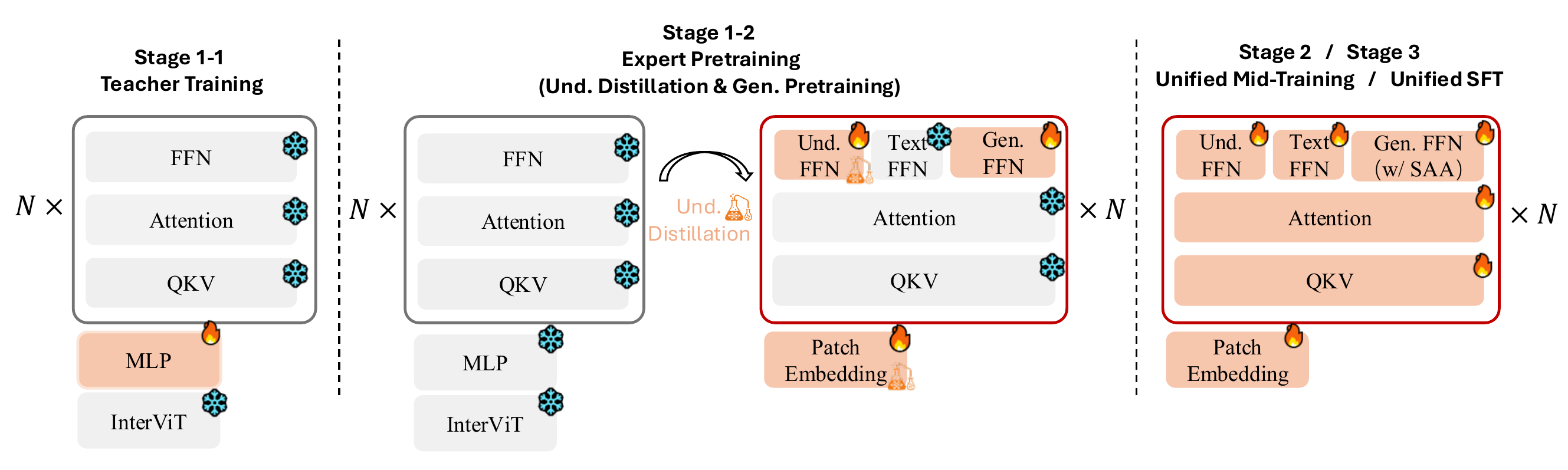}
\caption{Overview of the training pipeline. In Stage 1, we first prepare a teacher model by training a two-layer MLP to connect InternViT~\citep{internvit} and the Qwen2.5 LLM~\citep{qwen2.5}. This teacher model is then used to perform understanding distillation for the \texttt{Und.~FFN} and the \texttt{Patch  Embedding} layer. Simultaneously, we perform generation pretraining to optimize the \texttt{Gen.~FFN}. All other parameters of the LLM remain frozen to preserve its pretrained language capabilities. In Stage 2 and 3, the entire model is unfrozen to conduct unified mid-training and supervised fine-tuning (SFT), respectively. The VAE component for visual generation is omitted from the figure for clarity.}
\label{figs-training-pipeline}
\end{figure}

As shown in Fig.~\ref{figs-training-pipeline} and Table~\ref{tab:training_recipe}, we employ a three-stage training pipeline for OneCAT.

\subsection{Stage-1: Multilmodal Pretraining}
The objective of this stage is to equip the OneCAT with foundational visual perception and generation abilities while maintaining the linguistic capabilities from the pretrained LLM. The core challenge is that, for visual perception, the \texttt{Und.~FFN} is initialized from the weights of the LLM's text-focused FFN. While this "warm start" benefits abstract reasoning, it inherently lacks pretrained visual knowledge, making the training process highly data-intensive.
To address this limitation, we  leverage the visual perception  capabilities of an MLLM teacher and introduce an understanding distillation strategy to optimize \texttt{Und.~FFN} that significantly enhances visual learning efficiency. In parallel, we conduct generation pretraining to optimize \texttt{Gen.~FFN}.

\subsubsection{Stage 1-1: Teacher Training}
Instead of using an off-the-shelf MLLM as teacher (\textit{e.g.}, Qwen2.5-VL~\citep{Qwen2.5_arxiv_2025}), we customize a teacher to ensure parameter consistency between the LLM backbones of the teacher and student, and thus, improve distillation stability and efficiency (refer to Fig.~\ref{fig:abla}). Specifically, 
the teacher is built by connecting a pre-trained ViT (\texttt{InternViT}~\citep{internvit}) and a LLM (\texttt{Qwen2.5}~\citep{qwen2.5}) with a two-layer \texttt{MLP}. We freeze both ViT and LLM and only train the MLP connector on a small-scale image-text caption dataset with NTP loss. This alignment training endows the custom MLLM teacher with strong visual perception ability.

\subsubsection{Stage 1-2: Expert Pretraining}
We then train OneCAT based on the teacher model. We freeze the \texttt{QKV}, \texttt{Attention}, and \texttt{Text FFN}, and optimize the task-specific modules: the \texttt{Und.~FFN} and \texttt{Patch Embedding} layer for visual understanding, and the \texttt{Gen.~FFN} for visual generation.

\begin{table}[t!]
\centering
\resizebox{\textwidth}{!}{%
\begin{tabular}{l|rrrr}
\toprule
\textbf{Hyperparameter / Config} & \textbf{Stage 1-1} & \textbf{Stage 1-2} & \textbf{Stage 2} & \textbf{Stage 3} \\
& \small{(Teacher Training)} & \small{(Expert Pretraining)} & \small{(Unified Mid-Training)} & \small{(Unified SFT)} \\
\midrule
Learning Rate & $2 \times 10^{-3}$  & $2 \times 10^{-4}$ & $2 \times 10^{-5}$ & $1 \times 10^{-5}$ \\
LR Scheduler & Cosine & Cosine & Cosine & Cosine \\
Weight Decay & 0 & 0 & 0.01 & 0.01 \\
Gradient Norm Clip & 1.0 & 1.0 & 1.0 & 1.0 \\
Batch Size & 512 & 2048 & 512 & 256 \\
Sequence Length & 1024 & 1024 & 8192 & 16384 \\

\midrule
Number of Sample: Text-Only & - & - & 40M& 2M \\
Number of Sample: Und. & 10M & 436M & 70M & 11M  \\
Number of Sample: Gen. & - & 52M & 60M & 3M  \\
Number of Token (Total) & 5B & 0.3T & 0.6T& 57B \\
Token Ratio (T:U:G):& 0:1:0 & 0:8:1&  1:2:6 & 1:5:6 \\
\midrule
Resolution: Und. & 448$\times$448 & 448$\times$448 & Native  & Native   \\
Use  thumbnail & $\times$ & $\times$ & $\checkmark$& $\checkmark$ \\ 
\midrule
Resolution:  Gen. & - & 256$\times$256 &  \makecell[r]{Dynamical \\ (\#sides: 288$\sim$864)} & \makecell[r]{Dynamical \\ (\#sides: 288$\sim$1728)} \\
Number of Scales : Gen. &-& 7& 10 & 10$\sim$13\\
\bottomrule
\end{tabular}%
}
\caption{Detailed hyperparameter and configuration of the training recipe across different stages.}
\label{tab:training_recipe}
\end{table}

\textbf{Understanding  Distillation}:
We optimize the \texttt{Und.~FFN} on a large-scale dataset of image-to-text pairs. 
 The training objective is a combination of the NTP loss ($\mathcal{L}_{\text{NTP}}$) and a distillation loss ($\mathcal{L}_{\text{Distill}}$). Specifically, $\mathcal{L}_{\text{NTP}}$ is the cross-entropy loss for text token generation.
For distillation, instead of matching the final output logits, we align the student's internal hidden states with those of the teacher model through deep feature-level matching across each transformer layer. This enables the student to not only mimic the teacher's final text prediction but also its internal computational patterns across all token (including both visual and text tokens) for better visual knowledge transfer.  
The distillation loss is formulated as $\mathcal{L}_{\text{Distill}} = \sum_{n=1}^{N} \text{MSE}(\mathbf{h}_S^{(n)}, \mathbf{h}_T^{(n)})$, 
where $\mathbf{h}_S^{(n)}$ and $\mathbf{h}_T^{(n)}$ represent the hidden state outputs from the $n$-th transformer block of the student and teacher models, respectively. The final objective is $   \mathcal{L}_{\text{Und}} = \mathcal{L}_{\text{NTP}} + \lambda \mathcal{L}_{\text{Distill}}
$,
where $\lambda = 0.02$ is a balancing hyperparameter. Throughout distillation, all models process images at a fixed resolution of $448 \times 448$ to balance computational load and the granularity of visual features.

\textbf{Generation Pretraining}:
In parallel, we optimize the \texttt{Gen.~FFN} on a delicate text-to-image generation dataset to enable the LLM to learn the spatial relationships and   cross-scale dependencies. We also adopt the cross-entropy loss for next-scale prediction~\citep{VAR_nips_2024} and the output image resolution is fixed to 256$\times$256.

\subsection{Stage-2: Unified Mid-Training}
In Stage-2, we unfreeze the entire model to perform unified mid-training across multiple tasks (\textit{i.e.,} image-to-text, text-to-image, image editing, and text-only dialogues). We incorporate the proposed \texttt{scale-aware adapter} in this stage, which is optimized with other modules together to extract scale-specific representation for enhanced image generation quality. 
We also introduce native resolution strategy in this stage. 
For visual understanding, the model is trained to process images at their original resolutions, thereby preserving fine-grained details and reducing information loss. Additionally, a thumbnail of resolution 448$\times$448 is included to provide global visual context.
For visual generation,  we train the model  with dynamical resolution and aspect ratios where the side lengths sampled from a range of 288 to 864 pixels, which enhancing its generation versatility and real-world applicability.

\subsection{Stage-3: Unified Supervised Fine-tuning }

The final stage involves unified supervised fine-tuning (SFT) using a curated dateset of higher-quality data to enhance instruction-following and visual generation quality. The native resolution strategy was continued, with the size of generated image  expanded to support side lengths between 288 to 1776 pixels, enabling higher-resolution results.

%% file: sections/results.tex
\section{Data Setup}
\textbf{Stage-1:}
For the multimodal understanding, we curate a large-scale dataset of approximately 436 million image-text pairs, 
which is meticulously compiled and processed through comprehensive filtering and deduplication. This dataset is collected from two primary sources:
(1) Public Available Image-Text Caption Pairs: We incorporate several publicly available, high-quality image-caption datasets, including Recap-DataComp-1B~\citep{li2024if}, Capsfusion~\citep{yu2024capsfusion}, Detailed-Caption~\citep{li2024monkey}, SA1B-Dense-Caption~\citep{Tongyi_SA1B}, and Moondream2-COYO-5M-Captions~\citep{isidentical_2024}.
(2) Re-captioned Image Datasets: We generate new image-text pairs by re-captioning large-scale public image collections using Qwen2-VL~\citep{Qwen2-VL_arxiv_2024}. The source image datasets for this process include COYO700M~\citep{kakaobrain2022coyo-700m}, CC12M~\citep{changpinyo2021conceptual}, CC3M~\citep{sharma2018conceptual}, LAION-400M~\citep{schuhmann2021laion}, and Zeor250M~\citep{xie2023ccmb}.
From this large-scale dataset, we randomly sample a small-scale subset of 10 million samples to train the custom teacher.

For image generation, we construct a dataset of 52 million text-to-image samples after a rigorous filtering process to remove samples with low resolution or poor aesthetic scores.  This collection consists of 1 million class-labeled images from ImageNet-1k~\citep{deng2009imagenet}, 20 million pairs from public collections (\textit{i.e.,}  COYO700M~\citep{kakaobrain2022coyo-700m}, LAION-400M~\citep{schuhmann2021laion} and CC12M~\citep{changpinyo2021conceptual}), and 30 million in-house synthetic images generated by FLUX. 
The overall training token ratio across multimodal understanding and visual generation samples  in Stage-I is  approximately \textbf{8:1}.

\textbf{Stage-2}:  In the unified mid-training, for multimodal understanding we leverage an curated dataset of 70 million visual instruction samples. This dataset is specifically curated to be highly diverse tasks, including general VQA, detailed image captioning, OCR, multimodal reasoning(\textit{i.e.,} STEM problem-solving), knowledge, and visual grounding, which are sourced from  Detailed-Caption~\citep{li2024monkey}, ALLaVA~\citep{chen2024allava}, ShareGPT4V~\citep{chen2024sharegpt4v}, SA1B-Dense-Caption~\citep{Tongyi_SA1B}, WIT~\citep{srinivasan2021wit}, pdfa-eng-wds~\citep{pdfa-eng-wds}, UReader~\citep{ye2023ureader}, DVQA~\citep{mathew2021docvqa}, OCR-VQA~\citep{mishra2019ocr}, WebSRC~\citep{chen2021websrc}, GQA~\citep{hudson2019gqa}, visual-genome~\citep{krishna2017visual},  GRIT~\citep{Kosmos2}, and other in-house synthetic visual instruction data.

For visual generation, we supplement the text-to-image samples of Stage-1 with a additional collection of 8 million image editing samples, resulting a total of 60 million visual generation samples. These additional image editing samples are sourced from several public image editing datasets, including AnyEdit~\citep{yu2025anyedit}, UltraEdit~\citep{zhao2024ultraedit}, HQ-Edit~\citep{hui2024hq} and OmniEdit~\citep{wei2024omniedit}. 

Additionally, we incorporate 40 million text-only instruction samples to preserve the language ability of LLM. To ensure a strong focus on visual generation in Stage-II, we oversample the visual generation data, resulting a final training token ratio of approximately \textbf{1~:2~:6} across text-only, multimodal understanding, and visual generation tasks, respectively.

\begin{table}[t]
\centering
\caption{Model configurations for the two variants of OneCAT. $^{\dagger}$ A params indicates the activated parameters and T params indicates the total parameters.}
\resizebox{0.9\textwidth}{!}{
\begin{tabular}{lccccc}
\toprule
& \multicolumn{3}{c}{{Model Recipe}} & \multicolumn{2}{c}{{Understanding Distillation}} \\
\cmidrule(lr){2-4} \cmidrule(lr){5-6}
&Base Model&{A Params$^{\dagger}$} &{T Params$^{\dagger}$} &{Teacher ViT} &{Teacher LLM} \\
\midrule
OneCAT-1.5B &Qwen2.5-1.5B-instruct &1.5B &4.5B &InternViT-300M & Qwen2.5-1.5B-instruct \\
OneCAT-3B &Qwen2.5-3B-instruct &3B &9B &InternViT-300M & Qwen2.5-3B-instruct \\
\bottomrule
\end{tabular}
}

\label{tab:model_config}
\end{table}

\textbf{Stage-3}:  In the SFT stage, for multimodal understanding and text-only instruction, we construct a high-quality dataset of 13 million samples. This dataset comprises 10 million filtered samples from MAmmoTH-VL dataset~\citep{dataset_2024_mammoth} and 3 million samples from other open-source datasets AI2D~\citep{kembhavi2016diagram}, OKVQA~\citep{marino2019ok}, VQAv2~\citep{goyal2017making}, ART500K~\citep{mao2017deepart}, ScienceQA~\citep{saikh2022scienceqa}, GQA~\citep{hudson2019gqa},  CLEVR-Math~\citep{lindstrom2022clevr}, COCO-ReM~\citep{singh2024benchmarking}, TallyQA~\citep{acharya2019tallyqa},  Docmatix~\citep{laurenccon2024building}, DVQA~\citep{mathew2021docvqa}, DreamSim~\citep{fu2023dreamsim}, ShareGPT4o~\citep{InternVL2_arxiv_2024}.

For visual generation, we utilize a total of 3 million samples, aggregated from UniWorld~\citep{lin2025uniworld}, BLIP3o-60k~\citep{BLIP3-O_arxiv_2025}, ShareGPT-4o-Image~\citep{chen2025sharegpt}, and additional synthetic data generated by GPT-4o~\citep{hurst2024gpt} and FLUX~\citep{FLUX.1-dev_github_2024} using the partial prompts from JourneyDB~\citep{sun2023journeydb}.
The overall training token ratio across text-only, multimodal understanding, and visual generation for unified sft is 
approximately \textbf{1~:5~:6}.

\section{Implementation details}

\textbf{Model Configurations:}  We conduct experiments on two model variants, OneCAT-1.5B and OneCAT-3B, which is initialized from Qwen2.5-1.5B-instruct and Qwen2.5-3B-instruct, respectively.  The total parameters count of OneCAT-3B is 9B, but the activated parameters count for each token is 3B during forward process. Refer to Table~\ref{tab:model_config} for more details.

\begin{table*}[t]
\centering
\caption{Performance comparison across multiple multimodal understanding benchmarks. 
Higher scores are better, as indicated by the up-arrow ($\uparrow$). 
\textbf{A-LLM} denotes the number of activated LLM parameters, while \textbf{Vis.} indicates the parameter count of the vision encoder or tokenizer for multimodal understanding.  Chameleon~\citep{Chameleon-7B_arxiv_2024} does not report the parameter count of its vision tokenizer.  slash (/) denotes that models do not require a vision encoder or tokenizer for multimodal understanding. Best in \textbf{bold}, second best is \underline{underlined} (across unified models).}
\label{tab:multi-modal-understanding-vqa}
\resizebox{0.85\textwidth}{!}{%
    \begin{tabular}{l c c ccccccc}
    \toprule
    \multirow{2}{*}{\textbf{Model}} & \multicolumn{2}{c}{\textbf{\# Params}} & \multirow{2}{*}{\textbf{TextVQA$\uparrow$}} & \multirow{2}{*}{\textbf{ChartQA$\uparrow$}} & \multirow{2}{*}{\textbf{InfoVQA$\uparrow$}} & \multirow{2}{*}{\textbf{DocVQA$\uparrow$}} & \multirow{2}{*}{\textbf{GQA$\uparrow$}} & \multirow{2}{*}{\textbf{AI2D$\uparrow$}} \\
    \cmidrule(lr){2-3}
    & \textbf{A-LLM} & \textbf{Vis.} & & & & & & & \\
    \midrule
    \multicolumn{10}{l}{\textit{Encoder-based Understanding Only Models}} \\
    \addlinespace 
    InternVL2-2B~\citep{InternVL2_arxiv_2024} & 1.8B &0.3B & 73.4 & 76.2& 58.9& 86.9&- &74.1  \\
    InternVL2.5-2B~\citep{InternVL2.5_arxiv_2024} & 1.8B & 0.3B& 74.3 &79.2 &60.9 &88.7 & -& 74.9 \\
    Qwen2-VL-3B~\citep{Qwen2.5_arxiv_2025} & 1.5B & 0.6B& 79.7&73.5 &65.5 &90.1 &- &74.7 \\
    Qwen2.5-VL-3B~\citep{Qwen2.5_arxiv_2025} & 3B & 0.6B& 79.3 &84.0 &77.1 &93.9  &- &81.6 \\
    \midrule
    \multicolumn{10}{l}{\textit{Encoder-free Understanding Only Models}} \\
    \addlinespace

    Mono-InternVL~\citep{Mono-InternVL_CVPR_2025} & 1.8B & /& 72.6& 73.7&43.0 &83.0  &59.5 &68.6 \\
    EvE~\citep{EvE_nips_2024} & 7B &/ & 56.8 &59.1 &- &-  &62.6 &61.0 \\
    EvEv2~\citep{EvEv2.0_arxiv_2025} & 7B &/ & 71.1 &73.9 &- &-  &62.9 &74.8 \\
    VoRA~\citep{vora_2025_arxiv} & 7B & /& 56.3&-&-&-&-&65.6 \\
SAIL~\citep{SAIL_2025_ICCV} & 7B&/&77.1 &-&-&-&-&76.7\\
        HoVLE~\citep{HoVLE_CVPR_2025} & 2.6B & /& 70.9 &78.6 &55.7 & 86.1& 64.9&73.0 \\
    \midrule
    \multicolumn{10}{l}{\textit{Unified Models}} \\
    \addlinespace
    Chameleon~\citep{Chameleon-7B_arxiv_2024} & 7B & -& 4.8& 2.9& 5.0&1.5 &- &46.0 \\
    Emu3~\citep{Emu3-Gen_arxiv_2024} & 8B &0.3B & 64.7& 68.6& 43.8& 76.3& 60.3&70.0 \\
    Harmon-1.5B~\citep{Harmon-1.5B_arxiv_2025} & 1.5B &0.9B &- &- &- &- & 58.9&- \\
    Show-o2-1.5B~\citep{show-o2_arxiv_2025} & 1.5B & 0.5B&- & -& -&- &60.0 & 69.0\\
    Janus-Pro-1.5B~\citep{Janus-pro_arxiv_2025} & 1.5B &0.3B &- & -&- &- & 59.3&- \\
        \rowcolor{tablerowcolor}
    OneCAT-1.5B&  1.5B &/ & \underline{67.0}& \underline{76.2}& \underline{56.3}& \underline{87.1}&60.9 &72.4 \\
    ILLUME+~\citep{ILLUME+3B_arxiv_2025}  & 3B & 0.6B& -&69.9 &44.1 &80.8  &- &74.2 \\
    VILA-U~\citep{VILA-U_arxiv_2024} & 7B & 0.4B& 60.8 & - & - & - &  60.8 & - \\
    Janus-Pro-7B~\citep{Janus-pro_arxiv_2025} & 7B &0.3B &- & -&- & -&62.0&- \\
    Tar-7B~\citep{Tar-7B_arxiv_2025} & 7B &0.4B &- & -&- &- & 61.3& -\\
    Show-o2-7B~\citep{show-o2_arxiv_2025} & 7B &0.5B &- &- &- &-  &\underline{63.1} &\textbf{78.6} \\
    \rowcolor{tablerowcolor}
    OneCAT-3B &  3B &/ & \textbf{73.9}& \textbf{81.2}& \textbf{64.8}& \textbf{91.2} &\textbf{63.1} &\underline{77.8} \\
    \bottomrule
    \end{tabular}%
}
\end{table*}

\textbf{Data Packing and Gradient Accumulation:} To optimize workload balance across distributed processes and increase training throughput, we employ a data packing strategy that concatenates multiple variable-length samples into contiguous sequences. Furthermore, to manage the gradient contributions and token ratios between modalities as in Table~\ref{tab:training_recipe}, we utilize an \textit{uneven} gradient accumulation strategy: prior to each optimizer step, we accumulate a \textit{distinct} number of micro-batches' gradients for the text and image generation tasks to obtain a gradient of desired token ratios. 
Such an approach provides fine-grained control over the effective batch sizes of different tasks, ensuring a balanced and stable joint-training.

\textbf{Unbiased Global Batch Gradients:} When training on $N$ distributed processes, naively averaging local loss can lead to biased gradients when per-process token counts vary. The ideal objective is to optimize the \textit{Global Batch Loss}, defined as the loss summed over tokens for all micro-batches, normalized by the global token count, denoted as $T_{global}$. To this end, we first prefetch all micro-batches for the next optimizer step, enabling each process to compute the local token counts; a subsequent \textit{All-Reduce} collective operation then aggregates these local token counts into the final global token count, \textit{i.e.}, $T_{global}$. Similar to \cite{liao2025mogao}, we then employ \textit{Global Batch Reduced Loss} by dividing each micro-batch loss by the averaged token count per process, $\tfrac{T_{global}}{N}$, which can be shown that the final synchronized gradient for the subsequent optimizer step is mathematically equivalent to the gradient of the global batch loss, enabling training with unbiased gradients. 

\textbf{Classifier-free guidance (CFG).}
 We follow previous works~\citep{chen2025sharegpt,BAGEL-7B_arxiv_2025} to use CFG for enhanced visual generation quality. For training, we randomly drop conditional text and reference image tokens.
For inference, we combines conditional and unconditional predicted logits. For more details, please refer to Appendix~\ref{sec-cfg}.

\begin{table*}[t]
\centering
\caption{Performance comparison across multiple multimodal understanding benchmarks. 
Higher scores are better, as indicated by the up-arrow ($\uparrow$). 
\textbf{A-LLM} denotes the number of activated LLM parameters, while \textbf{Vis.} indicates the parameter count of the vision encoder or tokenizer for multimodal understanding. Chameleon~\citep{Chameleon-7B_arxiv_2024} does not report the parameter count of its vision tokenizer. slash (/) denotes that models do not require a vision encoder or tokenizer for multimodal understanding. Top-1 accuracy is reported (Best in \textbf{bold}, second best is \underline{underlined}).}
\label{tab:multi-modal-understanding-general}
\resizebox{\textwidth}{!}{%
    \begin{tabular}{l c c cccccccc}
    \toprule
    \multirow{2}{*}{\textbf{Model}} & \multicolumn{2}{c}{\textbf{\# Params}} & \multirow{2}{*}{\textbf{MME-P$\uparrow$}} & \multirow{2}{*}{\textbf{MME-S$\uparrow$}} & \multirow{2}{*}{\textbf{MMBench$\uparrow$}} & \multirow{2}{*}{\textbf{MMMU$\uparrow$}} & \multirow{2}{*}{\textbf{MM-Vet$\uparrow$}} & \multirow{2}{*}{\textbf{MathVista$\uparrow$}} & \multirow{2}{*}{\textbf{SEED$\uparrow$}} \\
    \cmidrule(lr){2-3}
    & \textbf{A-LLM} & \textbf{Vis.} & & & & & & & \\
    \midrule
    \multicolumn{10}{l}{\textit{Encoder-based Understanding Only Models}} \\
    \addlinespace 
    InternVL2~\citep{InternVL2_arxiv_2024} & 1.8B & 0.3B & 1440 & 1877 & 73.2 & 34.3 & 44.6 & 46.4 & 71.6 \\
    InternVL2.5~\citep{InternVL2.5_arxiv_2024} & 1.8B & 0.3B & - & 2138 & 74.7 & 43.6 & 60.8 & 51.3 & - \\
    Qwen2-VL~\citep{Qwen2-VL_arxiv_2024} & 1.5B & 0.6B & - & 1872 & 74.9 & 41.1 & 49.5 & 43.0 & - \\
    Qwen2.5-VL~\citep{Qwen2.5_arxiv_2025} & 3B & 0.6B & - & 2157 & 79.1 & 53.1 & 61.8 & 62.3 & - \\
    \midrule
    \multicolumn{10}{l}{\textit{Encoder-free Understanding Only Models}} \\
    \addlinespace

    Mono-InternVL~\citep{Mono-InternVL_CVPR_2025} & 1.8B & / &  - & 1875 & 65.5 &33.7 &40.1 & 45.7 & 67.4 \\
    EvE~\citep{EvE_nips_2024} & 7B & / &  - & 1628 & 52.3 &32.6 &25.7 & - & 64.6 \\
    EvEv2.0~\citep{EvEv2.0_arxiv_2025} & 7B & / &  - & 1709 & 66.3 &39.3 &45.0 & - & 71.4 \\
    VoRA~\citep{vora_2025_arxiv} &7B &/& 1363&1674&64.2&32.2&33.7&-&64.2\\
    SAIL~\citep{SAIL_2025_ICCV}&7B &/ &-&1719&70.1&-&46.3&57.0&72.9\\
    HoVLE~\citep{HoVLE_CVPR_2025} & 2.6B & / &  - & 1864 & 71.9 &33.7 &44.3 & 46.2 & 70.7 \\
    \midrule
    \multicolumn{10}{l}{\textit{Unified Models}} \\
    \addlinespace

    Chameleon~\citep{Chameleon-7B_arxiv_2024} & 7B & - & - & - & 35.7 & 28.4 & 8.3 & - & 30.6 \\
    Emu3~\citep{Emu3-Gen_arxiv_2024} & 8B & 0.3B &- &-&58.5&31.6&37.2&-&68.2\\
    Harmon~\citep{Harmon-1.5B_arxiv_2025}&1.5B& 0.9B & 1155&1476&65.5&38.9&-&-&67.1\\
    Show-o2~\citep{show-o2_arxiv_2025} &1.5B& 0.5B & 1450 & - & 67.4 &37.1 & -&-&65.6\\
    Janus-Pro~\citep{Janus-pro_arxiv_2025} & 1.5B & 0.3B & 1444 & - & 75.5 & 36.3 & 39.8 & - & - \\
     \rowcolor{tablerowcolor}
    OneCAT-1.5B &  1.5B & / & 1509& 1893& 72.4 & 39.0 & 42.4&55.6&70.9\\ 
    ILLUME+~\citep{ILLUME+3B_arxiv_2025} & 3B & 0.6B & 1414 & - &\textbf{80.8} & \underline{44.3} & 40.3 & - & \textbf{73.3} \\
    VILA-U~\citep{VILA-U_arxiv_2024} & 7B & 0.4B& 1401 & - & - & - & 33.5 & - & 59.0 \\
    Janus-Pro~\citep{Janus-pro_arxiv_2025} & 7B & 0.3B & 1567 & - & 79.2 & 41.0 & \underline{50.0} & - & - \\
    Tar~\citep{Tar-7B_arxiv_2025} & 7B & 0.4B & 1571 &\underline{1926} &74.4 & 39.0 &- & -& - \\
    Show-o2~\citep{show-o2_arxiv_2025} &7B& 0.5B & \underline{1620} & - & \underline{79.3} &\textbf{48.9} & -&-&69.8\\
    \rowcolor{tablerowcolor}
    OneCAT-3B &  3B & / & \textbf{1630}& \textbf{2051}& 78.8 & 41.9 & \textbf{52.2}&\textbf{61.7}&\underline{72.5} \\
    \bottomrule
    \end{tabular}%
}
\end{table*}

\begin{figure}[!t]
\centering
\includegraphics[width=1\linewidth]{figures/draw_t2i.pdf}
\caption{Text-to-Image comparison.}
\label{draw_t2i}
\end{figure}

\begin{figure}[!t]
\centering
\includegraphics[width=1\linewidth]{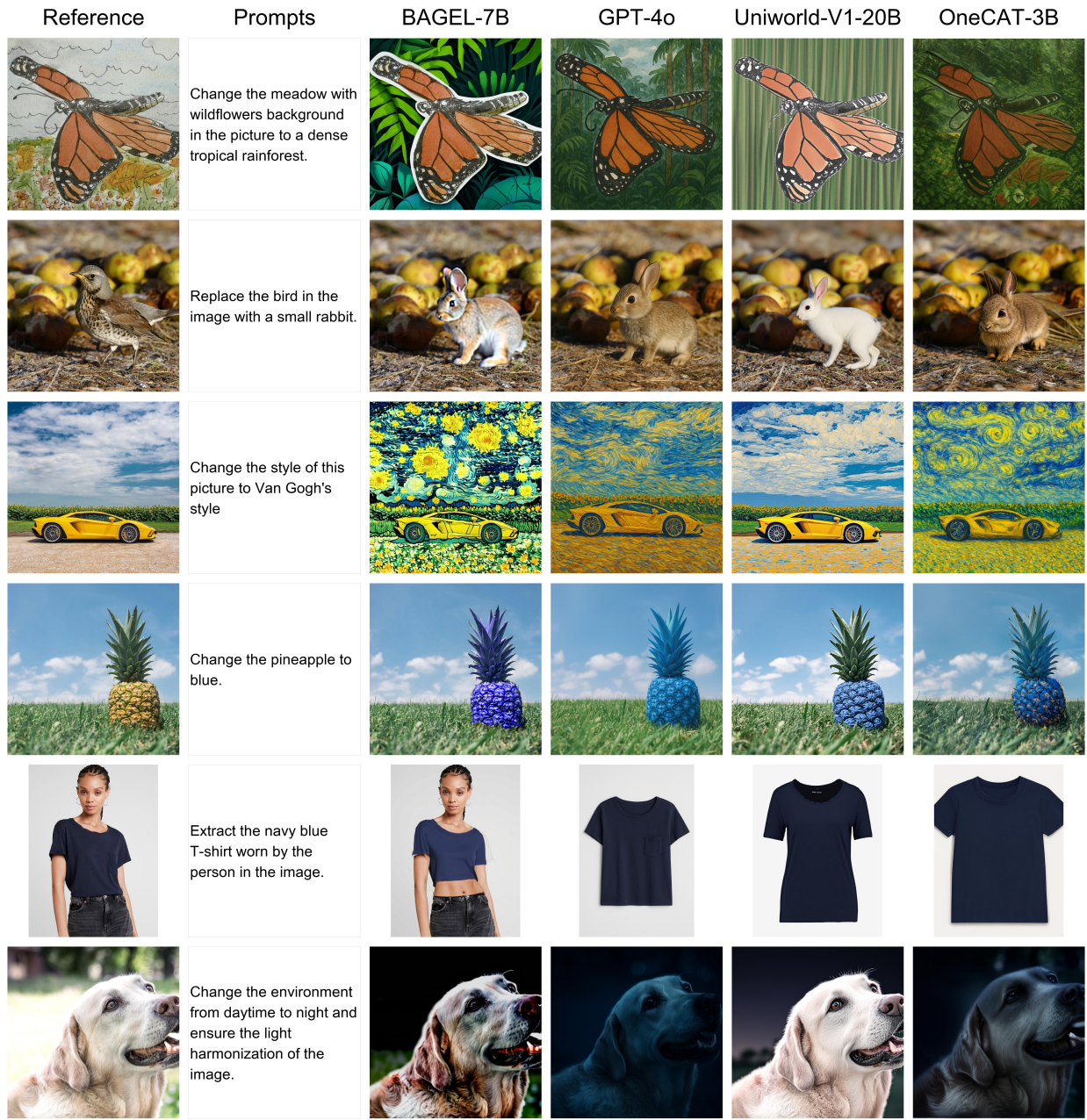}
\caption{Image-Editing comparison.}
\label{draw_edit}
\end{figure}

\section{Evaluation}
\subsection{Multimodal understanding.}

We evaluate OneCAT on several multimodal understanding benchmarks: MMbench~\citep{benchmark_mmbench_ECCV_2024}, MME~\citep{benchmark_MME_ECCV_2024}, MMMU~\citep{benchmark_MMMU_CVPR_2024}, MM-Vet~\citep{benchmark_MM-vet_arxiv_2023}, and SEED~\citep{benchmark_SEED_arxiv_2023} assess  general multimodal perception and reasoning. MathVista~\citep{benchmark_mathvista_arxiv_2023} focuses on mathematical reasoning. TextVQA~\citep{benchmark_TextVQA_CVPR_2019}, ChartQA~\citep{benchmark_ChartQA_arxiv_2022}, InfoVQA~\citep{benchmark_infoVQA_WACV_2022}, and DocVQA~\citep{benchmark_DocVQA_WACV_2021} evaluate OCR and text-related visual question answering; while GQA~\citep{benchmark_GQA_CVPR_2019} evaluates visual scene understanding and AI2D~\citep{benchmark_AI2D_ECCV_2016} evaluates scientific diagram comprehension.

\begin{table*}[t]
\centering
\caption{Performance comparison on the GenEval~\citep{Benchmark_geneval_nips_2023} benchmark. The dagger ($^{\dagger}$) indicates methods that employ an LLM for prompt rewriting. Best in \textbf{bold}, second best is \underline{underlined}.}
\label{tab:gen_model_comparison}

\small
\setlength{\tabcolsep}{5pt}

\resizebox{0.85\textwidth}{!}{
\begin{tabular}{lccccccc}
\toprule
\textbf{Model} & \textbf{Single Obj.} & \textbf{Two Obj.} & \textbf{Counting} & \textbf{Colors} & \textbf{Position} & \textbf{Color Attri.} & \textbf{Overall$\uparrow$} \\

\midrule
\multicolumn{8}{l}{\textit{Generation-only Models}} \\
    \addlinespace
SDXL~\citep{sdxl_arxiv_2023} & 0.98 & 0.74 & 0.39 & 0.85 & 0.15 & 0.23 & 0.55 \\
DALL-E 3~\citep{DALL-E_3_arxiv_2020} & 0.96 & 0.87 & 0.47 & 0.83 & 0.43 & 0.45 & 0.67 \\
Infinity$^{\dagger}$~\citep{infinity_CVPR_2025} & -&0.85&-&-&0.49&0.57&0.73\\
SD3-Medium~\citep{SD3-Medium_ICML_2024} & 0.99 & 0.94 & 0.72 & 0.89 & 0.33 & 0.60 & 0.74 \\
FLUX.1-dev$^{\dagger}$~\citep{FLUX.1-dev_github_2024} & 0.98 & 0.93 & 0.75 & 0.93 & 0.68 & 0.65 & 0.82 \\
\midrule

\multicolumn{8}{l}{\textit{Unified Models}} \\
    \addlinespace
Chameleon-7B~\citep{Chameleon-7B_arxiv_2024} & - & - & - & - & - & - & 0.39 \\
Transfusion-7B~\citep{Transfusion-7B_arxiv_2024} & - & - & - & - & - & - & 0.63 \\
Emu3-8B$^{\dagger}$~\citep{Emu3-Gen_arxiv_2024} & 0.99 & 0.81 & 0.42 & 0.80 & 0.49 & 0.45 & 0.66 \\
ILLUME+ 3B~\citep{ILLUME+3B_arxiv_2025} & 0.99 & 0.88 & 0.62 & 0.84 & 0.42 & 0.53 & 0.72 \\
Harmon-1.5B~\citep{Harmon-1.5B_arxiv_2025} 
&0.99&0.86&0.66&0.85&0.74&0.48&0.76\\
Show-o2-7B~\citep{show-o2_arxiv_2025} & 1.00 & 0.87 & 0.58 & 0.92 & 0.52 & 0.62 & 0.76 \\
Janus-Pro-7B~\citep{Janus-pro_arxiv_2025} & 0.99 & 0.89 & 0.59 & 0.90 & 0.79 & 0.66 & 0.80 \\
  Mogao-7B$^{\dagger}$~\citep{Mogao_arxiv_2025}       & \textbf{1.00} & \textbf{0.97} & 0.83 & 0.93 &\textbf{0.84}& \textbf{0.80}& \underline{0.89} \\
BLIP3-o-8B$^{\dagger}$~\citep{BLIP3-O_arxiv_2025} & -&-&-&-&-&-&0.84 \\
Tar-7B~\citep{Tar-7B_arxiv_2025} & 0.99&0.92&0.83&0.85&0.80&0.65&0.84\\
   UniWorld-V1-20B~\citep{UniWorld-V1_arxiv_2025}  & 0.99&0.93&0.79&0.89&0.49&0.70&0.80 \\
   UniWorld-V1-20B$^{\dagger}$~\citep{UniWorld-V1_arxiv_2025}  & 0.98&0.93&0.81&0.89&0.74&0.71&0.84 \\
BAGEL-7B~\citep{BAGEL-7B_arxiv_2025} & 0.99 & 0.94 & 0.81 & 0.88 & 0.64 & 0.63 & 0.82 \\
BAGEL-7B$^{\dagger}$~\citep{BAGEL-7B_arxiv_2025}  & 0.98 & 0.95 & \textbf{0.84} & \textbf{0.95} & 0.78 & 0.77 & 0.88 \\
   \rowcolor{tablerowcolor}
\textbf{OneCAT-1.5B}&0.99&0.92&0.83&0.91&0.72&0.75&0.85\\
   \rowcolor{tablerowcolor}
\textbf{OneCAT-3B}&\textbf{1.00}&\underline{0.96}&\textbf{0.84}&\underline{0.94}&\textbf{0.84}&\textbf{0.80}&\textbf{0.90}\\
\hline
\end{tabular}}
\end{table*}

As shown in Tab.~\ref{tab:multi-modal-understanding-vqa} and Tab.~\ref{tab:multi-modal-understanding-general}, we compare OneCAT with three types of models: encoder-based understanding-only models, encoder-free understanding-only models, and unified MLLMs. Our OneCAT-3B model demonstrates superior performance, significantly outperforming all existing encoder-free understanding-only MLLMs, e.g., HoVLE~\citep{HoVLE_CVPR_2025} and EvEv2~\citep{EvEv2.0_arxiv_2025}, across nearly all benchmarks. For instance, on OCR-related tasks including AI2D (77.8), ChartQA (81.2), InfoVQA (64.8), and DocVQA (91.2), OneCAT-3B achieves new state-of-the-art results among encoder-free models. It also excels in general vision-language benchmarks such as MME-P (1630), MMBench (78.8), and MM-Vet (52.2).

Moreover, OneCAT-3B outperforms recent unified MLLMs that rely on external vision encoders or tokenizers—such as Janus-Pro-7B~\citep{Janus-pro_arxiv_2025} (using SigLIP~\citep{MLLMs_2023_arxiv_siglip}) and Tar-7B~\citep{Tar-7B_arxiv_2025} (using SigLip2~\citep{MLLMs_2025_arxiv_siglip2})—despite activating fewer parameters.  Compared to top-tier encoder-based understanding-only models like Qwen2.5-VL-3B~\citep{Qwen2.5_arxiv_2025}, our model exhibits a slight performance gap, which we primarily attribute to differences in the scale and quality of training data. Specifically, Qwen2.5-VL was trained on 4T tokens, whereas our OneCAT was trained on only 0.5T tokens for multimodal understanding. We believe this gap can be bridged in the future by scaling up the pretraining data and incorporating more higher-quality instruction data.

\subsection{Visual Generation.}

We evaluate our model on three widely-used visual generation benchmarks: two for text-to-image generation, GenEval~\citep{Benchmark_geneval_nips_2023} and DPG-Bench~\citep{Benchmark_DPG-Bench_arxiv_2024}, and one for  instruction-based image editing, ImgEdit~\citep{Benchmark_imgedit_arxiv_2025}. We follow previous works~\citep{chen2025sharegpt,Mogao_arxiv_2025,BAGEL-7B_arxiv_2025} to use Classifier-free guidance(CFG)~\citep{ho2022classifier} to enhance visual generation quality.  During training, we randomly drop tokens of conditional text and reference image with probabilities 0.1, 0.1, respectively.
During inference, we combines conditional and unconditional predicted logits to produce outputs that better adhere to the given conditions. To ensure a fair comparison, we adhere strictly to the original raw prompts for the GenEval benchmark, unlike some previous works that employ LLM-based prompt rewriting to enhance performance.
As shown in  Tab.\ref{tab:gen_model_comparison}, \ref{tab:dpg_bench_eval_vertical_resized}, and~\ref{tab:imgedit_bench}, our OneCAT-3B model demonstrates highly competitive performance across all tasks.

On GenEval (Tab.~\ref{tab:gen_model_comparison}), which evaluates fine-grained instruction following over object counts, colors, and spatial relationships, OneCAT-3B achieves a SOTA overall score of 0.90. This performance surpasses most all unified models, including the strong baseline BAGEL-7B (0.88 with prompt rewriting) and Mogao-7B (0.89 with prompt rewriting). Notably, OneCAT-3B excels in challenging categories such as Position and Color Attribute, where it achieves the best performance (0.84 and 0.80 respectively), showcasing its superior ability to interpret complex spatial and attribute-based instructions.

On DPG-Bench (Tab.~\ref{tab:dpg_bench_eval_vertical_resized}), a benchmark focused on compositional text-to-image generation, OneCAT-3B attains a strong overall score of 84.53. This result is highly competitive among unified models, outperforming strong counterparts like Janus-Pro-7B (84.19) and Mogao-7B (84.33). 

On ImgEdit-Bench (Tab.~\ref{tab:imgedit_bench}), a challenging and diverse image editing benchmark, OneCAT-3B achieves an overall score of 3.43. This places it firmly among the top-performing unified models and significantly outperforms many specialized editing models. OneCAT-3B demonstrates exceptional capabilities in categories requiring precise local and global adjustments, securing the top scores in Adjust (3.70), Extract (2.42), and Background (3.79) manipulation. This highlights the effectiveness of our model's ability to condition its generation on fine-grained visual cues from a reference image.

\subsection{Qualitative results.}
We present qualitative comparisons for the text-to-image and image-editing tasks in Fig.~\ref{draw_t2i} and \ref{draw_edit}, respectively. OneCAT-3B exhibits leading instruction-following and world-understanding capabilities. We also present more visual generation and understanding showcases in Appendix~\ref{sec-appendix-visualization}.

\begin{table*}[t!]
\centering
\caption{
Performance comparison on the DPG-Bench~\citep{Benchmark_DPG-Bench_arxiv_2024} benchmark. Best in \textbf{bold}, second best is \underline{underlined}.
}
\label{tab:dpg_bench_eval_vertical_resized}
\resizebox{0.7\columnwidth}{!}{%
    \begin{tabular}{l ccccc c}
    \toprule
    \textbf{Model} & \textbf{Global} & \textbf{Entity} & \textbf{Attribute} & \textbf{Relation} & \textbf{Other} & \textbf{Overall$\uparrow$} \\
    \midrule
    \multicolumn{7}{l}{\textit{Generation-only Models}} \\
    \addlinespace 
        Hunyuan-DiT~\citep{Hunyuan-DiT_arxiv_2024}      & 84.59 & 80.59 & 88.01 & 74.36 & 86.41 & 78.87 \\
        Playground v2.5~\citep{Playgroundv2.5_arxiv_2024}  & 83.06 & 82.59 & 81.20 & 84.08 & 83.50 & 75.47 \\
        PixArt-$\Sigma$~\citep{Pixart_ECCV_2024}  & 86.89 & 82.89 & 88.94 & 86.59 & 87.68 & 80.54 \\
        DALL-E 3~\citep{DALL-E_3_arxiv_2020}         & 90.97 & 89.61 & 88.39 & 90.58 & 89.83 & 83.50 \\
        Infinity~\citep{infinity_CVPR_2025} &93.11&-&-&90.76&-&83.46\\
        SD3-Medium~\citep{SD3-Medium_ICML_2024}       & 87.90 & 91.01 & 88.83 & 80.70 & 88.68 & 84.08 \\
        FLUX.1-dev~\citep{FLUX.1-dev_github_2024}&82.10&89.50&88.80&91.10&89.40&84.00 \\
    \midrule
    \multicolumn{7}{l}{\textit{Unified Models}} \\
    \addlinespace
        Emu3-8B~\citep{Emu3-Gen_arxiv_2024}   & -     & -     & -     & -     & -     & 81.60 \\
        Janus-Pro-7B~\citep{Janus-pro_arxiv_2025}   & 86.90 & 88.90 & \underline{89.40} & 89.32 & 89.48 & 84.19 \\
        Mogao-7B~\citep{Mogao_arxiv_2025}       & 82.37 & 90.03 & 88.26 & \underline{93.18} & 85.40 & 84.33 \\
        BLIP3-o-8B~\citep{BLIP3-O_arxiv_2025} &-&-&-&-&-&81.60 \\
        Tar-7B~\citep{Tar-7B_arxiv_2025} &83.98 &88.62&88.05&\textbf{93.98}&84.86&84.19\\
        Show-o2-7B~\citep{show-o2_arxiv_2025}   & \underline{89.00} & \textbf{91.78} & \textbf{89.96} & 91.81 & \textbf{91.64} & \textbf{86.14} \\
   \rowcolor{tablerowcolor}
         \textbf{OneCAT-1.5B}   & \textbf{90.48} & 86.70 & 86.75 & 89.32 & 84.93 & 81.72 \\
   \rowcolor{tablerowcolor}
        \textbf{OneCAT-3B}   & 85.46 & \underline{90.81} & 89.00 & 90.40 & \underline{89.56} & \underline{84.53} \\
    \bottomrule
    \end{tabular}%
}
\end{table*}

\begin{table*}[t!]
\centering
\caption{Comprehensive comparison on ImgEdit-Bench~\citep{Benchmark_imgedit_arxiv_2025} showing performance across nine editing categories. Higher scores are better for all metrics. Best in \textbf{bold}, second best is \underline{underlined}.}
\label{tab:imgedit_bench}
\resizebox{\textwidth}{!}{%
    \begin{tabular}{l cccccccccc}
    \toprule
    \textbf{Model} & \textbf{Add} & \textbf{Adjust} & \textbf{Extract} & \textbf{Replace} & \textbf{Remove} & \textbf{Background} & \textbf{Style} & \textbf{Hybrid} & \textbf{Action} & \textbf{Overall} \\
    \midrule
    \multicolumn{11}{l}{\textit{Editing-only Models}} \\
    MagicBrush~\citep{Magicbrush_2023_nips}      & 2.84 & 1.58 & 1.51 & 1.97 & 1.58 & 1.75 & 2.38 & 1.62 & 1.22 & 1.90 \\
    Instruct-Pix2Pix~\citep{Instruct-Pix2Pix_2023_CVPR} & 2.45 & 1.83 & 1.44 & 2.01 & 1.50 & 1.44 & 3.55 & 1.20 & 1.46 & 1.88 \\
    AnyEdit~\citep{Anyedit_arxiv_2025}        & 3.18 & 2.95 & 1.88 & 2.47 & 2.23 & 2.24 & 2.85 & 1.56 & 2.65 & 2.45 \\
    UltraEdit~\citep{UltraEdit_2024_nips}       & 3.44 & 2.81 & 2.13 & 2.96 & 1.45 & 2.83 & 3.76 & 1.91 & 2.98 & 2.70 \\
    Step1X-Edit~\citep{Step1X-Edit_arxiv_2025}     & 3.88 & 3.14 & 1.76 & 3.40 & 2.41 & 3.16 & 4.63 & 2.64 & 2.52 & 3.06 \\
    ICEdit~\citep{ICEdit_arxiv_2025}          & 3.58 & 3.39 & 1.73 & 3.15 & 2.93 & 3.08 & 3.84 & 2.04 & 3.68 & 3.05 \\
    \midrule
    \multicolumn{11}{l}{\textit{Unified Models}} \\
    OmniGen~\citep{OmniGen_CVPR_2025}         & 3.47 & 3.04 & 1.71 & 2.94 & 2.43 & 3.21 & 4.19 & 2.24 & 3.38 & 2.96 \\
    OmniGen2~\citep{OmniGen2_arxiv_2025}        & 3.57 & 3.06 & 1.77 & \underline{3.74} & \underline{3.20} & \underline{3.57} & \textbf{4.81} & \underline{2.52} & \textbf{4.68} & \textbf{3.44} \\
    BAGEL-7B~\citep{BAGEL-7B_arxiv_2025}            & 3.56 & 3.31 & 1.70 & 3.30 & 2.62 & 3.24 & 4.49 & 2.38 & \underline{4.17} & 3.20 \\
    UniWorld-V1-20B~\citep{UniWorld-V1_arxiv_2025}     & \textbf{3.82} & \underline{3.64} & \underline{2.27} & 3.47 & \textbf{3.24} & 2.99 & 4.21 & \textbf{2.96} & 2.74 & 3.26 \\
    \midrule
    OneCAT-3B & \underline{3.65}&\textbf{3.70}&\textbf{2.42}&\textbf{3.92}&3.00&\textbf{3.79}&\underline{4.61}&2.23&3.53&\underline{3.43}\\
    \bottomrule
    \end{tabular}}
\end{table*}

\subsection{Comparison of Inference Efficiency}
In addition to its strong performance, our model's architectural design also yields significant improvements in inference efficiency for handling both high-resolution image input and output. 

The left section of Tab.~\ref{tab:combined_speed} compares the inference efficiency of OneCAT and Qwen2.5-VL for multimodal understanding. 
Benefiting from our pure decoder-only architecture, which removes the external ViT encoder, OneCAT achieves significantly faster prefilling, reducing first-token latency by up to 61\% compared to Qwen2.5-VL on high-resolution inputs.
As shown in the right section of Tab.~\ref{tab:combined_speed},  OneCAT exhibits a  substantial speed advantage over the diffusion-based BAGEL in image generation.  When producing a $1024 \times 1024$ image, OneCAT requires only 2.85s for T2I generation and 4.61s for image editing—approximately $10\times$ faster than BAGEL.
This improvement stems from two key elements: our multi-scale autoregressive mechanism within the LLM, and a VAE tokenizer-free design for editing that reduces both  encoding and decoding overhead.

\begin{table}[!t]
\centering
\caption{Efficiency comparison for understanding (\textbf{Left}) and generation (\textbf{Right}), tested on one NVIDIA H800. \textbf{Left:}  We report the time to first token (TTFT). The number of input text tokens are fixed to 24. 256* is the number of visual tokens of thumbnail. \textbf{Right:} We report total inference time for Text-to-Image (T2I) and Image-Editing.}
    \setlength{\belowcaptionskip}{-10pt}
\label{tab:combined_speed}

\begin{subtable}{0.51\linewidth}
    \centering
    \small
    \renewcommand{\arraystretch}{1.2}
    \label{tab:speed}
    \resizebox{\linewidth}{!}{
    \begin{tabular}{lcll}
    \toprule
    \textbf{Model} & \makecell{\textbf{Resolution of}\\\textbf{Input Image}} & \makecell{\textbf{\#Tokens}\\\textbf{Visual}} & \textbf{TTFT(s)} \\
    \hline
    Qwen2.5-VL-3B & $768 \times 768$ & 731 & 0.135  \\
    OneCAT-3B     & $768 \times 768$ & 731 +$256^{*}$& 0.067 \textcolor{blue!75}{(50\%$\downarrow$)}  \\
    \hline
    Qwen2.5-VL-3B & $1792 \times 1792$ & 4098 & 0.583  \\
    OneCAT-3B     & $1792 \times 1792$ & 4098 +$256^{*}$ & 0.225 \textcolor{blue!75}{(61\%$\downarrow$)}   \\
    \bottomrule
    \end{tabular}
    }
\end{subtable}
\hfill 
\begin{subtable}{0.48\linewidth}
    \centering
    \small
    \renewcommand{\arraystretch}{1.2}
    \label{tab:speed-t2i}
    \resizebox{\linewidth}{!}{
    \begin{tabular}{lcll}
    \toprule
    \textbf{Model} & \makecell{\textbf{Resolution of}\\\textbf{Output Image}} & \makecell{\textbf{T2I Infer.}\\\textbf{Time (s)}} & \makecell{\textbf{Edit Infer.}\\\textbf{Time (s)}} \\
    \hline
    BAGEL-7B  & $512 \times 512$ & 8.76 & 13.45 \\
    OneCAT-3B & $512 \times 512$ & 1.40 \textcolor{blue!75}{(84\%$\downarrow$)} & 2.03 \textcolor{blue!75}{(84\%$\downarrow$)} \\
    \hline
    BAGEL-7B  & $1024 \times 1024$ & 26.29 & 46.44 \\
    OneCAT-3B & $1024 \times 1024$ & 2.85 \textcolor{blue!75}{(89\%$\downarrow$)} & 4.61 \textcolor{blue!75}{(90\%$\downarrow$)} \\
    \bottomrule
    \end{tabular}
    }
\end{subtable}

\end{table}

\subsection{Ablation Study}
\subsubsection{Ablation Study on  Understanding  Distillation (\textit{Und. Distil.}) Stage}

We ablate each module in the \textit{Und.~Distil.} stage using the OneCAT-1.5B.  Each experiment uses 8B sampled tokens ($\sim$10M image-text pairs from Stage-1 dataset) to optimize the \texttt{Und.~FFN}, followed by a simplified SFT using LLaVA-665k~\citep{VLLMs_MLP_Llava_2023_nips} dataset.

\begin{figure}[htbp]
    \centering
    \begin{subfigure}{0.45\textwidth}\includegraphics[width=\textwidth]{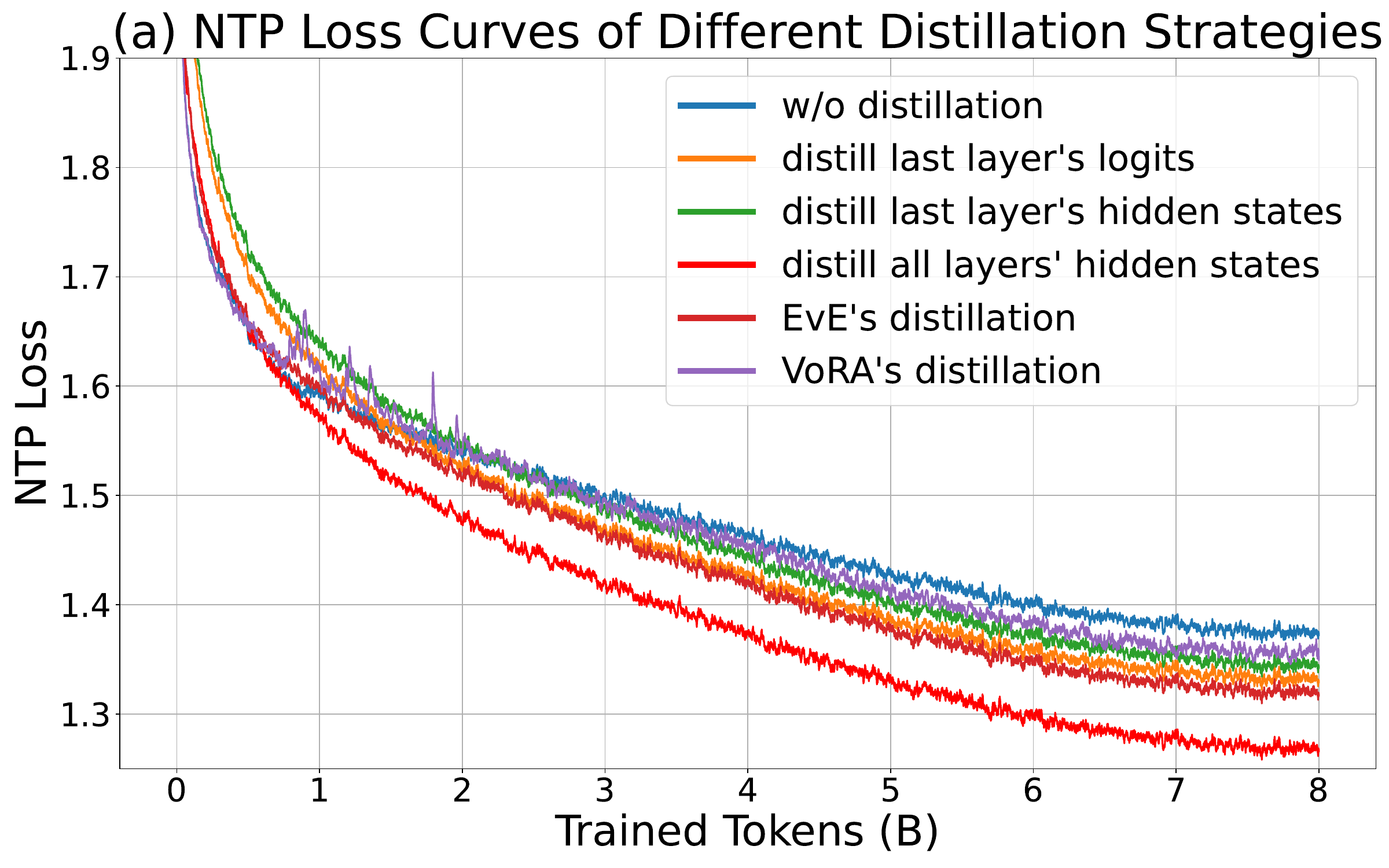}
        \label{fig:abla1}
    \end{subfigure}
    \begin{subfigure}{0.45\textwidth}
 \includegraphics[width=\textwidth]{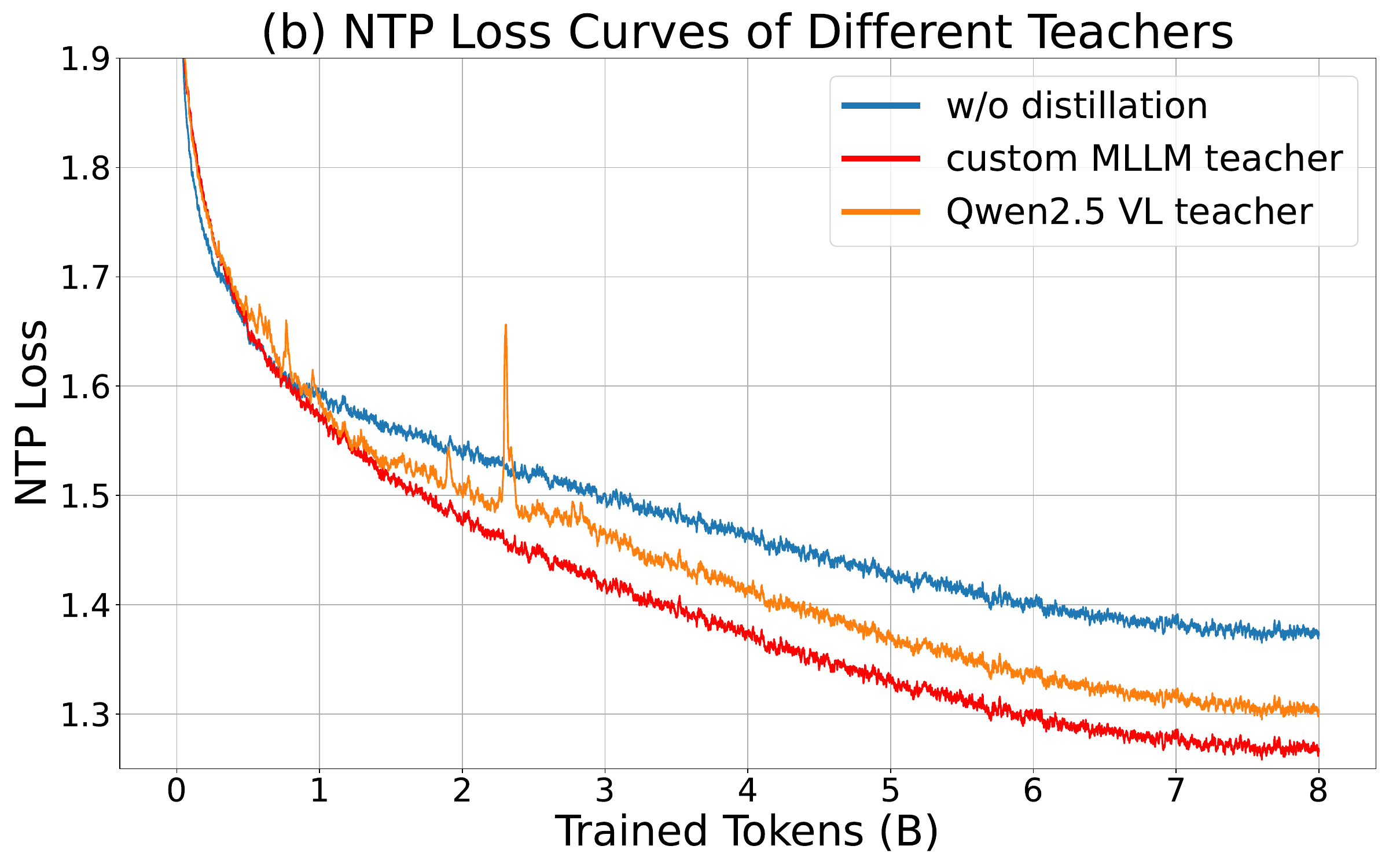}
        \label{fig:abla2}
    \end{subfigure}
    \caption{Comparision of different distillation strategies and teachers for stage-1 training.}
    \label{fig:abla}
\end{figure}

\textbf{On Different Distillation Strategies.}
 As shown in Tab.~\ref{table:effect-distillation} and Fig.~\ref{fig:abla}(a), our proposed distillation approach—distilling all layers' hidden states—significantly improves visual learning efficiency and multimodal performance. In addition, distilling only the last layer's hidden states or logits also brings gains over the baseline, yet remains inferior to our full distillation approach.  This highlights the necessity of full-layer distillation for effective visual knowledge transfer from the custom teacher MLLM.  Notably, our strategy yields better performance than applying the distillation approaches of EvE~\citep{EvE_nips_2024} and VoRA~\citep{vora_2025_arxiv} to our OneCAT-1.5B under the same setting, further validating the efficacy of our proposed method.
\begin{table}[htbp]
    \centering
            \small
            \centering
            \resizebox{0.95\textwidth}{!}{
                \begin{tabular}{c cccccccc cc}
                    \toprule
                    Methods & MMB & MME-S & MMVet & SEED & AI2D & ChartQA & TextVQA & \textbf{Avg.} \\
                    \midrule
                    w/o distillation & 42.0 & 1209 & 13.2 & 47.0 & 53.8 & 10.5 & 10.0 & 31.4 \\
                    distill last layer's logits & 44.8 & 1312 & 15.4 & 53.0 & 55.2 & 11.5 & 10.6 & 33.9 \\
                    distill last layer's hidden states & 46.0 & 1255 & 15.3 & 54.3 & 55.3 & 11.0 & 10.9 & 33.9 \\
                    \rowcolor{tablerowcolor}
                    distill all layers' hidden states & 49.4 & 1327 & 15.5 & 56.3 & 54.4 & 11.9 & 11.3 & \textbf{35.3} \\
                    EvE~\citep{EvE_nips_2024} & 44.5 & 1276 & 15.0 & 52.8 & 54.0 & 10.6 & 11.2 & 33.4 \\
                    VoRA~\citep{vora_2025_arxiv} & 41.5 &1234 &15.1 &50.6 &54.2 &10.4 &10.5 & 32.3\\
                    \bottomrule
                \end{tabular}
            }
             \caption{Effect of different distillation strategies.}  
            \label{table:effect-distillation}
\end{table}

    \begin{wraptable}[6]{r}{0.35\columnwidth}
    \vspace{-15pt}
        \begin{tabular}{c c}
            \toprule
            Methods & \textbf{Avg.} \\
            \midrule
            w/o distillation &  31.4\\
            \rowcolor{tablerowcolor}
            distill with our custom teacher &\textbf{35.3}  \\
            distill with Qwen2.5-VL teacher &33.7  \\
            \bottomrule
        \end{tabular}
         \caption{Effect of different teachers. \textbf{Avg.} denotes the average score in Tab.~\ref{table:effect-distillation}.}
    \label{table:abla-teacher}
\end{wraptable}

\textbf{On Different Teachers.}
We then analyze the effect of different teacher models. As shown in Tab.~\ref{table:abla-teacher} and Fig.~\ref{fig:abla}(b), using the pretrained Qwen2.5-VL-1.5B as the teacher results in unstable training and lower distillation efficiency due to parameter misalignment between the teacher and student LLMs—especially for the frozen attention and QKV layers—highlighting the importance of our custom teacher design.

\subsubsection{Ablation Study on Unified Mid-Training (Stage-2)}
\label{sec-abla-stage-2}

We initialize the model with OneCAT-1.5B from Stage-1 and perform a simplified unified mid-training to study (i) the effect of token ratio across tasks and (ii) the effect of SAA.
All models then undergo a simplified SFT  using LLaVA-665k~\citep{VLLMs_MLP_Llava_2023_nips} for understanding and BLIP3o-60k~\citep{BLIP3-O_arxiv_2025} for generation with a 1:1 token ratio.

\textbf{On Token Ratio.} We sample 5B and 10B tokens for text-only (T) and multimodal understanding (U), and vary the number of trained tokens in visual generation (G) to study the impact of token ratio. As shown in Table~\ref{table:effect-token-ratio-final}, increasing the training tokens for G does not significantly affect U but improves G. To balance training cost and performance, we adopt a token ratio of 1:2:6 for Stage-2. 
\begin{table}[htbp]
    \centering
        \resizebox{\textwidth}{!}{
            \begin{tabular}{ccccccccc|cc}
                \toprule
                \textbf{Token Ratio} & \multicolumn{8}{c}{\textbf{Understanding}} & \multicolumn{2}{c}{\textbf{Generation}} \\
                \cmidrule(lr){2-9} \cmidrule(lr){10-11}
                \textbf{(T:U:G)} & MMB & MME-S & MMVet & SEED & AI2D & ChartQA & TextVQA & \textbf{Avg.} & GenEval & DPG \\
                \midrule
                5B : 10B : 0B & 61.6 & 1556 & 28.9 & 67.2 & 60.1 & 56.1 & 56.2 &55.1& - & - \\
                5B : 10B : 10B & 62.7 & 1547 & 29.7 & 66.7 & 60.3 & 59.8 & 55.4 & 55.6& 80.7 & 74.6 \\
                \rowcolor{tablerowcolor}
                5B : 10B : 30B & 62.0 & 1602 & 29.4 & 66.7 & 59.8 & 61.6 & 55.5 &56.0& 81.2 & 74.9 \\
                5B : 10B : 45B & 62.1&1544 & 30.1&  66.8&58.4 &58.6 & 55.0& 55.2&81.4&75.6\\
                \bottomrule
            \end{tabular}
        }
            \caption{Effect of trained token ratio across text-only (T), understanding (U), and generation (G).}
        \label{table:effect-token-ratio-final}
    \end{table}%

    \begin{wraptable}[6]{r}{0.3\columnwidth}
    \centering 
    \vspace{-10pt}
    \begin{tabular}{l cc} 
        \toprule
        & GenEval & DPG \\
        \midrule
        \rowcolor{tablerowcolor}
        w/ SAA & 81.2 & 74.9 \\
        w/o SAA & 78.1 & 74.0 \\
        \bottomrule
    \end{tabular}
    
    \caption{Effect of SAA.}
    \label{table:effect-saa}
\end{wraptable}

    \textbf{On SAA.} We then remove the SAA during the stage-2 training for performance comparison, as shown in Table~\ref{table:effect-saa}. Each model train 5B, 10B, 30B tokens for T, U, and G for Stage-2, respectively. We can see that removing SAA leads to an obvious performance drop. We also present the visualization of frequency properties of tokens from different scales in Appendix~\ref{sec-visual-saa} to better understand the motivation of our SAA.



%% file: sections/conclusion.tex
\section{Conclusion}

In this work, we presented OneCAT, a pure decoder-only unified multimodal model that seamlessly integrates understanding, generation, and editing within a single, streamlined architecture. By eliminating external encoders and tokenizers, employing a modality-specific MoE design, and introducing a multi-scale autoregressive generation mechanism, OneCAT achieves strong performance across a wide range of benchmarks while significantly improving inference efficiency. Our results demonstrate the viability and advantages of a first-principles approach to multimodal modeling, offering a powerful new baseline for future research and applications in general-purpose multimodal intelligence.

%% file: sections/appendix.tex
\appendix

\section{Preliminary of Next-Scale Prediction }
\label{sec-preliminary-VAR}

\subsection{Multiscale Tokenization} 
Leveraging the inherent coarse-to-fine structure of natural images, VAR~\citep{VAR_nips_2024} introduces a multi-scale tokenizer that encodes an image into $K$ token scales $(R_1, R_2, \dots, R_K)$. The resolution of each scale $R_k$, denoted as $(h_k, w_k)$, increases monotonically with the scale index $k$. Specifically, given a feature map $F$ extracted from an image with an image encoder, VAR defines these token scales recursively:
\begin{align}
    R_1 &= \mathcal{Q}(\operatorname{interpolate}_1(F)), \\
    R_2 &= \mathcal{Q}(\operatorname{interpolate}_2(F - \operatorname{interpolate}_K(R_1))), \\
    &\vdots \nonumber \\
    R_k &= \mathcal{Q}\Big(\operatorname{interpolate}_k(F - \sum_{i=1}^{k-1} \operatorname{interpolate}_K(R_i)) \Big), \label{eq:var-tokenization} \\
    &\vdots \nonumber \\
    R_K &= \mathcal{Q}\Big(F - \sum_{i=1}^{K-1} \operatorname{interpolate}_K(R_i)\Big),
\end{align}
where $\operatorname{interpolate}_i$ is an operator that resizes its input to the resolution $(h_i, w_i)$, and $\mathcal{Q}$ is the quantization operator. For a given 3D feature map $x \in \mathbb{R}^{d \times h \times w}$, we implement $\mathcal{Q}$ using Binary Spherical Quantization (BSQ)~\citep{BSQ_2024_arxiv}, following~\citet{infinity_CVPR_2025}. The quantization is applied to each spatial feature vector $x_{ij} \in \mathbb{R}^d$ as:
\begin{equation}    
\mathcal{Q}(x_{ij}) = \frac{1}{\sqrt{d}} \operatorname{sign}\left(\frac{x_{ij}}{\|x_{ij}\|_2}\right).
\end{equation}

\subsection{Visual Auto-Regressive Training}
The premise for generation is that the feature map $F$ can be well approximated by summing all scales upsampled to the final resolution: $F \approx \sum_{i=1}^{K} \operatorname{interpolate}_K(R_i)$. It therefore suffices to generate the sequence of scales $R_{1:K}$ to synthesize an image. To achieve this, VAR models the joint distribution over the scales auto-regressively, factorizing the log-likelihood as:
\begin{equation}
    \log p_{\theta}(R_{1:K}) = \sum_{k=1}^K \log p_{\theta}(R_k \mid R_{1:k-1}). \label{eq:var-likelihood}
\end{equation}
The model, with parameters $\theta$, is trained to maximize this log-likelihood by learning to predict the current scale $R_k$ conditioned on all preceding scales $R_{1:k-1}$. To enable efficient parallel decoding, VAR assumes that all tokens within the current scale $R_k$ are conditionally independent given $R_{1:k-1}$.

However, this conditional independence assumption, coupled with imperfect model training, can lead to error propagation: mistakes in generating early-stage scales ($R_1, \dots, R_{k-1}$) are amplified when generating subsequent, higher-resolution scales $R_k$. To mitigate this issue, \citet{infinity_CVPR_2025} proposed \textit{Bitwise Self-Correction}. This technique involves training the model on corrupted versions of the conditioning scales $R_{1:k-1}$, thereby teaching it to generate the correct $R_k$ even when the preceding scales are imperfect. This robustifies the model against its own generation errors during inference.

\subsection{Predefined Scale Schedules}
We follows \citet{infinity_CVPR_2025}
and \citet{VAR_nips_2024} to establish a set of predefined scale schedules, thus ensuring efficient training across images with varying aspect ratios. As detailed in Table~\ref{tab:scale_schedules}, for each target aspect ratio $r$, we define a specific schedule as a sequence of $K$ resolution tuples: $\{(h_1^r, w_1^r), \dots, (h_K^r, w_K^r)\}$.

These schedules are designed based on two fundamental principles: \textbf{1.Aspect Ratio Consistency:} Each tuple $(h_k^r, w_k^r)$ within a schedule maintains an aspect ratio that is approximately equal to the target ratio $r$, especially at larger scales.
 \textbf{2.Consistent Area Across Scales:} For any given scale level $k$, the image area, calculated as $h_k^r \times w_k^r$, is kept roughly constant across different aspect ratio schedules. This standardization ensures that the training sequence lengths are similar for various aspect ratios, thereby improving overall training efficiency.
 During the inference stage, these predefined schedules enable the model to generate high-quality images covering a wide range of  aspect ratios.

\begin{table}[h!]
    \centering
    \small 
    \caption{Predefined scale schedules $\{(h_1^r, w_1^r), \dots, (h_K^r, w_K^r)\}$ for different aspect ratios. Following \citet{infinity_CVPR_2025}, OneCAT utilizes $K=13$ scales to generate the highest resolution image, such as a $1024 \times 1024$ image for the 1:1 aspect ratio, while lower-resolution images like $512 \times 512$ can be produced by truncating the schedule to $K=10$.}
    
    \label{tab:scale_schedules}
    \resizebox{1\textwidth}{!}{
    \begin{tabular}{c c l}
        \toprule
        \textbf{Aspect Ratio} & \textbf{Resolution} & \textbf{Scale Schedule} \\
        \midrule
        1.000 (1:1)  & $1024 \times 1024$ & (1,1) (2,2) (4,4) (6,6) (8,8) (12,12) (16,16) (20,20) (24,24) (32,32) (40,40) (48,48) (64,64) \\
        0.800 (4:5)  & $896 \times 1120$  & (1,1) (2,2) (3,3) (4,5) (8,10) (12,15) (16,20) (20,25) (24,30) (28,35) (36,45) (44,55) (56,70) \\
        1.250 (5:4)  & $1120 \times 896$  & (1,1) (2,2) (3,3) (5,4) (10,8) (15,12) (20,16) (25,20) (30,24) (35,28) (45,36) (55,44) (70,56) \\
        0.750 (3:4)  & $864 \times 1152$  & (1,1) (2,2) (3,4) (6,8) (9,12) (12,16) (15,20) (18,24) (21,28) (27,36) (36,48) (45,60) (54,72) \\
        1.333 (4:3)  & $1152 \times 864$  & (1,1) (2,2) (4,3) (8,6) (12,9) (16,12) (20,15) (24,18) (28,21) (36,27) (48,36) (60,45) (72,54) \\
        0.666 (2:3)  & $832 \times 1248$  & (1,1) (2,2) (2,3) (4,6) (6,9) (10,15) (14,21) (18,27) (22,33) (26,39) (32,48) (42,63) (52,78) \\
        1.500 (3:2)  & $1248 \times 832$  & (1,1) (2,2) (3,2) (6,4) (9,6) (15,10) (21,14) (27,18) (33,22) (39,26) (48,32) (63,42) (78,52) \\
        0.571 (4:7)  & $768 \times 1344$  & (1,1) (2,2) (3,3) (4,7) (6,11) (8,14) (12,21) (16,28) (20,35) (24,42) (32,56) (40,70) (48,84) \\
        1.750 (7:4)  & $1344 \times 768$  & (1,1) (2,2) (3,3) (7,4) (11,6) (14,8) (21,12) (28,16) (35,20) (42,24) (56,32) (70,40) (84,48) \\
        0.500 (1:2)  & $720 \times 1440$  & (1,1) (2,2) (2,4) (3,6) (5,10) (8,16) (11,22) (15,30) (19,38) (23,46) (30,60) (37,74) (45,90) \\
        2.000 (2:1)  & $1440 \times 720$  & (1,1) (2,2) (4,2) (6,3) (10,5) (16,8) (22,11) (30,15) (38,19) (46,23) (60,30) (74,37) (90,45) \\
        0.400 (2:5)  & $640 \times 1600$  & (1,1) (2,2) (2,5) (4,10) (6,15) (8,20) (10,25) (12,30) (16,40) (20,50) (26,65) (32,80) (40,100) \\
        2.500 (5:2)  & $1600 \times 640$  & (1,1) (2,2) (5,2) (10,4) (15,6) (20,8) (25,10) (30,12) (40,16) (50,20) (65,26) (80,32) (100,40) \\
        0.333 (1:3)  & $592 \times 1776$  & (1,1) (2,2) (2,6) (3,9) (5,15) (7,21) (9,27) (12,36) (15,45) (18,54) (24,72) (30,90) (37,111) \\
        3.000 (3:1)  & $1776 \times 592$  & (1,1) (2,2) (6,2) (9,3) (15,5) (21,7) (27,9) (36,12) (45,15) (54,18) (72,24) (90,30) (111,37) \\
        \bottomrule
    \end{tabular}}
\end{table}

\section{Details of Class-Free Guidance}
\label{sec-cfg}

We follow previous works~\citep{chen2025sharegpt,BAGEL-7B_arxiv_2025} to use CFG for enhanced visual generation quality. For training, we randomly drop tokens of conditional text and reference image with probabilities 0.1. For inference, we combine conditional and unconditional predicted logits to produce outputs.

Specifically, for \textbf{text-to-image generation}, the final logits \(\textbf{L}_{\text{final}}\) are computed as a linear combination of the conditional logits \(\textbf{L}_t\) (with text input) and unconditional logits \(\textbf{L}_{\emptyset}\) (without text input):

\begin{equation}
\textbf{L}_{\text{final}} = \lambda_t \cdot \textbf{L}_t + (1 - \lambda_t) \cdot \textbf{L}_{\emptyset}
\end{equation}

where \(\lambda_t\) is the text guidance scale controlling the influence of the text condition.

For \textbf{image editing} tasks, which involve both textual and reference image conditions, we employ a dual-guidance mechanism. Let \(\textbf{L}_{t,i}\) denote the logits with both text and reference image conditions, \(\textbf{L}_t\) the logits with text condition only, and \(\textbf{L}_{\emptyset}\) the logits without any conditions. The refined logits \(\textbf{L}_c\) are first obtained by blending \(\textbf{L}_{t,i}\) and \(\textbf{L}_t\) using a reference image guidance scale \(\lambda_i\):

\begin{equation}
\textbf{L}_c = \frac{\textbf{L}_{t,i} + \lambda_i \cdot \textbf{L}_t}{1 + \lambda_i}
\end{equation}

Then, the final output logits \(\textbf{L}_{\text{final}}\) are computed by combining \(\textbf{L}_c\) with the fully unconditional logits \(\textbf{L}_{\emptyset}\) using the text guidance scale \(\lambda_t\):

\begin{equation}
\textbf{L}_{\text{final}} = \textbf{L}_{\emptyset} + \lambda_t \cdot (\textbf{L}_c - \textbf{L}_{\emptyset})
\end{equation}

This approach allows flexible control over the influence of both textual and visual conditions during the image editing process. 

In our experiments, for text-to-image generation, we set \(\lambda_t = 20\). For image editing, we set \(\lambda_i = 1\) and \(\lambda_t = 3\).

\section{Additional ablation study}
\label{sec-appendix-addition-ablation}

\subsection{Effect on Early and Late Fusion in MLLM.}
\begin{figure}
    \centering
    \includegraphics[width=1\linewidth]{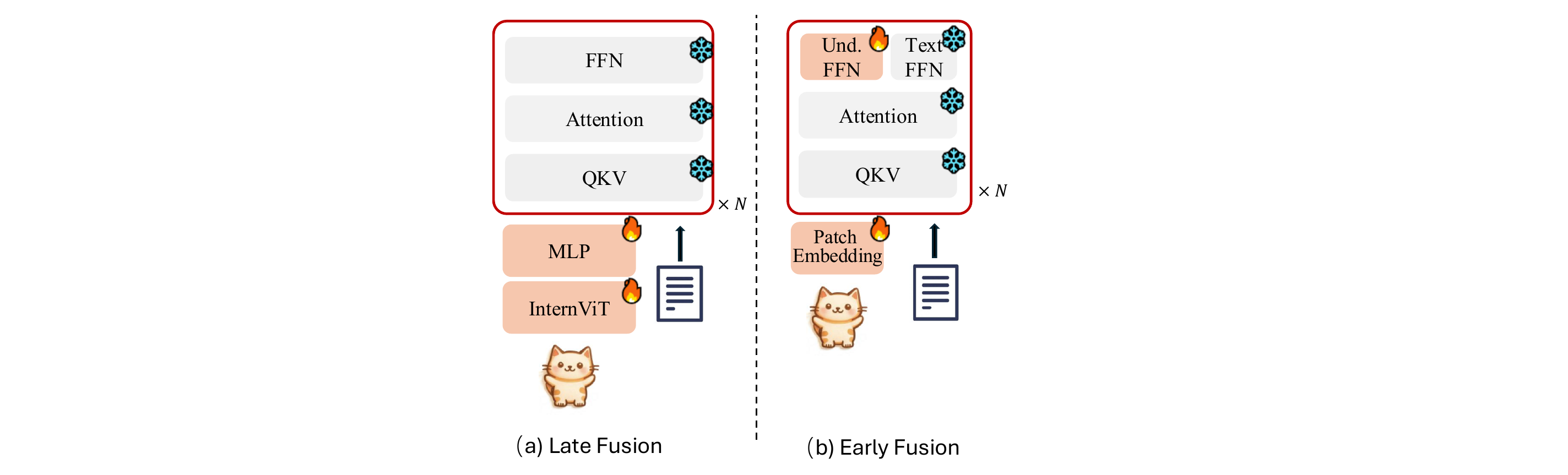}
    \caption{Comparison of late fusion and early fusion for multimodal understanding in our setting. We omit the \texttt{Gen.~FFN} in this figure for clarity.}
    \label{fig:appendix-early-fusion}
\end{figure}

\begin{table}[!t]
\small
\centering
\captionof{table}{Performance comparison of different fusion strategies and distillation methods for multimodal understanding across varying training scales.}
\setlength{\belowcaptionskip}{0pt}\renewcommand{\arraystretch}{1.2}
\setlength\tabcolsep{2.5pt}
\resizebox{0.75\textwidth}{!}{
    \begin{tabular}{l c ccccccc c}
        \toprule
        Methods & \#Trained Tokens of Stage-1& MMB & MME-S & MMVet & SEED & AI2D & ChartQA & TextVQA&\textbf{Avg.} \\
        \midrule
        &8B& 43.0 & 1222 & 13.3 & 46.8 & 52.1 &10.1 & 10.6 & 31.4 \\
        Encoder-based Late Fusion& 20B  
        & 49.0 & 1437 & 16.7 & 50.4 & 54.4 & 11.6 & 11.3 & 35.0 \\
        & 70B & 51.7 & 1426 & 19.2 & 55.4 & 55.9 & 12.0 & 19.7 & 37.8 \\
        \midrule
         & 8B & 42.0 & 1209 & 13.2 & 47.0 & 53.8 & 10.5 & 10.0 & 31.4 \\
        Decoder-only Early Fusion& 20B 
        & 45.7 & 1312 & 16.7 & 51.8 & 56.4 & 11.7 &11.9 & 34.4 \\

        & 70B & 50.9 & 1423 & 16.9 & 57.4 & 57.2 & 14.2 & 21.0 & 38.3 \\
        \midrule
        & 8B &  49.4 & 1327 & 15.5 & 56.3 & 54.4 & 11.9 & 11.3 & 35.3  \\
         Decoder-only Early Fusion& 20B & 54.3 & 1410 &17.7 & 61.0&55.4 &13.6 & 15.8 & 38.3 \\
        + Proposed Distillation& 70B & 57.6 & 1476 & 23.4 & 63.0 & 57.2 & 15.0 & 25.0 & 42.0 \\
        & 300B & 60.7 & 1526 & 27.1 & 63.4 & 60.0 & 19.2 & 35.7 & 45.8 \\
        \bottomrule
    \end{tabular}
}
\label{table:scaling_comparison}
\end{table}

\begin{figure}[!t]
    \centering
\includegraphics[width=0.75\linewidth]{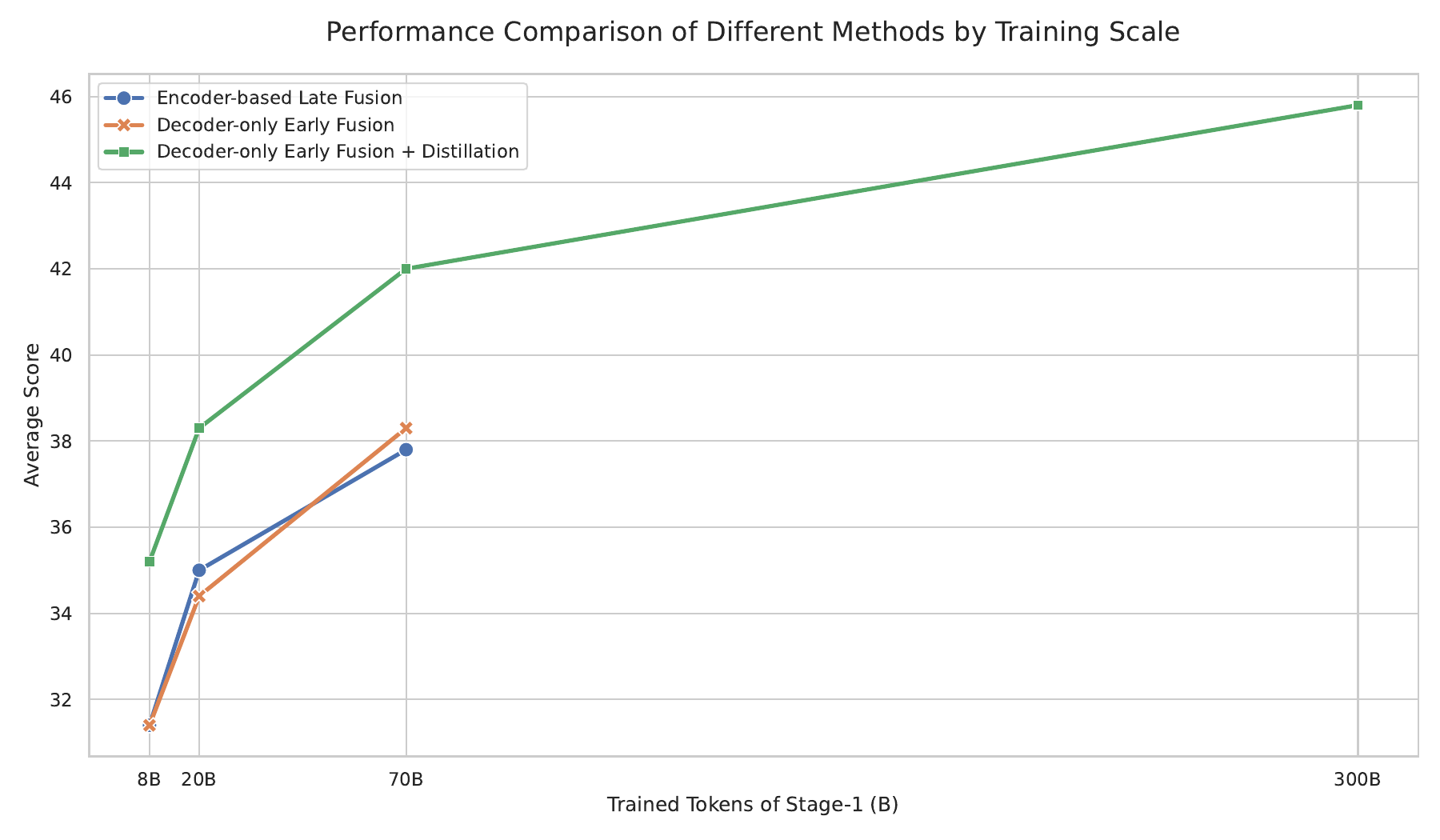}
    \caption{Performance comparison of different  methods for multimodal understanding across varying training scales.}
    \label{fig:scaling_comparison}
\end{figure}
We conduct an ablation study to evaluate the impact of encoder-based early fusion versus encoder-free late fusion strategies in multimodal understanding, aiming to validate the effectiveness of the decoder-only architecture employed in our model. As illustrated in Fig.~\ref{fig:appendix-early-fusion}, late fusion corresponds to the conventional encoder-based MLLM approach, where images are first processed by a vision encoder before being fed into the LLM. In contrast, early fusion represents the decoder-only MLLM paradigm of our proposed OneCAT.

To ensure a fair comparison, the vision encoder (InternViT) in the late fusion model was randomly initialized without pretrained weights, while the LLM in both models was initialized from the pretrained Qwen2.5-1.5B-instruct and kept frozen, aligning with the setup of our stage-1 training. All variants are further empoloyed a simplified SFT using LLaVA-665k dataset.

 Fig.~\ref{fig:scaling_comparison} compare the scaling properties of both models with varying trained token budgets for stage-1 and the detailed values are shown in Tab.~\ref{table:scaling_comparison}.  
The experimental results demonstrate that the early and late fusion perform on par, while the early fusion model offers a distinct advantage in computational efficiency, aligning with the findings of \cite{shukor2025scaling}.

\subsection{Effect on Distilling Only Visual Tokens}

We use the same setting of Sec.~\ref{sec-abla-stage-2} to conduct an ablation study to evaluate the impact of distilling different types of tokens. Tab.~\ref{table:effect-distill-only-visual} shows that distilling only the continuous visual tokens results in a slight overall performance drop, suggesting that it is crucial to distill both visual and text tokens.
\begin{table}[h!]
    \small
    \centering
 \caption{Effect of distilling only visual tokens}  
    \setlength{\belowcaptionskip}{0pt}\renewcommand{\arraystretch}{1.2}
        \setlength\tabcolsep{2.5pt}
            \resizebox{0.8\textwidth}{!}{
                \begin{tabular}{c cccccccc cc}
                    \toprule
                    Methods & MMB & MME-S & MMVet & SEED & AI2D & ChartQA & TextVQA & \textbf{Avg.} \\
                    \midrule
                    w/o distillation & 42.0 & 1209 & 13.2 & 47.0 & 53.8 & 10.5 & 10.0 & 31.4 \\
                    distill only visual tokens & 48.1 & 1299 & 16.7 & 55.7 & 55.2 & 11.4 & 10.5 & 34.8 \\
                    \rowcolor{tablerowcolor}
                    distill both visual and text tokens & 49.4 & 1327 & 15.5 & 56.3 & 54.4 & 11.9 & 11.3 & \textbf{35.3} \\
                    \bottomrule
                \end{tabular}
            }
            \label{table:effect-distill-only-visual}
            \end{table}


\subsection{Effect on The Increase of  Training Tokens of Unified Mid-Training (Stage-2) }

\begin{figure}[!t]
    \centering
\includegraphics[width=0.9\linewidth]{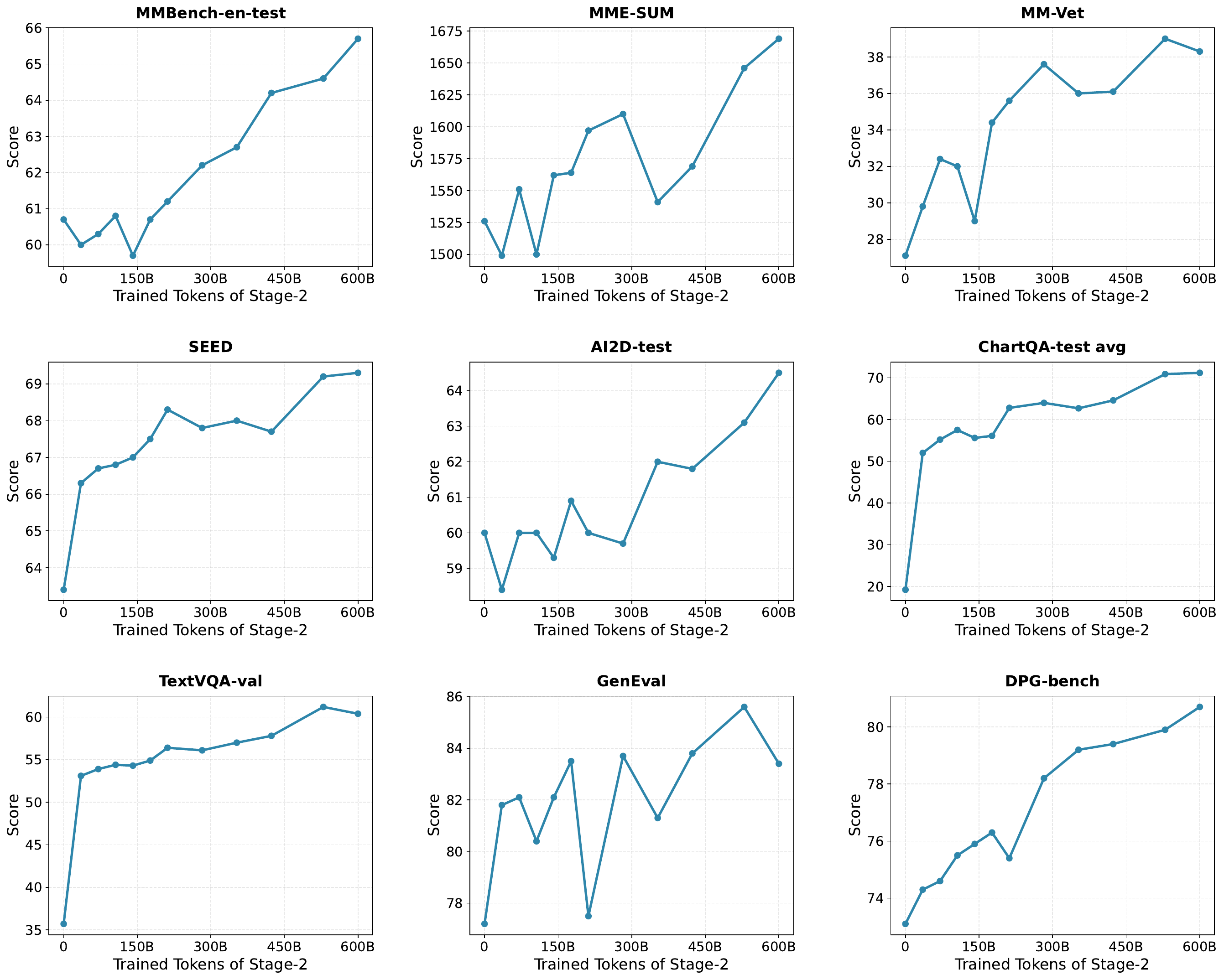}
    \caption{Performance of OneCAT-1.5B on different multimodal understanding and generation benchmarks with the increase of training tokens of unified mid-training (Stage-2).}
    \label{fig:scaling-pt2}
\end{figure}

Fig.~\ref{fig:scaling-pt2} provides the performance of our OneCAT-1.5B model on multimodal understanding and generation benchmarks at various checkpoints throughout the unified mid-training stage, corresponding to different amounts of training tokens.

Before downstream evaluation, the model of each checkpoint undergoes a simplified unified SFT with a combined instruction dataset (LLaVA-665K for understanding and BLIP3o-60K for generation), and we oversample the BLIP3o-60K dataset to achieve a 1:1 training token ratio for understanding and generation tasks.

\section{Visualization of discrete visual tokens of different scales}
\label{sec-visual-saa}
\begin{figure}[h!]
    \centering
    \includegraphics[width=1\linewidth]{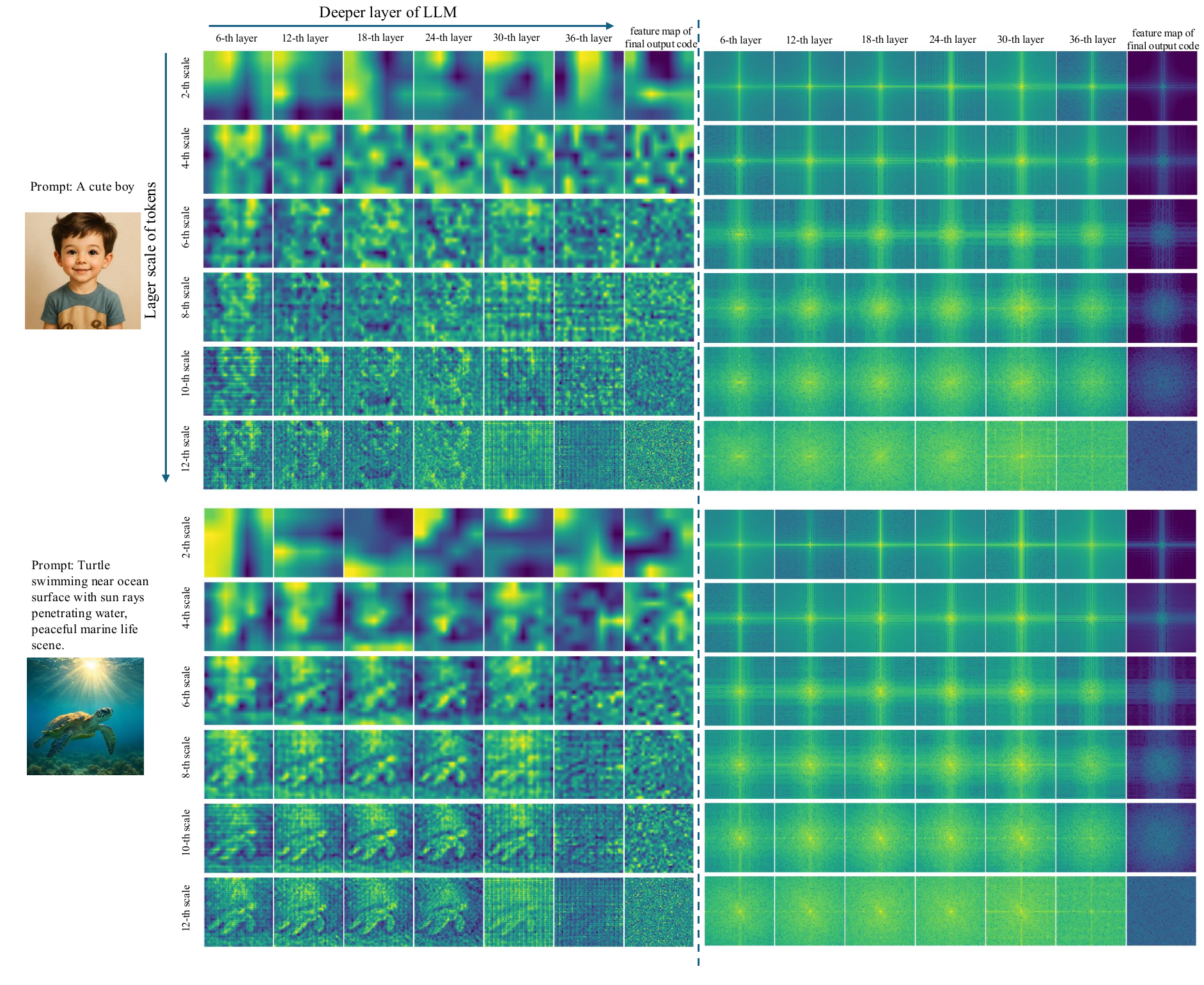}
    \caption{Visualization of discrete visual tokens across scales and LLM layers. \textbf{Left:} Each row shows the reshaped feature maps of token hidden states at a specific scale throughout LLM layers. The final column displays the feature map of final output codes fed to the image detokenizer for image reconstruction. All features are resized to 64$\times$64 for display. \textbf{Right:} Frequency intensity map of the corresponding feature maps. Lighter colors indicate larger
magnitudes, while pixels closer to the center represent lower frequencies. Zoom in better.}
    \label{fig:appendix-SAA}
\end{figure}

We generate two 1024$\times$ 1024 images and present the visualization of discrete visual tokens across different scales and LLM layers.  We also visualize the intensity of frequency component by
applying Fast Fourier Transform (FFT) to the corresponding tokens' feature maps. As shown in Fig~\ref{fig:appendix-SAA}, the results show that tokens at lower scales primarily encode low-frequency global information, while higher-scale tokens capture high-frequency details, validating the design rationale of our scale-aware adapter.

\section{More Qualitative results}
\label{sec-appendix-visualization}
\begin{figure}[!t]
\centering
\includegraphics[width=0.75\linewidth]{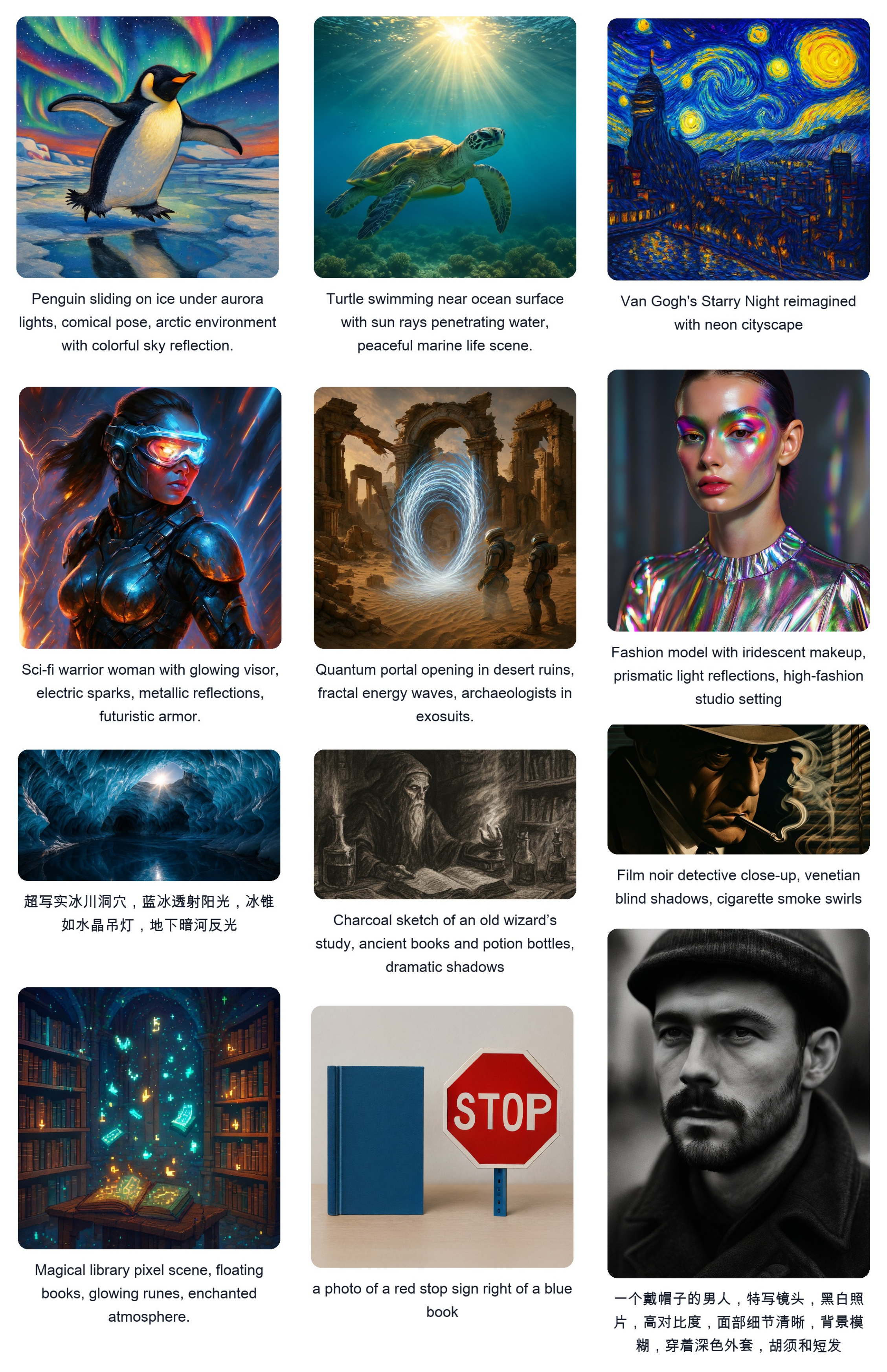}
\caption{Showcase of the text-to-image abilities of the \ourmodel{} model.}
\label{draw_t2i_2}
\end{figure}

\begin{figure}[!t]
\centering
\includegraphics[width=0.75\linewidth]{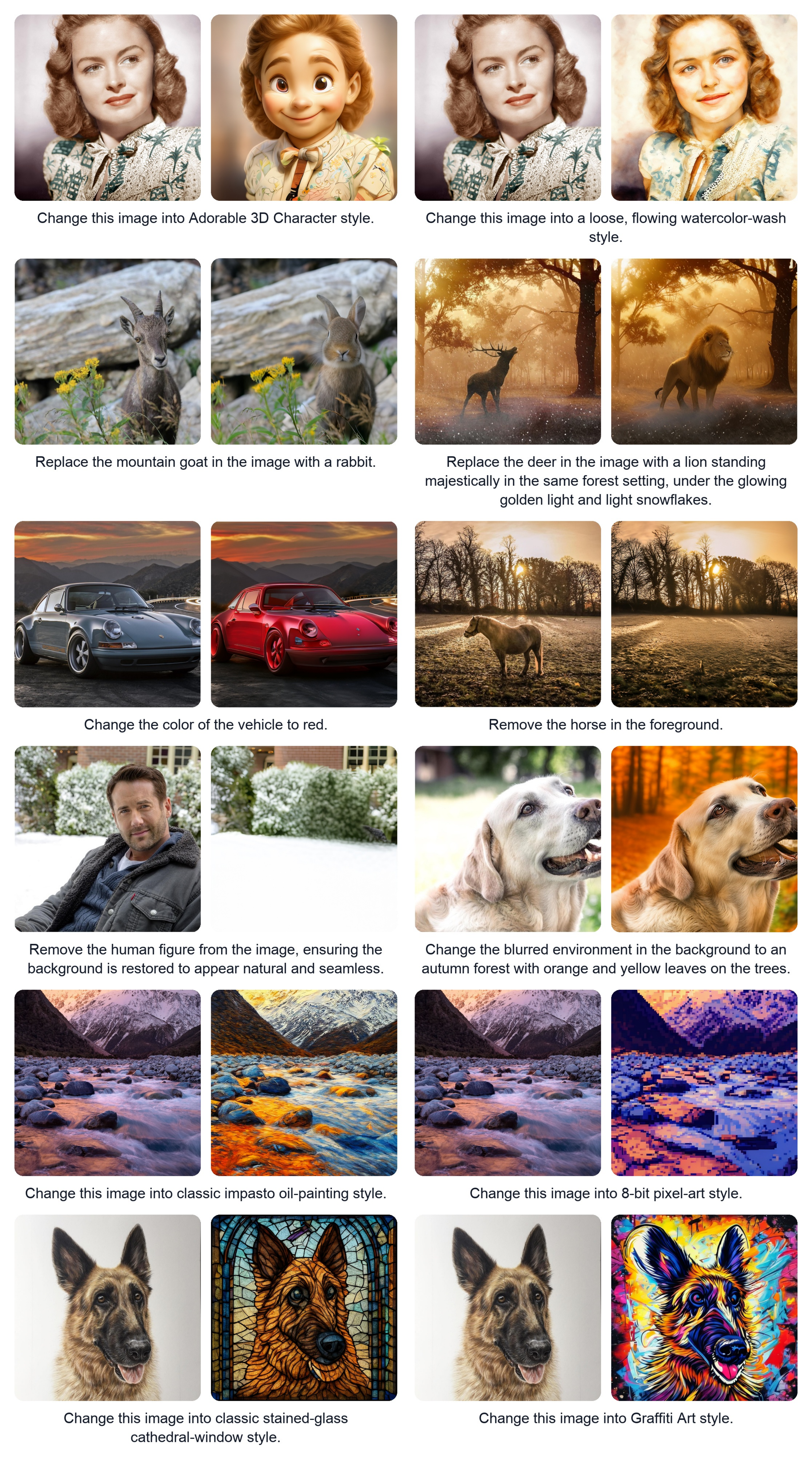}
\caption{Showcase of the image editing abilities of the \ourmodel{} model.}
\label{draw_edit_2}
\end{figure}

\begin{figure}[!t]
\centering
\includegraphics[width=1\linewidth]{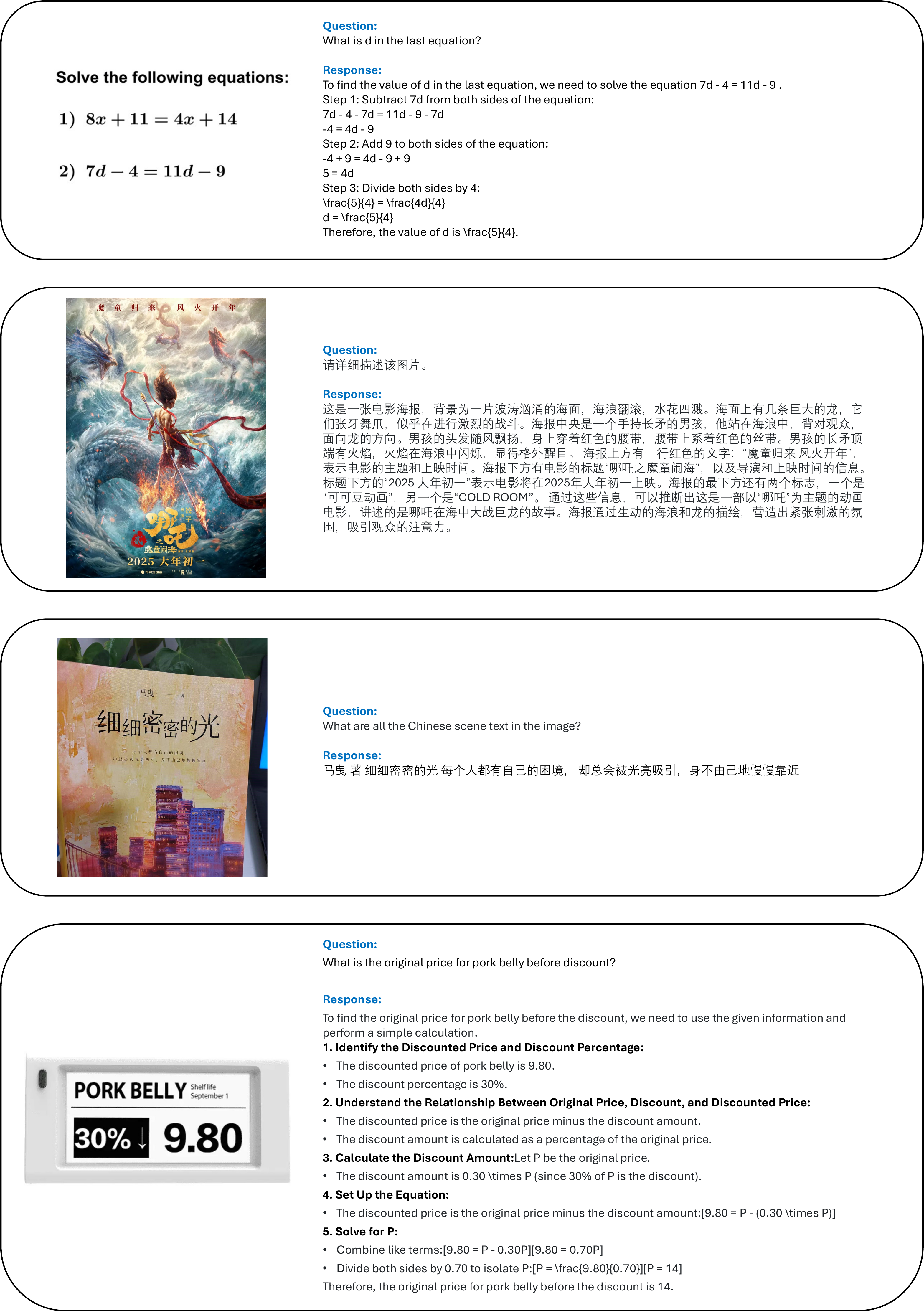}
\caption{Showcase of the visual understanding abilities of the \ourmodel{} model.}
\label{draw_und}
\end{figure}

In Fig.~\ref{draw_t2i} and ~\ref{draw_edit}, we present qualitative comparisons for text-to-image generation and image editing against several open-source models—including Janus-Pro~\citep{Janus-pro_arxiv_2025}, BAGEL~\citep{BAGEL-7B_arxiv_2025}, and UniWorld-V1~\citep{UniWorld-V1_arxiv_2025}—as well as the proprietary model GPT-4o-image~\citep{GPT4o_2024}.
We further present additional qualitative results to comprehensively demonstrate the capabilities of our model. Fig.~\ref{draw_t2i_2} presents text-to-image generation results from OneCAT under various aspect ratios and resolutions. Fig.~\ref{draw_edit_2} showcases OneCAT's performance on a range of image editing tasks, such as style transfer, object adjustment, attribute modification, object removal, and background editing. Additionally, Fig.~\ref{draw_und} provides examples of OneCAT's multimodal understanding abilities across mathematical reasoning, optical character recognition (OCR), and detailed image captioning.

%% file: main.bbl
\begin{thebibliography}{121}
\providecommand{\natexlab}[1]{#1}
\providecommand{\url}[1]{\texttt{#1}}
\expandafter\ifx\csname urlstyle\endcsname\relax
  \providecommand{\doi}[1]{doi: #1}\else
  \providecommand{\doi}{doi: \begingroup \urlstyle{rm}\Url}\fi

\bibitem[pdf()]{pdfa-eng-wds}
pdfa-eng-wds.
\newblock URL \url{https://huggingface.co/datasets/pixparse/pdfa-eng-wds}.

\bibitem[Acharya et~al.(2019)Acharya, Kafle, and Kanan]{acharya2019tallyqa}
Manoj Acharya, Kushal Kafle, and Christopher Kanan.
\newblock Tallyqa: Answering complex counting questions.
\newblock In \emph{Proceedings of the AAAI conference on artificial intelligence}, volume~33, pages 8076--8084, 2019.

\bibitem[Alayrac et~al.(2022)Alayrac, Donahue, Luc, Miech, Barr, Hasson, Lenc, Mensch, Millican, Reynolds, et~al.]{VLLMs_flamingo_2022_nips}
Jean-Baptiste Alayrac, Jeff Donahue, Pauline Luc, Antoine Miech, Iain Barr, Yana Hasson, Karel Lenc, Arthur Mensch, Katherine Millican, Malcolm Reynolds, et~al.
\newblock Flamingo: a visual language model for few-shot learning.
\newblock \emph{Advances in neural information processing systems}, 35:\penalty0 23716--23736, 2022.

\bibitem[Bai et~al.(2023)Bai, Bai, Yang, Wang, Tan, Wang, Lin, Zhou, and Zhou]{Qwen-VL_arxiv_2023}
Jinze Bai, Shuai Bai, Shusheng Yang, Shijie Wang, Sinan Tan, Peng Wang, Junyang Lin, Chang Zhou, and Jingren Zhou.
\newblock Qwen-vl: A versatile vision-language model for understanding, localization, text reading, and beyond.
\newblock \emph{arXiv preprint arXiv:2308.12966}, 2023.

\bibitem[Bai et~al.(2025)Bai, Chen, Liu, Wang, Ge, Song, Dang, Wang, Wang, Tang, et~al.]{Qwen2.5_arxiv_2025}
Shuai Bai, Keqin Chen, Xuejing Liu, Jialin Wang, Wenbin Ge, Sibo Song, Kai Dang, Peng Wang, Shijie Wang, Jun Tang, et~al.
\newblock Qwen2. 5-vl technical report.
\newblock \emph{arXiv preprint arXiv:2502.13923}, 2025.

\bibitem[Bavishi et~al.(2023)Bavishi, Elsen, Hawthorne, Nye, Odena, Somani, and Ta\c{s}\i{}rlar]{fuyu-8b_2023}
Rohan Bavishi, Erich Elsen, Curtis Hawthorne, Maxwell Nye, Augustus Odena, Arushi Somani, and Sa\u{g}nak Ta\c{s}\i{}rlar.
\newblock Introducing our multimodal models, 2023.
\newblock URL \url{https://www.adept.ai/blog/fuyu-8b}.

\bibitem[Brooks et~al.(2023)Brooks, Holynski, and Efros]{Instruct-Pix2Pix_2023_CVPR}
Tim Brooks, Aleksander Holynski, and Alexei~A Efros.
\newblock Instructpix2pix: Learning to follow image editing instructions.
\newblock In \emph{Proceedings of the IEEE/CVF conference on computer vision and pattern recognition}, pages 18392--18402, 2023.

\bibitem[Byeon et~al.(2022)Byeon, Park, Kim, Lee, Baek, and Kim]{kakaobrain2022coyo-700m}
Minwoo Byeon, Beomhee Park, Haecheon Kim, Sungjun Lee, Woonhyuk Baek, and Saehoon Kim.
\newblock Coyo-700m: Image-text pair dataset.
\newblock \url{https://github.com/kakaobrain/coyo-dataset}, 2022.

\bibitem[Changpinyo et~al.(2021)Changpinyo, Sharma, Ding, and Soricut]{changpinyo2021conceptual}
Soravit Changpinyo, Piyush Sharma, Nan Ding, and Radu Soricut.
\newblock Conceptual 12m: Pushing web-scale image-text pre-training to recognize long-tail visual concepts.
\newblock In \emph{Proceedings of the IEEE/CVF conference on computer vision and pattern recognition}, pages 3558--3568, 2021.

\bibitem[Chen et~al.(2024{\natexlab{a}})Chen, Chen, Zhang, Chen, Wu, Zhang, Chen, Li, Wan, and Wang]{chen2024allava}
Guiming~Hardy Chen, Shunian Chen, Ruifei Zhang, Junying Chen, Xiangbo Wu, Zhiyi Zhang, Zhihong Chen, Jianquan Li, Xiang Wan, and Benyou Wang.
\newblock Allava: Harnessing gpt4v-synthesized data for lite vision-language models.
\newblock \emph{arXiv preprint arXiv:2402.11684}, 2024{\natexlab{a}}.

\bibitem[Chen et~al.(2025{\natexlab{a}})Chen, Xu, Pan, Hu, Qin, Goldstein, Huang, Zhou, Xie, Savarese, et~al.]{BLIP3-O_arxiv_2025}
Jiuhai Chen, Zhiyang Xu, Xichen Pan, Yushi Hu, Can Qin, Tom Goldstein, Lifu Huang, Tianyi Zhou, Saining Xie, Silvio Savarese, et~al.
\newblock Blip3-o: A family of fully open unified multimodal models-architecture, training and dataset.
\newblock \emph{arXiv preprint arXiv:2505.09568}, 2025{\natexlab{a}}.

\bibitem[Chen et~al.(2024{\natexlab{b}})Chen, Ge, Xie, Wu, Yao, Ren, Wang, Luo, Lu, and Li]{Pixart_ECCV_2024}
Junsong Chen, Chongjian Ge, Enze Xie, Yue Wu, Lewei Yao, Xiaozhe Ren, Zhongdao Wang, Ping Luo, Huchuan Lu, and Zhenguo Li.
\newblock Pixart-$\sigma$: Weak-to-strong training of diffusion transformer for 4k text-to-image generation.
\newblock In \emph{European Conference on Computer Vision}, pages 74--91. Springer, 2024{\natexlab{b}}.

\bibitem[Chen et~al.(2025{\natexlab{b}})Chen, Cai, Chen, Chen, Ji, Wang, Yang, and Wang]{chen2025sharegpt}
Junying Chen, Zhenyang Cai, Pengcheng Chen, Shunian Chen, Ke~Ji, Xidong Wang, Yunjin Yang, and Benyou Wang.
\newblock Sharegpt-4o-image: Aligning multimodal models with gpt-4o-level image generation.
\newblock \emph{arXiv preprint arXiv:2506.18095}, 2025{\natexlab{b}}.

\bibitem[Chen et~al.(2024{\natexlab{c}})Chen, Li, Dong, Zhang, He, Wang, Zhao, and Lin]{chen2024sharegpt4v}
Lin Chen, Jinsong Li, Xiaoyi Dong, Pan Zhang, Conghui He, Jiaqi Wang, Feng Zhao, and Dahua Lin.
\newblock Sharegpt4v: Improving large multi-modal models with better captions.
\newblock In \emph{European Conference on Computer Vision}, pages 370--387. Springer, 2024{\natexlab{c}}.

\bibitem[Chen et~al.(2025{\natexlab{c}})Chen, Wu, Liu, Pan, Liu, Xie, Yu, and Ruan]{Janus-pro_arxiv_2025}
Xiaokang Chen, Zhiyu Wu, Xingchao Liu, Zizheng Pan, Wen Liu, Zhenda Xie, Xingkai Yu, and Chong Ruan.
\newblock Janus-pro: Unified multimodal understanding and generation with data and model scaling.
\newblock \emph{arXiv preprint arXiv:2501.17811}, 2025{\natexlab{c}}.

\bibitem[Chen et~al.(2021)Chen, Zhao, Chen, Zhang, Ji, Luo, Xiong, and Yu]{chen2021websrc}
Xingyu Chen, Zihan Zhao, Lu~Chen, Danyang Zhang, Jiabao Ji, Ao~Luo, Yuxuan Xiong, and Kai Yu.
\newblock Websrc: A dataset for web-based structural reading comprehension.
\newblock \emph{arXiv preprint arXiv:2101.09465}, 2021.

\bibitem[Chen et~al.(2024{\natexlab{d}})Chen, Wang, Cao, Liu, Gao, Cui, Zhu, Ye, Tian, Liu, et~al.]{InternVL2.5_arxiv_2024}
Zhe Chen, Weiyun Wang, Yue Cao, Yangzhou Liu, Zhangwei Gao, Erfei Cui, Jinguo Zhu, Shenglong Ye, Hao Tian, Zhaoyang Liu, et~al.
\newblock Expanding performance boundaries of open-source multimodal models with model, data, and test-time scaling.
\newblock \emph{arXiv preprint arXiv:2412.05271}, 2024{\natexlab{d}}.

\bibitem[Chen et~al.(2024{\natexlab{e}})Chen, Wang, Tian, Ye, Gao, Cui, Tong, Hu, Luo, Ma, et~al.]{InternVL2_arxiv_2024}
Zhe Chen, Weiyun Wang, Hao Tian, Shenglong Ye, Zhangwei Gao, Erfei Cui, Wenwen Tong, Kongzhi Hu, Jiapeng Luo, Zheng Ma, et~al.
\newblock How far are we to gpt-4v? closing the gap to commercial multimodal models with open-source suites.
\newblock \emph{arXiv preprint arXiv:2404.16821}, 2024{\natexlab{e}}.

\bibitem[Chen et~al.(2024{\natexlab{f}})Chen, Wu, Wang, Su, Chen, Xing, Zhong, Zhang, Zhu, Lu, Li, Luo, Lu, Qiao, and Dai]{internvit}
Zhe Chen, Jiannan Wu, Wenhai Wang, Weijie Su, Guo Chen, Sen Xing, Muyan Zhong, Qinglong Zhang, Xizhou Zhu, Lewei Lu, Bin Li, Ping Luo, Tong Lu, Yu~Qiao, and Jifeng Dai.
\newblock Internvl: Scaling up vision foundation models and aligning for generic visual-linguistic tasks.
\newblock In \emph{Proceedings of the IEEE/CVF Conference on Computer Vision and Pattern Recognition (CVPR)}, 2024{\natexlab{f}}.

\bibitem[Data(2024)]{Tongyi_SA1B}
Tongyi Data.
\newblock Sa1b-dense-caption dataset, 2024.
\newblock URL \url{https://www.modelscope.cn/datasets/Tongyi-DataEngine/SA1B-Dense-Caption}.

\bibitem[Deng et~al.(2025)Deng, Zhu, Li, Gou, Li, Wang, Zhong, Yu, Nie, Song, et~al.]{BAGEL-7B_arxiv_2025}
Chaorui Deng, Deyao Zhu, Kunchang Li, Chenhui Gou, Feng Li, Zeyu Wang, Shu Zhong, Weihao Yu, Xiaonan Nie, Ziang Song, et~al.
\newblock Emerging properties in unified multimodal pretraining.
\newblock \emph{arXiv preprint arXiv:2505.14683}, 2025.

\bibitem[Deng et~al.(2009)Deng, Dong, Socher, Li, Li, and Fei-Fei]{deng2009imagenet}
Jia Deng, Wei Dong, Richard Socher, Li-Jia Li, Kai Li, and Li~Fei-Fei.
\newblock Imagenet: A large-scale hierarchical image database.
\newblock In \emph{2009 IEEE conference on computer vision and pattern recognition}, pages 248--255. Ieee, 2009.

\bibitem[Diao et~al.(2024)Diao, Cui, Li, Wang, Lu, and Wang]{EvE_nips_2024}
Haiwen Diao, Yufeng Cui, Xiaotong Li, Yueze Wang, Huchuan Lu, and Xinlong Wang.
\newblock Unveiling encoder-free vision-language models.
\newblock \emph{Advances in Neural Information Processing Systems}, 37:\penalty0 52545--52567, 2024.

\bibitem[Diao et~al.(2025)Diao, Li, Cui, Wang, Deng, Pan, Wang, Lu, and Wang]{EvEv2.0_arxiv_2025}
Haiwen Diao, Xiaotong Li, Yufeng Cui, Yueze Wang, Haoge Deng, Ting Pan, Wenxuan Wang, Huchuan Lu, and Xinlong Wang.
\newblock Evev2: Improved baselines for encoder-free vision-language models.
\newblock \emph{arXiv preprint arXiv:2502.06788}, 2025.

\bibitem[Esser et~al.(2024)Esser, Kulal, Blattmann, Entezari, M{\"u}ller, Saini, Levi, Lorenz, Sauer, Boesel, et~al.]{SD3-Medium_ICML_2024}
Patrick Esser, Sumith Kulal, Andreas Blattmann, Rahim Entezari, Jonas M{\"u}ller, Harry Saini, Yam Levi, Dominik Lorenz, Axel Sauer, Frederic Boesel, et~al.
\newblock Scaling rectified flow transformers for high-resolution image synthesis.
\newblock In \emph{Forty-first international conference on machine learning}, 2024.

\bibitem[Fu et~al.(2023)Fu, Tamir, Sundaram, Chai, Zhang, Dekel, and Isola]{fu2023dreamsim}
Stephanie Fu, Netanel Tamir, Shobhita Sundaram, Lucy Chai, Richard Zhang, Tali Dekel, and Phillip Isola.
\newblock Dreamsim: Learning new dimensions of human visual similarity using synthetic data.
\newblock \emph{arXiv preprint arXiv:2306.09344}, 2023.

\bibitem[Ghosh et~al.(2023)Ghosh, Hajishirzi, and Schmidt]{Benchmark_geneval_nips_2023}
Dhruba Ghosh, Hannaneh Hajishirzi, and Ludwig Schmidt.
\newblock Geneval: An object-focused framework for evaluating text-to-image alignment.
\newblock \emph{Advances in Neural Information Processing Systems}, 36:\penalty0 52132--52152, 2023.

\bibitem[Goyal et~al.(2017)Goyal, Khot, Summers-Stay, Batra, and Parikh]{goyal2017making}
Yash Goyal, Tejas Khot, Douglas Summers-Stay, Dhruv Batra, and Devi Parikh.
\newblock Making the v in vqa matter: Elevating the role of image understanding in visual question answering.
\newblock In \emph{Proceedings of the IEEE conference on computer vision and pattern recognition}, pages 6904--6913, 2017.

\bibitem[Guo et~al.(2024)Guo, Zheng, Bai, Li, Wang, Zhu, Li, Neubig, Chen, and Yue]{dataset_2024_mammoth}
Jarvis Guo, Tuney Zheng, Yuelin Bai, Bo~Li, Yubo Wang, King Zhu, Yizhi Li, Graham Neubig, Wenhu Chen, and Xiang Yue.
\newblock Mammoth-vl: Eliciting multimodal reasoning with instruction tuning at scale.
\newblock \emph{arXiv preprint arXiv:2412.05237}, 2024.

\bibitem[Han et~al.(2025{\natexlab{a}})Han, Chen, Zhao, Wang, Zhao, Yang, He, Yue, and Jiang]{Tar-7B_arxiv_2025}
Jiaming Han, Hao Chen, Yang Zhao, Hanyu Wang, Qi~Zhao, Ziyan Yang, Hao He, Xiangyu Yue, and Lu~Jiang.
\newblock Vision as a dialect: Unifying visual understanding and generation via text-aligned representations.
\newblock \emph{arXiv preprint arXiv:2506.18898}, 2025{\natexlab{a}}.

\bibitem[Han et~al.(2025{\natexlab{b}})Han, Liu, Jiang, Yan, Zhang, Yuan, Peng, and Liu]{infinity_CVPR_2025}
Jian Han, Jinlai Liu, Yi~Jiang, Bin Yan, Yuqi Zhang, Zehuan Yuan, Bingyue Peng, and Xiaobing Liu.
\newblock Infinity: Scaling bitwise autoregressive modeling for high-resolution image synthesis.
\newblock In \emph{Proceedings of the Computer Vision and Pattern Recognition Conference}, pages 15733--15744, 2025{\natexlab{b}}.

\bibitem[Ho and Salimans(2022)]{ho2022classifier}
Jonathan Ho and Tim Salimans.
\newblock Classifier-free diffusion guidance.
\newblock \emph{arXiv preprint arXiv:2207.12598}, 2022.

\bibitem[Hu et~al.(2022)Hu, Shen, Wallis, Allen-Zhu, Li, Wang, Wang, Chen, et~al.]{LORA_ICLR_2022}
Edward~J Hu, Yelong Shen, Phillip Wallis, Zeyuan Allen-Zhu, Yuanzhi Li, Shean Wang, Lu~Wang, Weizhu Chen, et~al.
\newblock Lora: Low-rank adaptation of large language models.
\newblock \emph{ICLR}, 1\penalty0 (2):\penalty0 3, 2022.

\bibitem[Hu et~al.(2024)Hu, Wang, Fang, Fu, Cheng, and Yu]{Benchmark_DPG-Bench_arxiv_2024}
Xiwei Hu, Rui Wang, Yixiao Fang, Bin Fu, Pei Cheng, and Gang Yu.
\newblock Ella: Equip diffusion models with llm for enhanced semantic alignment.
\newblock \emph{arXiv preprint arXiv:2403.05135}, 2024.

\bibitem[Huang et~al.(2025)Huang, Wang, Yang, Lu, Yuan, Han, Hou, Zhang, Hong, Zhao, et~al.]{ILLUME+3B_arxiv_2025}
Runhui Huang, Chunwei Wang, Junwei Yang, Guansong Lu, Yunlong Yuan, Jianhua Han, Lu~Hou, Wei Zhang, Lanqing Hong, Hengshuang Zhao, et~al.
\newblock Illume+: Illuminating unified mllm with dual visual tokenization and diffusion refinement.
\newblock \emph{arXiv preprint arXiv:2504.01934}, 2025.

\bibitem[Hudson and Manning(2019{\natexlab{a}})]{benchmark_GQA_CVPR_2019}
Drew~A Hudson and Christopher~D Manning.
\newblock Gqa: A new dataset for real-world visual reasoning and compositional question answering.
\newblock In \emph{Proceedings of the IEEE/CVF conference on computer vision and pattern recognition}, pages 6700--6709, 2019{\natexlab{a}}.

\bibitem[Hudson and Manning(2019{\natexlab{b}})]{hudson2019gqa}
Drew~A Hudson and Christopher~D Manning.
\newblock Gqa: A new dataset for real-world visual reasoning and compositional question answering.
\newblock In \emph{Proceedings of the IEEE/CVF conference on computer vision and pattern recognition}, pages 6700--6709, 2019{\natexlab{b}}.

\bibitem[Hui et~al.(2024)Hui, Yang, Zhao, Shi, Wang, Wang, Zhou, and Xie]{hui2024hq}
Mude Hui, Siwei Yang, Bingchen Zhao, Yichun Shi, Heng Wang, Peng Wang, Yuyin Zhou, and Cihang Xie.
\newblock Hq-edit: A high-quality dataset for instruction-based image editing.
\newblock \emph{arXiv preprint arXiv:2404.09990}, 2024.

\bibitem[Hurst et~al.(2024)Hurst, Lerer, Goucher, Perelman, Ramesh, Clark, Ostrow, Welihinda, Hayes, Radford, et~al.]{hurst2024gpt}
Aaron Hurst, Adam Lerer, Adam~P Goucher, Adam Perelman, Aditya Ramesh, Aidan Clark, AJ~Ostrow, Akila Welihinda, Alan Hayes, Alec Radford, et~al.
\newblock Gpt-4o system card.
\newblock \emph{arXiv preprint arXiv:2410.21276}, 2024.

\bibitem[isidentical(2024)]{isidentical_2024}
isidentical.
\newblock moondream2-coyo-5m-captions dataset, 2024.
\newblock URL \url{https://hf-mirror.com/datasets/isidentical/moondream2-coyo-5M-captions}.

\bibitem[Jiang et~al.(2025)Jiang, Fang, Zhang, Ma, Wan, Wang, He, and Chua]{Anyedit_arxiv_2025}
Houcheng Jiang, Junfeng Fang, Ningyu Zhang, Guojun Ma, Mingyang Wan, Xiang Wang, Xiangnan He, and Tat-seng Chua.
\newblock Anyedit: Edit any knowledge encoded in language models.
\newblock \emph{arXiv preprint arXiv:2502.05628}, 2025.

\bibitem[Kembhavi et~al.(2016{\natexlab{a}})Kembhavi, Salvato, Kolve, Seo, Hajishirzi, and Farhadi]{benchmark_AI2D_ECCV_2016}
Aniruddha Kembhavi, Mike Salvato, Eric Kolve, Minjoon Seo, Hannaneh Hajishirzi, and Ali Farhadi.
\newblock A diagram is worth a dozen images.
\newblock pages 235--251, 2016{\natexlab{a}}.

\bibitem[Kembhavi et~al.(2016{\natexlab{b}})Kembhavi, Salvato, Kolve, Seo, Hajishirzi, and Farhadi]{kembhavi2016diagram}
Aniruddha Kembhavi, Mike Salvato, Eric Kolve, Minjoon Seo, Hannaneh Hajishirzi, and Ali Farhadi.
\newblock A diagram is worth a dozen images.
\newblock In \emph{European conference on computer vision}, pages 235--251. Springer, 2016{\natexlab{b}}.

\bibitem[Krishna et~al.(2017)Krishna, Zhu, Groth, Johnson, Hata, Kravitz, Chen, Kalantidis, Li, Shamma, et~al.]{krishna2017visual}
Ranjay Krishna, Yuke Zhu, Oliver Groth, Justin Johnson, Kenji Hata, Joshua Kravitz, Stephanie Chen, Yannis Kalantidis, Li-Jia Li, David~A Shamma, et~al.
\newblock Visual genome: Connecting language and vision using crowdsourced dense image annotations.
\newblock \emph{International journal of computer vision}, 123\penalty0 (1):\penalty0 32--73, 2017.

\bibitem[Labs(2024)]{FLUX.1-dev_github_2024}
Black~Forest Labs.
\newblock Flux.
\newblock \url{https://github.com/black-forest-labs/flux}, 2024.

\bibitem[Lauren{\c{c}}on et~al.(2024)Lauren{\c{c}}on, Marafioti, Sanh, and Tronchon]{laurenccon2024building}
Hugo Lauren{\c{c}}on, Andr{\'e}s Marafioti, Victor Sanh, and L{\'e}o Tronchon.
\newblock Building and better understanding vision-language models: insights and future directions.
\newblock \emph{arXiv preprint arXiv:2408.12637}, 2024.

\bibitem[Lei et~al.(2025)Lei, Wang, Wang, Li, Liew, Feng, and Huang]{SAIL_2025_ICCV}
Weixian Lei, Jiacong Wang, Haochen Wang, Xiangtai Li, Jun~Hao Liew, Jiashi Feng, and Zilong Huang.
\newblock The scalability of simplicity: Empirical analysis of vision-language learning with a single transformer.
\newblock \emph{arXiv preprint arXiv:2504.10462}, 2025.

\bibitem[Li et~al.(2023{\natexlab{a}})Li, Wang, Wang, Ge, Ge, and Shan]{benchmark_SEED_arxiv_2023}
Bohao Li, Rui Wang, Guangzhi Wang, Yuying Ge, Yixiao Ge, and Ying Shan.
\newblock Seed-bench: Benchmarking multimodal llms with generative comprehension.
\newblock \emph{arxiv:2307.16125}, 2023{\natexlab{a}}.

\bibitem[Li et~al.(2024{\natexlab{a}})Li, Kamko, Akhgari, Sabet, Xu, and Doshi]{Playgroundv2.5_arxiv_2024}
Daiqing Li, Aleks Kamko, Ehsan Akhgari, Ali Sabet, Linmiao Xu, and Suhail Doshi.
\newblock Playground v2. 5: Three insights towards enhancing aesthetic quality in text-to-image generation.
\newblock \emph{arXiv preprint arXiv:2402.17245}, 2024{\natexlab{a}}.

\bibitem[Li et~al.(2025)Li, Tian, Shao, Zhu, Wang, Zhu, Dou, Wang, Li, Lu, et~al.]{li2025synergen}
Hao Li, Changyao Tian, Jie Shao, Xizhou Zhu, Zhaokai Wang, Jinguo Zhu, Wenhan Dou, Xiaogang Wang, Hongsheng Li, Lewei Lu, et~al.
\newblock Synergen-vl: Towards synergistic image understanding and generation with vision experts and token folding.
\newblock In \emph{Proceedings of the Computer Vision and Pattern Recognition Conference}, pages 29767--29779, 2025.

\bibitem[Li et~al.(2023{\natexlab{b}})Li, Li, Savarese, and Hoi]{VLLMs_ICML_2023_BLIP_2}
Junnan Li, Dongxu Li, Silvio Savarese, and Steven Hoi.
\newblock Blip-2: Bootstrapping language-image pre-training with frozen image encoders and large language models.
\newblock In \emph{International conference on machine learning}, pages 19730--19742. PMLR, 2023{\natexlab{b}}.

\bibitem[Li et~al.(2024{\natexlab{b}})Li, Tu, Hui, Wang, Zhao, Xiao, Ren, Mei, Liu, Zheng, et~al.]{li2024if}
Xianhang Li, Haoqin Tu, Mude Hui, Zeyu Wang, Bingchen Zhao, Junfei Xiao, Sucheng Ren, Jieru Mei, Qing Liu, Huangjie Zheng, et~al.
\newblock What if we recaption billions of web images with llama-3?
\newblock \emph{arXiv preprint arXiv:2406.08478}, 2024{\natexlab{b}}.

\bibitem[Li et~al.(2024{\natexlab{c}})Li, Yang, Liu, Ma, Zhang, Yang, Sun, Liu, and Bai]{li2024monkey}
Zhang Li, Biao Yang, Qiang Liu, Zhiyin Ma, Shuo Zhang, Jingxu Yang, Yabo Sun, Yuliang Liu, and Xiang Bai.
\newblock Monkey: Image resolution and text label are important things for large multi-modal models.
\newblock In \emph{proceedings of the IEEE/CVF conference on computer vision and pattern recognition}, pages 26763--26773, 2024{\natexlab{c}}.

\bibitem[Li et~al.(2024{\natexlab{d}})Li, Zhang, Lin, Xiong, Long, Deng, Zhang, Liu, Huang, Xiao, et~al.]{Hunyuan-DiT_arxiv_2024}
Zhimin Li, Jianwei Zhang, Qin Lin, Jiangfeng Xiong, Yanxin Long, Xinchi Deng, Yingfang Zhang, Xingchao Liu, Minbin Huang, Zedong Xiao, et~al.
\newblock Hunyuan-dit: A powerful multi-resolution diffusion transformer with fine-grained chinese understanding.
\newblock \emph{arXiv preprint arXiv:2405.08748}, 2024{\natexlab{d}}.

\bibitem[Liao et~al.(2025{\natexlab{a}})Liao, Liu, Wang, Luo, Zhang, Zhao, Wu, Li, Tian, and Huang]{Mogao_arxiv_2025}
Chao Liao, Liyang Liu, Xun Wang, Zhengxiong Luo, Xinyu Zhang, Wenliang Zhao, Jie Wu, Liang Li, Zhi Tian, and Weilin Huang.
\newblock Mogao: An omni foundation model for interleaved multi-modal generation.
\newblock \emph{arXiv preprint arXiv:2505.05472}, 2025{\natexlab{a}}.

\bibitem[Liao et~al.(2025{\natexlab{b}})Liao, Liu, Wang, Luo, Zhang, Zhao, Wu, Li, Tian, and Huang]{liao2025mogao}
Chao Liao, Liyang Liu, Xun Wang, Zhengxiong Luo, Xinyu Zhang, Wenliang Zhao, Jie Wu, Liang Li, Zhi Tian, and Weilin Huang.
\newblock Mogao: An omni foundation model for interleaved multi-modal generation.
\newblock \emph{arXiv preprint arXiv:2505.05472}, 2025{\natexlab{b}}.

\bibitem[Lin et~al.(2025{\natexlab{a}})Lin, Li, Cheng, Niu, Ye, He, Yuan, Yu, Wang, Ge, et~al.]{UniWorld-V1_arxiv_2025}
Bin Lin, Zongjian Li, Xinhua Cheng, Yuwei Niu, Yang Ye, Xianyi He, Shenghai Yuan, Wangbo Yu, Shaodong Wang, Yunyang Ge, et~al.
\newblock Uniworld: High-resolution semantic encoders for unified visual understanding and generation.
\newblock \emph{arXiv preprint arXiv:2506.03147}, 2025{\natexlab{a}}.

\bibitem[Lin et~al.(2025{\natexlab{b}})Lin, Li, Cheng, Niu, Ye, He, Yuan, Yu, Wang, Ge, et~al.]{lin2025uniworld}
Bin Lin, Zongjian Li, Xinhua Cheng, Yuwei Niu, Yang Ye, Xianyi He, Shenghai Yuan, Wangbo Yu, Shaodong Wang, Yunyang Ge, et~al.
\newblock Uniworld: High-resolution semantic encoders for unified visual understanding and generation.
\newblock \emph{arXiv preprint arXiv:2506.03147}, 2025{\natexlab{b}}.

\bibitem[Lindstr{\"o}m and Abraham(2022)]{lindstrom2022clevr}
Adam~Dahlgren Lindstr{\"o}m and Savitha~Sam Abraham.
\newblock Clevr-math: A dataset for compositional language, visual and mathematical reasoning.
\newblock \emph{arXiv preprint arXiv:2208.05358}, 2022.

\bibitem[Liu et~al.(2023)Liu, Li, Wu, and Lee]{VLLMs_MLP_Llava_2023_nips}
Haotian Liu, Chunyuan Li, Qingyang Wu, and Yong~Jae Lee.
\newblock Visual instruction tuning.
\newblock \emph{Advances in neural information processing systems}, 36, 2023.

\bibitem[Liu et~al.(2025)Liu, Han, Xing, Yin, Wang, Cheng, Liao, Wang, Fu, Han, et~al.]{Step1X-Edit_arxiv_2025}
Shiyu Liu, Yucheng Han, Peng Xing, Fukun Yin, Rui Wang, Wei Cheng, Jiaqi Liao, Yingming Wang, Honghao Fu, Chunrui Han, et~al.
\newblock Step1x-edit: A practical framework for general image editing.
\newblock \emph{arXiv preprint arXiv:2504.17761}, 2025.

\bibitem[Liu et~al.(2024)Liu, Duan, Zhang, Li, Zhang, Zhao, Yuan, Wang, He, Liu, et~al.]{benchmark_mmbench_ECCV_2024}
Yuan Liu, Haodong Duan, Yuanhan Zhang, Bo~Li, Songyang Zhang, Wangbo Zhao, Yike Yuan, Jiaqi Wang, Conghui He, Ziwei Liu, et~al.
\newblock Mmbench: Is your multi-modal model an all-around player?
\newblock In \emph{European conference on computer vision}, pages 216--233. Springer, 2024.

\bibitem[Lu et~al.(2023)Lu, Bansal, Xia, Liu, Li, Hajishirzi, Cheng, Chang, Galley, and Gao]{benchmark_mathvista_arxiv_2023}
Pan Lu, Hritik Bansal, Tony Xia, Jiacheng Liu, Chunyuan Li, Hannaneh Hajishirzi, Hao Cheng, Kai-Wei Chang, Michel Galley, and Jianfeng Gao.
\newblock Mathvista: Evaluating mathematical reasoning of foundation models in visual contexts.
\newblock \emph{arXiv preprint arXiv:2310.02255}, 2023.

\bibitem[Luo et~al.(2025)Luo, Yang, Dou, Wang, Liu, Dai, Qiao, and Zhu]{Mono-InternVL_CVPR_2025}
Gen Luo, Xue Yang, Wenhan Dou, Zhaokai Wang, Jiawen Liu, Jifeng Dai, Yu~Qiao, and Xizhou Zhu.
\newblock Mono-internvl: Pushing the boundaries of monolithic multimodal large language models with endogenous visual pre-training.
\newblock In \emph{Proceedings of the Computer Vision and Pattern Recognition Conference}, pages 24960--24971, 2025.

\bibitem[Mao et~al.(2017)Mao, Cheung, and She]{mao2017deepart}
Hui Mao, Ming Cheung, and James She.
\newblock Deepart: Learning joint representations of visual arts.
\newblock In \emph{Proceedings of the 25th ACM international conference on Multimedia}, pages 1183--1191, 2017.

\bibitem[Marino et~al.(2019)Marino, Rastegari, Farhadi, and Mottaghi]{marino2019ok}
Kenneth Marino, Mohammad Rastegari, Ali Farhadi, and Roozbeh Mottaghi.
\newblock Ok-vqa: A visual question answering benchmark requiring external knowledge.
\newblock In \emph{Proceedings of the IEEE/cvf conference on computer vision and pattern recognition}, pages 3195--3204, 2019.

\bibitem[Masry et~al.(2022)Masry, Long, Tan, Joty, and Hoque]{benchmark_ChartQA_arxiv_2022}
Ahmed Masry, Do~Xuan Long, Jia~Qing Tan, Shafiq Joty, and Enamul Hoque.
\newblock Chartqa: A benchmark for question answering about charts with visual and logical reasoning.
\newblock \emph{arxiv:2203.10244}, 2022.

\bibitem[Mathew et~al.(2021{\natexlab{a}})Mathew, Karatzas, and Jawahar]{benchmark_DocVQA_WACV_2021}
Minesh Mathew, Dimosthenis Karatzas, and CV~Jawahar.
\newblock Docvqa: A dataset for vqa on document images.
\newblock In \emph{WACV}, pages 2200--2209, 2021{\natexlab{a}}.

\bibitem[Mathew et~al.(2021{\natexlab{b}})Mathew, Karatzas, and Jawahar]{mathew2021docvqa}
Minesh Mathew, Dimosthenis Karatzas, and CV~Jawahar.
\newblock Docvqa: A dataset for vqa on document images.
\newblock In \emph{Proceedings of the IEEE/CVF winter conference on applications of computer vision}, pages 2200--2209, 2021{\natexlab{b}}.

\bibitem[Mathew et~al.(2022)Mathew, Bagal, Tito, Karatzas, Valveny, and Jawahar]{benchmark_infoVQA_WACV_2022}
Minesh Mathew, Viraj Bagal, Rub{\`e}n Tito, Dimosthenis Karatzas, Ernest Valveny, and CV~Jawahar.
\newblock Infographicvqa.
\newblock In \emph{WACV}, pages 1697--1706, 2022.

\bibitem[Mishra et~al.(2019)Mishra, Shekhar, Singh, and Chakraborty]{mishra2019ocr}
Anand Mishra, Shashank Shekhar, Ajeet~Kumar Singh, and Anirban Chakraborty.
\newblock Ocr-vqa: Visual question answering by reading text in images.
\newblock In \emph{2019 international conference on document analysis and recognition (ICDAR)}, pages 947--952. IEEE, 2019.

\bibitem[OpenAI(2024)]{GPT4o_2024}
OpenAI.
\newblock Hello gpt-4o.
\newblock \url{https://openai.com/index/hello-gpt-4o/}, 2024.

\bibitem[Pan et~al.(2025)Pan, Shukla, Singh, Zhao, Mishra, Wang, Xu, Chen, Li, Juefei-Xu, et~al.]{Metaqueries_arxiv_2025}
Xichen Pan, Satya~Narayan Shukla, Aashu Singh, Zhuokai Zhao, Shlok~Kumar Mishra, Jialiang Wang, Zhiyang Xu, Jiuhai Chen, Kunpeng Li, Felix Juefei-Xu, et~al.
\newblock Transfer between modalities with metaqueries.
\newblock \emph{arXiv preprint arXiv:2504.06256}, 2025.

\bibitem[Peng et~al.(2023)Peng, Wang, Dong, Hao, Huang, Ma, and Wei]{Kosmos2}
Zhiliang Peng, Wenhui Wang, Li~Dong, Yaru Hao, Shaohan Huang, Shuming Ma, and Furu Wei.
\newblock Kosmos-2: Grounding multimodal large language models to the world.
\newblock \emph{ArXiv}, abs/2306.14824, 2023.

\bibitem[Podell et~al.(2023)Podell, English, Lacey, Blattmann, Dockhorn, M{\"u}ller, Penna, and Rombach]{sdxl_arxiv_2023}
Dustin Podell, Zion English, Kyle Lacey, Andreas Blattmann, Tim Dockhorn, Jonas M{\"u}ller, Joe Penna, and Robin Rombach.
\newblock Sdxl: Improving latent diffusion models for high-resolution image synthesis.
\newblock \emph{arXiv preprint arXiv:2307.01952}, 2023.

\bibitem[{PyTorch Team}(2024)]{pytorch_team_flexattention_2024}
{PyTorch Team}.
\newblock {FlexAttention}: The flexibility of pytorch with the performance of flashattention.
\newblock PyTorch Blog, 2024.
\newblock URL \url{https://pytorch.org/blog/flexattention/}.
\newblock Accessed: 2024-09-03.

\bibitem[Radford et~al.(2021)Radford, Kim, Hallacy, Ramesh, Goh, Agarwal, Sastry, Askell, Mishkin, Clark, et~al.]{MLLMs_2021_ICML_CLIP}
Alec Radford, Jong~Wook Kim, Chris Hallacy, Aditya Ramesh, Gabriel Goh, Sandhini Agarwal, Girish Sastry, Amanda Askell, Pamela Mishkin, Jack Clark, et~al.
\newblock Learning transferable visual models from natural language supervision.
\newblock In \emph{International conference on machine learning}, pages 8748--8763. PMLR, 2021.

\bibitem[Saikh et~al.(2022)Saikh, Ghosal, Mittal, Ekbal, and Bhattacharyya]{saikh2022scienceqa}
Tanik Saikh, Tirthankar Ghosal, Amish Mittal, Asif Ekbal, and Pushpak Bhattacharyya.
\newblock Scienceqa: A novel resource for question answering on scholarly articles.
\newblock \emph{International Journal on Digital Libraries}, 23\penalty0 (3):\penalty0 289--301, 2022.

\bibitem[Schuhmann et~al.(2021)Schuhmann, Vencu, Beaumont, Kaczmarczyk, Mullis, Katta, Coombes, Jitsev, and Komatsuzaki]{schuhmann2021laion}
Christoph Schuhmann, Richard Vencu, Romain Beaumont, Robert Kaczmarczyk, Clayton Mullis, Aarush Katta, Theo Coombes, Jenia Jitsev, and Aran Komatsuzaki.
\newblock Laion-400m: Open dataset of clip-filtered 400 million image-text pairs.
\newblock \emph{arXiv preprint arXiv:2111.02114}, 2021.

\bibitem[Sharma et~al.(2018)Sharma, Ding, Goodman, and Soricut]{sharma2018conceptual}
Piyush Sharma, Nan Ding, Sebastian Goodman, and Radu Soricut.
\newblock Conceptual captions: A cleaned, hypernymed, image alt-text dataset for automatic image captioning.
\newblock In \emph{Proceedings of the 56th Annual Meeting of the Association for Computational Linguistics (Volume 1: Long Papers)}, pages 2556--2565, 2018.

\bibitem[Shi et~al.(2020)Shi, Zhou, Qiu, and Zhu]{DALL-E_3_arxiv_2020}
Zhan Shi, Xu~Zhou, Xipeng Qiu, and Xiaodan Zhu.
\newblock Improving image captioning with better use of captions.
\newblock \emph{arXiv preprint arXiv:2006.11807}, 2020.

\bibitem[Shukor et~al.(2025)Shukor, Fini, da~Costa, Cord, Susskind, and El-Nouby]{shukor2025scaling}
Mustafa Shukor, Enrico Fini, Victor Guilherme~Turrisi da~Costa, Matthieu Cord, Joshua Susskind, and Alaaeldin El-Nouby.
\newblock Scaling laws for native multimodal models.
\newblock \emph{arXiv preprint arXiv:2504.07951}, 2025.

\bibitem[Singh et~al.(2019)Singh, Natarajan, Shah, Jiang, Chen, Batra, Parikh, and Rohrbach]{benchmark_TextVQA_CVPR_2019}
Amanpreet Singh, Vivek Natarajan, Meet Shah, Yu~Jiang, Xinlei Chen, Dhruv Batra, Devi Parikh, and Marcus Rohrbach.
\newblock Towards vqa models that can read.
\newblock In \emph{CVPR}, pages 8317--8326, 2019.

\bibitem[Singh et~al.(2024)Singh, Yadav, Jain, Shi, Johnson, and Desai]{singh2024benchmarking}
Shweta Singh, Aayan Yadav, Jitesh Jain, Humphrey Shi, Justin Johnson, and Karan Desai.
\newblock Benchmarking object detectors with coco: A new path forward.
\newblock In \emph{European Conference on Computer Vision}, pages 279--295. Springer, 2024.

\bibitem[Srinivasan et~al.(2021)Srinivasan, Raman, Chen, Bendersky, and Najork]{srinivasan2021wit}
Krishna Srinivasan, Karthik Raman, Jiecao Chen, Michael Bendersky, and Marc Najork.
\newblock Wit: Wikipedia-based image text dataset for multimodal multilingual machine learning.
\newblock In \emph{Proceedings of the 44th international ACM SIGIR conference on research and development in information retrieval}, pages 2443--2449, 2021.

\bibitem[Sun et~al.(2023)Sun, Pan, Ge, Li, Duan, Wu, Zhang, Zhou, Qin, Wang, et~al.]{sun2023journeydb}
Keqiang Sun, Junting Pan, Yuying Ge, Hao Li, Haodong Duan, Xiaoshi Wu, Renrui Zhang, Aojun Zhou, Zipeng Qin, Yi~Wang, et~al.
\newblock Journeydb: A benchmark for generative image understanding.
\newblock \emph{Advances in neural information processing systems}, 36:\penalty0 49659--49678, 2023.

\bibitem[Tao et~al.(2025)Tao, Su, Zhu, Zhang, Chen, Liu, Wang, Lu, Huang, Qiao, et~al.]{HoVLE_CVPR_2025}
Chenxin Tao, Shiqian Su, Xizhou Zhu, Chenyu Zhang, Zhe Chen, Jiawen Liu, Wenhai Wang, Lewei Lu, Gao Huang, Yu~Qiao, et~al.
\newblock Hovle: Unleashing the power of monolithic vision-language models with holistic vision-language embedding.
\newblock In \emph{Proceedings of the Computer Vision and Pattern Recognition Conference}, pages 14559--14569, 2025.

\bibitem[Team(2024)]{Chameleon-7B_arxiv_2024}
Chameleon Team.
\newblock Chameleon: Mixed-modal early-fusion foundation models.
\newblock \emph{arXiv preprint arXiv:2405.09818}, 2024.

\bibitem[Tian et~al.(2024)Tian, Jiang, Yuan, Peng, and Wang]{VAR_nips_2024}
Keyu Tian, Yi~Jiang, Zehuan Yuan, Bingyue Peng, and Liwei Wang.
\newblock Visual autoregressive modeling: Scalable image generation via next-scale prediction.
\newblock \emph{Advances in neural information processing systems}, 37:\penalty0 84839--84865, 2024.

\bibitem[Tschannen et~al.(2025)Tschannen, Gritsenko, Wang, Naeem, Alabdulmohsin, Parthasarathy, Evans, Beyer, Xia, Mustafa, et~al.]{MLLMs_2025_arxiv_siglip2}
Michael Tschannen, Alexey Gritsenko, Xiao Wang, Muhammad~Ferjad Naeem, Ibrahim Alabdulmohsin, Nikhil Parthasarathy, Talfan Evans, Lucas Beyer, Ye~Xia, Basil Mustafa, et~al.
\newblock Siglip 2: Multilingual vision-language encoders with improved semantic understanding, localization, and dense features.
\newblock \emph{arXiv preprint arXiv:2502.14786}, 2025.

\bibitem[Wang et~al.(2025{\natexlab{a}})Wang, Ye, Li, Nie, Lu, Tang, Wang, and Huang]{vora_2025_arxiv}
Han Wang, Yongjie Ye, Bingru Li, Yuxiang Nie, Jinghui Lu, Jingqun Tang, Yanjie Wang, and Can Huang.
\newblock Vision as lora.
\newblock \emph{arXiv preprint arXiv:2503.20680}, 2025{\natexlab{a}}.

\bibitem[Wang et~al.(2024{\natexlab{a}})Wang, Bai, Tan, Wang, Fan, Bai, Chen, Liu, Wang, Ge, et~al.]{Qwen2-VL_arxiv_2024}
Peng Wang, Shuai Bai, Sinan Tan, Shijie Wang, Zhihao Fan, Jinze Bai, Keqin Chen, Xuejing Liu, Jialin Wang, Wenbin Ge, et~al.
\newblock Qwen2-vl: Enhancing vision-language model's perception of the world at any resolution.
\newblock \emph{arXiv preprint arXiv:2409.12191}, 2024{\natexlab{a}}.

\bibitem[Wang et~al.(2025{\natexlab{b}})Wang, Gao, Gu, Pu, Cui, Wei, Liu, Jing, Ye, Shao, et~al.]{internvl3.5}
Weiyun Wang, Zhangwei Gao, Lixin Gu, Hengjun Pu, Long Cui, Xingguang Wei, Zhaoyang Liu, Linglin Jing, Shenglong Ye, Jie Shao, et~al.
\newblock Internvl3. 5: Advancing open-source multimodal models in versatility, reasoning, and efficiency.
\newblock \emph{arXiv preprint arXiv:2508.18265}, 2025{\natexlab{b}}.

\bibitem[Wang et~al.(2024{\natexlab{b}})Wang, Zhang, Luo, Sun, Cui, Wang, Zhang, Wang, Li, Yu, et~al.]{Emu3-Gen_arxiv_2024}
Xinlong Wang, Xiaosong Zhang, Zhengxiong Luo, Quan Sun, Yufeng Cui, Jinsheng Wang, Fan Zhang, Yueze Wang, Zhen Li, Qiying Yu, et~al.
\newblock Emu3: Next-token prediction is all you need.
\newblock \emph{arXiv preprint arXiv:2409.18869}, 2024{\natexlab{b}}.

\bibitem[Wei et~al.(2024)Wei, Xiong, Ren, Du, Zhang, and Chen]{wei2024omniedit}
Cong Wei, Zheyang Xiong, Weiming Ren, Xeron Du, Ge~Zhang, and Wenhu Chen.
\newblock Omniedit: Building image editing generalist models through specialist supervision.
\newblock In \emph{The Thirteenth International Conference on Learning Representations}, 2024.

\bibitem[Wu et~al.(2025{\natexlab{a}})Wu, Chen, Wu, Ma, Liu, Pan, Liu, Xie, Yu, Ruan, et~al.]{Janus_CVPR_2025}
Chengyue Wu, Xiaokang Chen, Zhiyu Wu, Yiyang Ma, Xingchao Liu, Zizheng Pan, Wen Liu, Zhenda Xie, Xingkai Yu, Chong Ruan, et~al.
\newblock Janus: Decoupling visual encoding for unified multimodal understanding and generation.
\newblock In \emph{Proceedings of the Computer Vision and Pattern Recognition Conference}, pages 12966--12977, 2025{\natexlab{a}}.

\bibitem[Wu et~al.(2025{\natexlab{b}})Wu, Zheng, Yan, Xiao, Luo, Wang, Li, Jiang, Liu, Zhou, et~al.]{OmniGen2_arxiv_2025}
Chenyuan Wu, Pengfei Zheng, Ruiran Yan, Shitao Xiao, Xin Luo, Yueze Wang, Wanli Li, Xiyan Jiang, Yexin Liu, Junjie Zhou, et~al.
\newblock Omnigen2: Exploration to advanced multimodal generation.
\newblock \emph{arXiv preprint arXiv:2506.18871}, 2025{\natexlab{b}}.

\bibitem[Wu et~al.(2025{\natexlab{c}})Wu, Zhang, Xu, Jin, Wu, Tao, Liu, Li, and Loy]{Harmon-1.5B_arxiv_2025}
Size Wu, Wenwei Zhang, Lumin Xu, Sheng Jin, Zhonghua Wu, Qingyi Tao, Wentao Liu, Wei Li, and Chen~Change Loy.
\newblock Harmonizing visual representations for unified multimodal understanding and generation.
\newblock \emph{arXiv preprint arXiv:2503.21979}, 2025{\natexlab{c}}.

\bibitem[Wu et~al.(2024)Wu, Zhang, Chen, Tang, Li, Fang, Zhu, Xie, Yin, Yi, et~al.]{VILA-U_arxiv_2024}
Yecheng Wu, Zhuoyang Zhang, Junyu Chen, Haotian Tang, Dacheng Li, Yunhao Fang, Ligeng Zhu, Enze Xie, Hongxu Yin, Li~Yi, et~al.
\newblock Vila-u: a unified foundation model integrating visual understanding and generation.
\newblock \emph{arXiv preprint arXiv:2409.04429}, 2024.

\bibitem[Xiao et~al.(2025)Xiao, Wang, Zhou, Yuan, Xing, Yan, Li, Wang, Huang, and Liu]{OmniGen_CVPR_2025}
Shitao Xiao, Yueze Wang, Junjie Zhou, Huaying Yuan, Xingrun Xing, Ruiran Yan, Chaofan Li, Shuting Wang, Tiejun Huang, and Zheng Liu.
\newblock Omnigen: Unified image generation.
\newblock In \emph{Proceedings of the Computer Vision and Pattern Recognition Conference}, pages 13294--13304, 2025.

\bibitem[Xie et~al.(2023)Xie, Cai, Li, Kong, Wu, Song, Morimitsu, Yao, Wang, Zhang, et~al.]{xie2023ccmb}
Chunyu Xie, Heng Cai, Jincheng Li, Fanjing Kong, Xiaoyu Wu, Jianfei Song, Henrique Morimitsu, Lin Yao, Dexin Wang, Xiangzheng Zhang, et~al.
\newblock Ccmb: A large-scale chinese cross-modal benchmark.
\newblock In \emph{Proceedings of the 31st ACM International Conference on Multimedia}, pages 4219--4227, 2023.

\bibitem[Xie et~al.(2025{\natexlab{a}})Xie, Darrell, Zettlemoyer, and Wang]{xie2025reconstructionalignmentimprovesunified}
Ji~Xie, Trevor Darrell, Luke Zettlemoyer, and XuDong Wang.
\newblock Reconstruction alignment improves unified multimodal models, 2025{\natexlab{a}}.
\newblock URL \url{https://arxiv.org/abs/2509.07295}.

\bibitem[Xie et~al.(2024)Xie, Mao, Bai, Zhang, Wang, Lin, Gu, Chen, Yang, and Shou]{show-o_arxiv_2024}
Jinheng Xie, Weijia Mao, Zechen Bai, David~Junhao Zhang, Weihao Wang, Kevin~Qinghong Lin, Yuchao Gu, Zhijie Chen, Zhenheng Yang, and Mike~Zheng Shou.
\newblock Show-o: One single transformer to unify multimodal understanding and generation.
\newblock \emph{arXiv preprint arXiv:2408.12528}, 2024.

\bibitem[Xie et~al.(2025{\natexlab{b}})Xie, Yang, and Shou]{show-o2_arxiv_2025}
Jinheng Xie, Zhenheng Yang, and Mike~Zheng Shou.
\newblock Show-o2: Improved native unified multimodal models.
\newblock \emph{arXiv preprint arXiv:2506.15564}, 2025{\natexlab{b}}.

\bibitem[Yang et~al.(2024)Yang, Yang, Zhang, Hui, Zheng, Yu, Li, Liu, Huang, Dong, Wei, Lin, Yang, Tu, Zhang, Yang, Yang, Zhou, Lin, Dang, Lu, Bao, Yang, Yu, Li, Xue, Zhang, Zhu, Men, Lin, Li, Xia, Ren, Ren, Fan, Su, Zhang, Wan, Liu, Cui, Zhang, Qiu, Quan, and Wang]{qwen2.5}
An~Yang, Baosong Yang, Beichen Zhang, Binyuan Hui, Bo~Zheng, Bowen Yu, Chengyuan Li, Dayiheng Liu, Fei Huang, Guanting Dong, Haoran Wei, Huan Lin, Jian Yang, Jianhong Tu, Jianwei Zhang, Jianxin Yang, Jiaxin Yang, Jingren Zhou, Junyang Lin, Kai Dang, Keming Lu, Keqin Bao, Kexin Yang, Le~Yu, Mei Li, Mingfeng Xue, Pei Zhang, Qin Zhu, Rui Men, Runji Lin, Tianhao Li, Tingyu Xia, Xingzhang Ren, Xuancheng Ren, Yang Fan, Yang Su, Yi-Chao Zhang, Yunyang Wan, Yuqi Liu, Zeyu Cui, Zhenru Zhang, Zihan Qiu, Shanghaoran Quan, and Zekun Wang.
\newblock Qwen2.5 technical report.
\newblock \emph{arXiv preprint arXiv:2412.15115}, 2024.

\bibitem[Ye et~al.(2023)Ye, Hu, Xu, Ye, Yan, Xu, Li, Tian, Qian, Zhang, et~al.]{ye2023ureader}
Jiabo Ye, Anwen Hu, Haiyang Xu, Qinghao Ye, Ming Yan, Guohai Xu, Chenliang Li, Junfeng Tian, Qi~Qian, Ji~Zhang, et~al.
\newblock Ureader: Universal ocr-free visually-situated language understanding with multimodal large language model.
\newblock \emph{arXiv preprint arXiv:2310.05126}, 2023.

\bibitem[Ye et~al.(2025)Ye, He, Li, Lin, Yuan, Yan, Hou, and Yuan]{Benchmark_imgedit_arxiv_2025}
Yang Ye, Xianyi He, Zongjian Li, Bin Lin, Shenghai Yuan, Zhiyuan Yan, Bohan Hou, and Li~Yuan.
\newblock Imgedit: A unified image editing dataset and benchmark.
\newblock \emph{arXiv preprint arXiv:2505.20275}, 2025.

\bibitem[Yin et~al.(2024)Yin, Fu, Zhao, Li, Sun, Xu, and Chen]{benchmark_MME_ECCV_2024}
Shukang Yin, Chaoyou Fu, Sirui Zhao, Ke~Li, Xing Sun, Tong Xu, and Enhong Chen.
\newblock A survey on multimodal large language models.
\newblock \emph{National Science Review}, 11\penalty0 (12):\penalty0 nwae403, 2024.

\bibitem[Yu et~al.(2025)Yu, Chow, Yue, Pan, Wu, Wan, Li, Tang, Zhang, and Zhuang]{yu2025anyedit}
Qifan Yu, Wei Chow, Zhongqi Yue, Kaihang Pan, Yang Wu, Xiaoyang Wan, Juncheng Li, Siliang Tang, Hanwang Zhang, and Yueting Zhuang.
\newblock Anyedit: Mastering unified high-quality image editing for any idea.
\newblock In \emph{Proceedings of the Computer Vision and Pattern Recognition Conference}, pages 26125--26135, 2025.

\bibitem[Yu et~al.(2024)Yu, Sun, Zhang, Cui, Zhang, Cao, Wang, and Liu]{yu2024capsfusion}
Qiying Yu, Quan Sun, Xiaosong Zhang, Yufeng Cui, Fan Zhang, Yue Cao, Xinlong Wang, and Jingjing Liu.
\newblock Capsfusion: Rethinking image-text data at scale.
\newblock In \emph{Proceedings of the IEEE/CVF Conference on Computer Vision and Pattern Recognition}, pages 14022--14032, 2024.

\bibitem[Yu et~al.(2023)Yu, Yang, Li, Wang, Lin, Liu, Wang, and Wang]{benchmark_MM-vet_arxiv_2023}
Weihao Yu, Zhengyuan Yang, Linjie Li, Jianfeng Wang, Kevin Lin, Zicheng Liu, Xinchao Wang, and Lijuan Wang.
\newblock Mm-vet: Evaluating large multimodal models for integrated capabilities.
\newblock \emph{arXiv preprint arXiv:2308.02490}, 2023.

\bibitem[Yue et~al.(2024)Yue, Ni, Zhang, Zheng, Liu, Zhang, Stevens, Jiang, Ren, Sun, et~al.]{benchmark_MMMU_CVPR_2024}
Xiang Yue, Yuansheng Ni, Kai Zhang, Tianyu Zheng, Ruoqi Liu, Ge~Zhang, Samuel Stevens, Dongfu Jiang, Weiming Ren, Yuxuan Sun, et~al.
\newblock Mmmu: A massive multi-discipline multimodal understanding and reasoning benchmark for expert agi.
\newblock In \emph{Proceedings of the IEEE/CVF Conference on Computer Vision and Pattern Recognition}, pages 9556--9567, 2024.

\bibitem[Zhai et~al.(2023)Zhai, Mustafa, Kolesnikov, and Beyer]{MLLMs_2023_arxiv_siglip}
Xiaohua Zhai, Basil Mustafa, Alexander Kolesnikov, and Lucas Beyer.
\newblock Sigmoid loss for language image pre-training.
\newblock In \emph{Proceedings of the IEEE/CVF International Conference on Computer Vision}, pages 11975--11986, 2023.

\bibitem[Zhang et~al.(2023)Zhang, Mo, Chen, Sun, and Su]{Magicbrush_2023_nips}
Kai Zhang, Lingbo Mo, Wenhu Chen, Huan Sun, and Yu~Su.
\newblock Magicbrush: A manually annotated dataset for instruction-guided image editing.
\newblock \emph{Advances in Neural Information Processing Systems}, 36:\penalty0 31428--31449, 2023.

\bibitem[Zhang et~al.(2025)Zhang, Xie, Lu, Yang, and Yang]{ICEdit_arxiv_2025}
Zechuan Zhang, Ji~Xie, Yu~Lu, Zongxin Yang, and Yi~Yang.
\newblock In-context edit: Enabling instructional image editing with in-context generation in large scale diffusion transformer.
\newblock \emph{arXiv preprint arXiv:2504.20690}, 2025.

\bibitem[Zhao et~al.(2024{\natexlab{a}})Zhao, Ma, Chen, Si, Wu, An, Yu, Zhang, Li, and Chang]{UltraEdit_2024_nips}
Haozhe Zhao, Xiaojian~Shawn Ma, Liang Chen, Shuzheng Si, Rujie Wu, Kaikai An, Peiyu Yu, Minjia Zhang, Qing Li, and Baobao Chang.
\newblock Ultraedit: Instruction-based fine-grained image editing at scale.
\newblock \emph{Advances in Neural Information Processing Systems}, 37:\penalty0 3058--3093, 2024{\natexlab{a}}.

\bibitem[Zhao et~al.(2024{\natexlab{b}})Zhao, Ma, Chen, Si, Wu, An, Yu, Zhang, Li, and Chang]{zhao2024ultraedit}
Haozhe Zhao, Xiaojian~Shawn Ma, Liang Chen, Shuzheng Si, Rujie Wu, Kaikai An, Peiyu Yu, Minjia Zhang, Qing Li, and Baobao Chang.
\newblock Ultraedit: Instruction-based fine-grained image editing at scale.
\newblock \emph{Advances in Neural Information Processing Systems}, 37:\penalty0 3058--3093, 2024{\natexlab{b}}.

\bibitem[Zhao et~al.(2024{\natexlab{c}})Zhao, Xiong, and Kr{\"a}henb{\"u}hl]{BSQ_2024_arxiv}
Yue Zhao, Yuanjun Xiong, and Philipp Kr{\"a}henb{\"u}hl.
\newblock Image and video tokenization with binary spherical quantization.
\newblock \emph{arXiv preprint arXiv:2406.07548}, 2024{\natexlab{c}}.

\bibitem[Zhou et~al.(2024)Zhou, Yu, Babu, Tirumala, Yasunaga, Shamis, Kahn, Ma, Zettlemoyer, and Levy]{Transfusion-7B_arxiv_2024}
Chunting Zhou, Lili Yu, Arun Babu, Kushal Tirumala, Michihiro Yasunaga, Leonid Shamis, Jacob Kahn, Xuezhe Ma, Luke Zettlemoyer, and Omer Levy.
\newblock Transfusion: Predict the next token and diffuse images with one multi-modal model.
\newblock \emph{arXiv preprint arXiv:2408.11039}, 2024.

\bibitem[Zhu et~al.(2025)Zhu, Wang, Chen, Liu, Ye, Gu, Tian, Duan, Su, Shao, Gao, Cui, Wang, Cao, Liu, Wei, Zhang, Wang, Xu, Li, Wang, Deng, Li, He, Jiang, Luo, Wang, He, Shi, Zhang, Shao, He, Xiong, Qu, Sun, Jiao, Lv, Wu, Zhang, Deng, Ge, Chen, Wang, Dou, Lu, Zhu, Lu, Lin, Qiao, Dai, and Wang]{zhu2025internvl3exploringadvancedtraining}
Jinguo Zhu, Weiyun Wang, Zhe Chen, Zhaoyang Liu, Shenglong Ye, Lixin Gu, Hao Tian, Yuchen Duan, Weijie Su, Jie Shao, Zhangwei Gao, Erfei Cui, Xuehui Wang, Yue Cao, Yangzhou Liu, Xingguang Wei, Hongjie Zhang, Haomin Wang, Weiye Xu, Hao Li, Jiahao Wang, Nianchen Deng, Songze Li, Yinan He, Tan Jiang, Jiapeng Luo, Yi~Wang, Conghui He, Botian Shi, Xingcheng Zhang, Wenqi Shao, Junjun He, Yingtong Xiong, Wenwen Qu, Peng Sun, Penglong Jiao, Han Lv, Lijun Wu, Kaipeng Zhang, Huipeng Deng, Jiaye Ge, Kai Chen, Limin Wang, Min Dou, Lewei Lu, Xizhou Zhu, Tong Lu, Dahua Lin, Yu~Qiao, Jifeng Dai, and Wenhai Wang.
\newblock Internvl3: Exploring advanced training and test-time recipes for open-source multimodal models, 2025.
\newblock URL \url{https://arxiv.org/abs/2504.10479}.

\bibitem[Zhuang et~al.(2025)Zhuang, Xie, Deng, Liang, Ru, Yin, and Zou]{zhuang2025vargpt}
Xianwei Zhuang, Yuxin Xie, Yufan Deng, Liming Liang, Jinghan Ru, Yuguo Yin, and Yuexian Zou.
\newblock Vargpt: Unified understanding and generation in a visual autoregressive multimodal large language model.
\newblock \emph{arXiv preprint arXiv:2501.12327}, 2025.

\end{thebibliography}
